\let\LaTeXcline\cline\documentclass[default,iicol]{sn-jnl}\let\cline\LaTeXcline

\usepackage{graphicx}%
\usepackage{multirow}%
\usepackage{amsmath,amssymb,amsfonts}%
\usepackage{amsthm}%
\usepackage{mathrsfs}%
\usepackage[title]{appendix}%
\usepackage{xcolor}%
\usepackage{textcomp}%
\usepackage{manyfoot}%
\usepackage{booktabs}%
\usepackage{algorithm}      
\usepackage{algorithmicx}%
\usepackage[noend]{algpseudocode} 
\usepackage{listings}%
\usepackage[sort]{natbib}%
\usepackage{pax}

\usepackage[utf8]{inputenc} 
\usepackage[T1]{fontenc}    
\usepackage{hyperref}       
\usepackage{url}            
\usepackage{nicefrac}       
\usepackage{microtype}      

\newcommand{\algrule}[1][.2pt]{\par\vskip.1\baselineskip\hrule height #1\par\vskip.1\baselineskip}

\theoremstyle{thmstyleone}%
\theoremstyle{thmstyletwo}%

\theoremstyle{thmstylethree}%

\begin{document}

\title{Masked Image Modeling: A Survey}


\author[1]{\fnm{Vlad} \sur{Hondru}}

\author[1]{\fnm{Florinel Alin} \sur{Croitoru}}

\author[2]{\fnm{Shervin} \sur{Minaee}}

\author*[1]{\fnm{Radu Tudor} \sur{Ionescu}}\email{raducu.ionescu@gmail.com}

\author[3]{\fnm{Nicu} \sur{Sebe}}

\affil[1]{\orgdiv{Department of Computer Science},
\orgname{University of Bucharest}, \orgaddress{\street{14 Academiei},
\city{Bucharest},
\postcode{010014},
\country{Romania}}}

\affil[2]{\orgdiv{Applied AI Team},
\orgname{Amazon}, \city{Seattle}, \country{USA}}

\affil[3]{\orgdiv{Department of Information Engineering and Computer Science},
\orgname{University of Trento}, \orgaddress{\street{9 via Sommarive},
\city{Povo-Trento}, \postcode{38123},
\country{Italy}}}


\abstract{In this work, we survey recent studies on masked image modeling (MIM), an approach that emerged as a powerful self-supervised learning technique in computer vision. The MIM task involves masking some information, e.g.~pixels, patches, or even latent representations, and training a model, usually an autoencoder, to predicting the missing information by using the context available in the visible part of the input. We identify and formalize two categories of approaches on how to implement MIM as a pretext task, one based on reconstruction and one based on contrastive learning. Then, we construct a taxonomy and review the most prominent papers in recent years. We complement the manually constructed taxonomy with a dendrogram obtained by applying a hierarchical clustering algorithm. We further identify relevant clusters via manually inspecting the resulting dendrogram. Our review also includes datasets that are commonly used in MIM research. We aggregate the performance results of various masked image modeling methods on the most popular datasets, to facilitate the comparison of competing methods. Finally, we identify research gaps and propose several interesting directions of future work. We supplement our survey with the following public repository containing organized references: \url{https://github.com/vladhondru25/MIM-Survey}.}




\maketitle

\section{Introduction}

\begin{figure*}[t]
\centering
\includegraphics[width=1.0\linewidth]{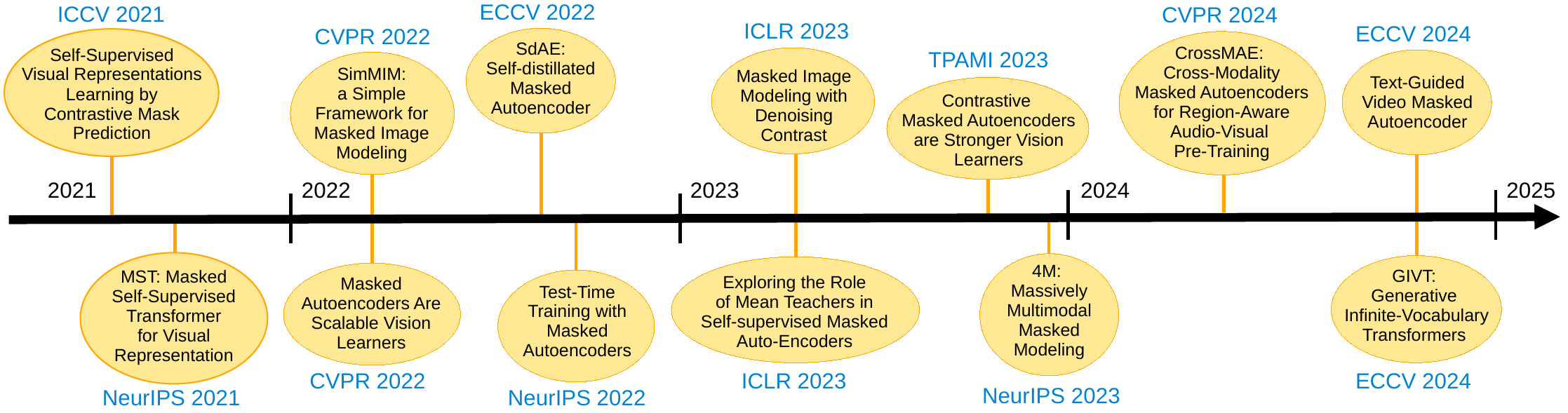}
\vspace{-0.3cm}
\caption{A timeline with the most prominent works in Masked Image Modeling. This timeline illustrates the evolution of MIM methods and the primary research directions in this area. Early works adapted masked language modeling directly to the image domain, as in MST~\citep{li-neurips-2021} and SimMIM~\citep{xie-cvpr-2022}. Subsequent efforts~\citep{yi-iclr-2023, huang-tpami-2023} tried to combine the MIM ideas with another successful self-supervised method applied in computer vision, namely contrastive learning. The most recent developments extend MIM to multimodal settings~\citep{Fan-ECCV-2024, guo-cvpr-2024} and tackle more complex tasks~\citep{Tschannen-ECCV-2024, qiu-cvpr-2024}.}
\label{fig:timeline}
\end{figure*}

Data samples have always played a crucial role in training large deep neural networks. Procuring a curated labeled dataset not only involves a great effort and a laborious process from a team of human annotators, but it also represents a significant expense. As a result, various self-supervised learning strategies have been explored, where the model is pre-trained with a different objective, which does not require human labels. Self-supervised learning can help the model to learn a rich feature representation, and even surpass supervised alternatives \citep{he-cvpr-2022}. Then, during the fine-tuning phase, the model is further optimized for a specific downstream task. For example, in most image classification tasks, a common approach is to train the network on the ImageNet dataset \citep{Deng-CVPR-2009,russakovsky-ijcv-2015}, and then change the classification head for a new task and train the resulting model on a relevant dataset. 

Self-supervised learning is a popular method to pre-train a deep learning model. It involves creating a supervised setting without annotating the data, but rather using the inherent structure of the data. Due to its potential, self-supervised pre-training algorithms have been rapidly adopted in computer vision in recent years. 
The early works in this direction by \cite{doersch-iccv-2015} and \cite{noroozi-eccv-2016} were inspired by the jigsaw puzzle idea, in which the authors proposed to split the image into patches and estimate the position of each patch. Another pre-training strategy was introduced by \cite{wang-iccv-2015}, where the objective was to have the distance between the initial and last frame in a video smaller than the distance between the initial frame and a random frame from another video. \cite{pathak-cvpr-2017} created a segmentation map from the motion of objects in videos, and then applied a segmentation task for pre-taining, using the artificial segmentation map as ground-truth. Other prominent pretext tasks are colorizing a gray-scale image \citep{zhang-eccv-2016} or rotating the input image and estimating the angle it was rotated with \citep{gidaris-iclr-2018}. 

The idea of pre-training a model by masking a part of the input and then predicting the masked information gained traction with the introduction of Bidirectional Encoder Representations from Transformers (BERT) \citep{devlin-naacl-2019}, which brought significant advancements in the Natural Language Processing domain. The main advantage was that huge amounts of unlabeled and unstructured text data could be used. Notably, this pretext strategy was applied in vision problems earlier, by \cite{vincent-jmlr-2010} and \cite{pathak-cvpr-2016}, where the original input signal was altered (corrupted or obscured) and then reconstructed using an architecture based on convolutional layers. The latter work even employed an encoder-decoder model to inpaint the missing regions from the input image. Nevertheless, as presented by \cite{he-cvpr-2022}, there are some differences from the recent Masked Image Modeling (MIM) literature, as the aforementioned papers framed the problem as a denoising task.

In Figure~\ref{fig:timeline}, we present a timeline with the most prominent works in masked image modeling. At ICCV 2021, \cite{zhao-iccv-2021} presented a self-supervised learning method that applies a contrastive loss between a masked patch from an image and other regions. At NeurIPS 2021, \cite{li-neurips-2021} proposed a teacher-student framework that employs both a reconstruction and a contrastive objective. At CVPR 2022, \cite{he-cvpr-2022} and \cite{xie-cvpr-2022} introduced a pre-training framework that involves masking a high portion of the input and reconstruct it using autoencoders based on the Vision Transformer (ViT) architecture \citep{dosovitskiy-arXiv-2020}. Their concurrent studies represent the base for the research that followed. Following \cite{li-neurips-2021}, \cite{chen-eccv-2022} formally presented a pre-training method that involves both objectives, but follows the masked autoencoder (MAE) framework. Besides employing MAE for pre-training, \cite{gandelsman-neurips-2022} demonstrated how to use MAE at inference time, improving performance on downstream tasks. At ICLR 2023, \cite{yi-iclr-2023} combined MIM with denoising contrastive learning for a better feature learning, while \cite{lee-iclr-2023} analyzed the teacher-student MIM framework and showed the advantages of updating the teacher's weights as an exponential moving average of the student's. \cite{huang-tpami-2023} is another important stepping stone in integrating both reconstruction and contrastive objectives in the MAE framework. At NeurIPS 2023, \cite{mizrahi-neurips-2023} presented a method that applies the masking pre-training method on multiple input vision modalities, as well as text. More recent contributions are focused on multimodal settings \citep{Fan-ECCV-2024, guo-cvpr-2024} and more complex tasks \citep{Tschannen-ECCV-2024, qiu-cvpr-2024}.

Within the context of more complex architectures of the latest neural networks and the large quantities of annotated data they require, pre-training such models has started to become a prerequisite. Masked Image Modeling represents a pretext task that consists of masking some information from the input (either the raw signal or some features obtained from it), and then estimating an output that should be the same as if the input was unaltered, or even predicting the original input. This pre-training strategy has quickly become popular, especially since the mechanism is easily implemented with the well-known transformer architecture, and thus it emerged in many domains and tasks. As a result, it is very difficult and time-consuming to study such a high number of research papers and find the necessary information. Our work aims to mitigate this challenge and facilitate further research or industrial endeavors. Firstly, we present a generic framework that all masked image modeling methods follow and identify two different categories of approaches: one involving input reconstruction and one employing a contrastive objective. Furthermore, we have carefully reviewed the most recent papers and extracted the main ideas and contributions from these studies. Furthermore, we manually organize the reviewed studies into a taxonomy based on multiple criteria, which is complemented by a dendrogram obtained via hierarchical clustering.

Given the increasing prominence of masked image modeling and the corresponding rise in research publications on this topic, several studies have been conducted with objectives similar to ours: to facilitate the literature review process. Among these, the survey on masked modeling by \cite{li-arxiv-2023} stands out as a notable contribution, with which our paper shares many characteristics, such as the general frameworks employed during pre-training, some criteria of the manual taxonomy, or even some presented papers. Nevertheless, our survey is focused on vision and how MIM is applied in the most recent techniques. The work of \cite{li-arxiv-2023} surveys a broad range of domains, while including a limited number of papers per domain. Furthermore, the more focused scope of our survey on the image domain allows us to review a greater number of vision papers for a more thorough study. Thus, we consider our work to be more comprehensive, particularly in the computer vision domain. In contrast to \cite{li-arxiv-2023}, we organize the papers via both manual and automatic clustering, providing complementary ways to categorize the surveyed papers. Other relevant surveys are the ones conducted by \cite{Zhang-IJCAI-2023} and \cite{Zhou-Access-2023}, where the authors focus specifically on masked autoencoders. Unlike these related surveys, we address the broader domain of masked image modeling, which is not necessarily coupled with autoencoders. Our survey is therefore more comprehensive, containing more than 100 additional references compared with both surveys on masked autoencoders \citep{Zhang-IJCAI-2023,Zhou-Access-2023}. For example, we review works that integrate masking into basic operations, e.g.~convolutions \citep{ristea-cvpr-2022, madan-tpami-2024}. As shown by the respective authors, masked convolutions can be integrated into any architecture. 

To highlight the aim of our survey, we sum up our contributions as follows:
\begin{itemize}
  \item We highlight two categories of approaches on how to implement masked image modeling as a pretext task.
  \item We review the most prominent papers in recent years, and construct a taxonomy that facilitates studying the related literature.
  \item We apply a hierarchical clustering algorithm on the abstracts and identify relevant clusters via manually inspecting the resulting dendrogram.
  \item We review commonly used datasets and aggregate the results of various masked image modeling methods in a single place, to facilitate the comparison of competing methods. 
  \item We identify research gaps and propose several interesting directions of future work in the area of masked image modeling.
\end{itemize}

\section{Generic Framework}

Masked Image Modeling is an unsupervised technique that is usually applied during the pre-training phase. It involves masking some information, either from the input or from the latent representation, and then estimating the original data, as if the data would not have been concealed. Although many masked image modeling techniques have been proposed, the research has been focused on two main schemes, either reconstructing the masked signal, or comparing two latent representations, one for the unaltered input signal and one for the masked input. On a few occasions, different approaches have been explored, but they are built on similar grounds. Therefore, in the following subsections, we aim to give a general formulation of the first two aforementioned schemes. 

\begin{figure}[t]
\centering
\includegraphics[width=0.98\linewidth]{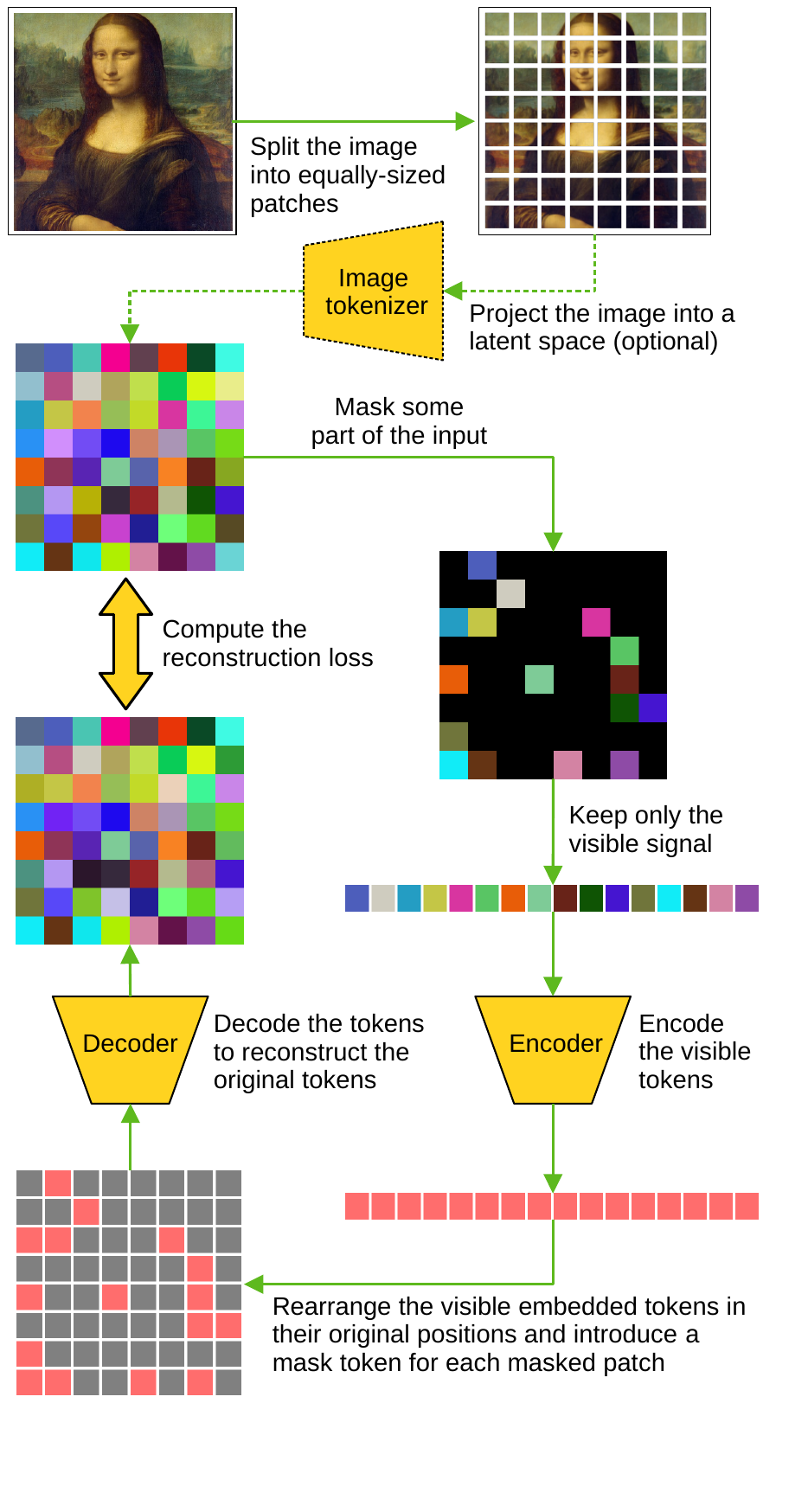}
\caption{Reconstruction-based MIM pipeline. The input image is split into patches. Some of the resulting patches are masked, and the remaining patches are passed through an encoder. Next, latent vectors corresponding to masked and visible patches are passed through a decoder. Finally, a reconstruction loss is computed between the output patches and the original input patches. The whole purpose of this self-supervised pipeline is to generate a robust latent representation by learning to reconstruct masked patches. Best viewed in color.}
\label{reconstruction}
\end{figure}

\subsection{Reconstruction}
\label{section_rec}
The first scheme that we identified revolves around the idea of masking some piece of information at any stage during the forward pass of the model, and then employing a decoder to reconstruct the missing data. We illustrate the reconstruction framework in Figure~\ref{reconstruction}. Typically, the input tokens in this pipeline correspond to patches of raw pixels \citep{he-cvpr-2022, xie-cvpr-2022}. However, there are scenarios in which these tokens reside in a latent space~\citep{chang-cvpr-2022, li-cvpr-2023a}. In Figure~\ref{reconstruction}, we illustrate this optional latent space projection through an image tokenizer, which can vary in complexity. For instance, \cite{chang-cvpr-2022} employed a VQ-GAN \citep{Esser-CVPR-2021} tokenizer to transform images into sequences of semantic visual tokens. We emphasize that all recent advances \citep{Zhu-NeurIPS-2024, Chen-CVPR-2023d, Wang-NeurIPS-2024, Li-CVPR-2025} on image tokenizers can be easily integrated into our generic formulation.

The reconstruction-based pre-training framework was concurrently introduced by \cite{he-cvpr-2022} and \cite{xie-cvpr-2022}. Both studies employ an encoder based on the ViT architecture \citep{dosovitskiy-arXiv-2020}, in which a significant portion of the input tokens is masked. The encoder processes only the visible tokens. After the encoding stage, the masked tokens are replaced with a special mask token, and a decoder reconstructs their original content. In the Masked Autoencoder (MAE) proposed by \cite{he-cvpr-2022}, the decoder is lighter than the encoder, which results in an asymmetric architecture. The loss function is applied only to the output corresponding to the masked tokens. 

After pre-training, the resulting encoder can be repurposed for feature extraction in downstream tasks, thanks to its strong representational capacity. Moreover, the decoder can be employed for generation, as demonstrated by \cite{chang-cvpr-2022} and \cite{li-cvpr-2023a}. The generative process involves iterative model evaluations and begins with an image in which all tokens are initially masked. During each iteration, some of the masked tokens are replaced with predictions from the model. This process continues for a fixed number of steps, with the final iteration replacing any remaining masked tokens with the model's predictions, ultimately producing a complete image.

\begin{algorithm}[t]
\caption{Reconstruction-based MIM \label{reconstruction_alg}}
\textbf{Models}:
$T_\omega$ -- the tokenizer;
$E_\theta$ -- the encoder;
$D_\phi$ -- the decoder.

\vspace{0.2em}
\algrule
\vspace{0.2em}
\textbf{Input}: 
$X$ -- the input image;
$h, w$ -- the patch dimensions;
$\alpha$ -- the proportion of input masking;
$\mbox{split}$ -- the function that splits an image into a number of patches;
$\mbox{mask}$ -- the function which chooses what patches should be masked;
$M$ -- the learnable embedding of the masked patches;
$d$ -- the optimization function;
$\eta$ -- the learning rate.
\vspace{0.2em}
\algrule
\vspace{0.2em}

\textbf{Computation}:
\begin{algorithmic}[1]

\State  $P = \{p_i| p_i \in \mathbb{R}^{h\times w\times c}\}_{i=1}^n \leftarrow \mbox{split}(X,h,w)$
\State $P = T_\omega(P)$
\State $I_v, I_m \leftarrow \mbox{mask}(P, \alpha, n)$

\For{$i \in \{1, \dots, n\}$}
    \If{$i \in I_v$}
        \State $H[i] \leftarrow E_\theta (p_i)$
    \ElsIf{$i \in I_m$}
        \State $H[i] \leftarrow M$
    \EndIf
    
\EndFor

\State $\hat{P} \leftarrow D_\phi(H)$
\State $\mathcal{L}\left(\phi,\varphi, \theta, M\right) \leftarrow d\left( \hat{P}, P\right)$
\State $\theta \leftarrow \theta - \eta \cdot \frac{\partial\mathcal{L}}{\partial \theta}$
\State $\phi \leftarrow \phi - \eta \cdot \frac{\partial\mathcal{L}}{\partial \phi}$
\State $M \leftarrow M - \eta \cdot \frac{\partial\mathcal{L}}{\partial M}$
\end{algorithmic}
\end{algorithm}

We formally present the reconstruction-based training strategy in Algorithm~\ref{reconstruction_alg}. The first step of the algorithm splits the input images into non-overlapping patches, resulting in a set $P$ that contains patches of the same size, namely of $h\times w$ pixels. The second step applies the tokenizer $T_\omega$, which can differ from one method to another. Notably, $T_\omega$ can be represented by the identity function when the MIM pre-training is directly applied on the raw pixels. The masking operation indicates which patches should be kept and which should be masked. The indexes of the visible and masked patches are stored in $I_v$ and $I_m$, respectively. In the next steps, the masked patches are usually dropped, and only the visible patches are processed by an encoder ($E_\theta$) that extracts a latent representation. Before the decoding step, the previously masked patches are replaced by a learnable representation ($M$), which resides in the latent space of the encoder. These transformations correspond to steps 3-8 in Algorithm~\ref{reconstruction_alg}.

Using the sequence formed by concatenating the masked representations with those from the encoder, the decoder ($D_\phi$) reconstructs the patches as depicted in step 9 of Algorithm~\ref{reconstruction_alg}. Finally, in steps 10-13 of Algorithm~\ref{reconstruction_alg}, a distance function is optimized by updating the encoder, the decoder, the projection layer, and the learnable masked representation.

\begin{figure}[t]
\centering
\includegraphics[width=0.90\linewidth]{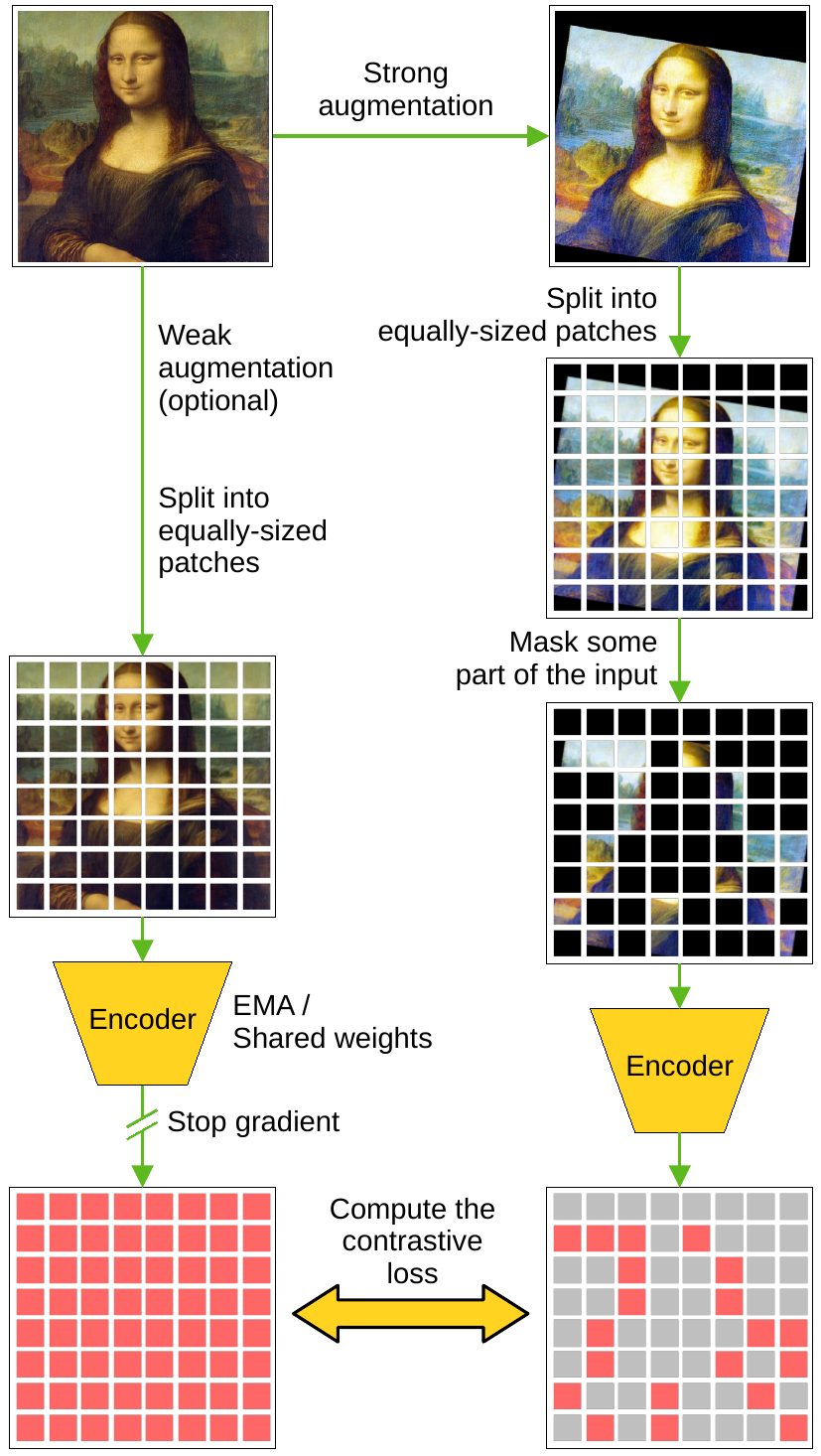}
\caption{Contrastive-based MIM pipeline. Two versions of the input image are used in this framework, one that is unaltered (or weakly augmented) and one that is strongly augmented and masked. The images are processed by two encoders, a teacher encoder (left) and student encoder (right). The teacher encoder is either identical to the student encoder, or an exponential moving average (EMA) of the student encoder. The training is based on a contrastive loss applied on the latent representations of the patches. Gradients are propagated only through the student encoder. Best viewed in color.}
\label{contrastive}
\end{figure}

\subsection{Contrastive}
\label{section_contrastive}
The second generic scheme is represented by comparing two different latent representations of the same input. One latent representation corresponds to an unaltered or weakly augmented input image, while the other corresponds to a masked and strongly augmented version of the same input image. The approach is based on a contrastive learning framework, as illustrated in Figure~\ref{contrastive}. There are two common architectural configurations. One configuration uses an encoder with shared weights, as in \citep{wu-cvpr-2023a, zhang-iclr-2023}. The other configuration uses an encoder for the masked input and an exponential moving average (EMA) version of the encoder for the original input, as in \citep{lee-iclr-2023}.

\begin{algorithm}[t]
\caption{Contrastive-based MIM \label{contrastive_alg}}
\textbf{Models}:
$V\!P_{\varphi}$ -- the visual projection layer;
$E_\theta$ -- the encoder.

\vspace{0.2em}
\algrule
\vspace{0.2em}
\textbf{Input}: 
$X$ -- the input image;
$h, w$ -- the patch dimensions;
$\alpha$ -- the proportion of input masking;
$\mbox{split}$ -- the function that splits an image into a number of patches;
$\mbox{mask}$ -- the function which chooses what patches should be masked;
$\mbox{augment}$ -- the image augmentation function;
$M$ -- the learnable embedding of the masked patches;
$d$ -- the optimization function;
$\eta$ -- the learning rate.
\vspace{0.2em}
\algrule
\vspace{0.2em}

\textbf{Computation}:
\begin{algorithmic}[1]
\State  $\hat{X} \leftarrow  \mbox{augment}(X, \mbox{`weak'})$
\State $\tilde{X} \leftarrow  \mbox{augment}(X, \mbox{`strong'})$
\State  $\hat{P} = \{\hat{p}_i| \hat{p}_i \in \mathbb{R}^{h\times w\times c}\}_{i=1}^n \leftarrow \mbox{split}(\hat{X},h,w)$
\State  $\tilde{P} = \{\tilde{p}_i| \tilde{p}_i \in \mathbb{R}^{h\times w\times c}\}_{i=1}^n \leftarrow \mbox{split}(\tilde{X},h,w)$
\State $I_v, I_m \leftarrow \mbox{mask}(\tilde{P}, \alpha, n)$

\For{$i \in \{1, \dots, n\}$}

    \State $\hat{H}[i] \leftarrow E_\theta ( V\!P_{\varphi}(\hat{p}_i) )$
    \If{$i \in I_v$}
        \State $\tilde{H}[i] \leftarrow E_\theta ( V\!P_{\varphi}(\tilde{p}_i) )$
    \ElsIf{$i \in I_m$}
        \State $\tilde{H}[i] \leftarrow E_\theta(M)$
    \EndIf

\EndFor
\State $\mathcal{L}\left(\varphi,\!\theta,\!M\right)\!\leftarrow\!d(\hat{H},\tilde{H})$
\State $\theta \leftarrow \theta - \eta \cdot \frac{\partial\mathcal{L}}{\partial \theta}$
\State $\varphi \leftarrow \varphi - \eta \cdot \frac{\partial\mathcal{L}}{\partial \varphi}$
\State $M \leftarrow M - \eta \cdot \frac{\partial\mathcal{L}}{\partial M}$
\end{algorithmic}
\end{algorithm}

The generic contrastive-based MIM framework is formalized in Algorithm~\ref{contrastive_alg}. The aim of this framework is to obtain a similar embedding, irrespective of the applied masking. Steps 1 and 2 of the algorithm generate two versions of the input image by augmenting them at different intensities. The version that undergoes masking is strongly augmented, while the other one is weakly augmented (or can even remain unaltered). In steps 3-4, each image is divided into non-overlapping patches, and in step 5, masking is applied solely to the strongly augmented image. The unmasked image undergoes processing by the projection layer and the encoder (step 7), whereas the masked image has its omitted patches replaced with a learnable vector before being encoded (steps 8-11). Notably, Algorithm~\ref{contrastive_alg} processes both input images using the same encoder, which can be regarded as two encoders with shared weights. Gradient propagation is restricted to the processing of the masked image. An alternative approach~\citep{lee-iclr-2023} is to process the unmasked image with an EMA-based encoder in steps 9 and 11. Step 12 of the algorithm computes the contrastive loss, in which the negative patches originate from different positions within the same image. The most common loss function used in Step 12 is some variant of the InfoNCE (Noise-Contrastive Estimation) loss:
\begin{equation}
\mathcal{L}\left(\varphi,\!\theta,\!M\right)\!\leftarrow\!-\!\sum_{i=1}^n\!\log{\frac{\exp(\langle\hat{H}[i],\tilde{H}[i]\rangle/\tau)}{\sum_{j=1}^n \exp(\langle\hat{H}[j],\tilde{H}[i]\rangle/\tau)}},    
\end{equation}
where $\tau$ is the temperature factor scaling the pairwise similarity. The goal of InfoNCE is to maximize the similarity between $\hat{H}[i]$ and $\tilde{H}[i]$, while minimizing the similarities $\hat{H}[j]$ and $\tilde{H}[i]$, $\forall j \neq i$. In practice, the negative patches denoted by $\hat{H}[j]$ can also be sourced from other images. Finally, the last steps (13-15) of the algorithm update the weights of the encoder, the projection layer, and the learnable representation $M$.
Although the contrastive learning method is technically different from the reconstruction-based scheme, \cite{zhang-neurips-2022} demonstrated that the contrastive approach strongly correlates with the reconstruction-based framework in terms of the learned latent representations.

\begin{figure*}[!th]
\centering
\includegraphics[width=0.965\linewidth]{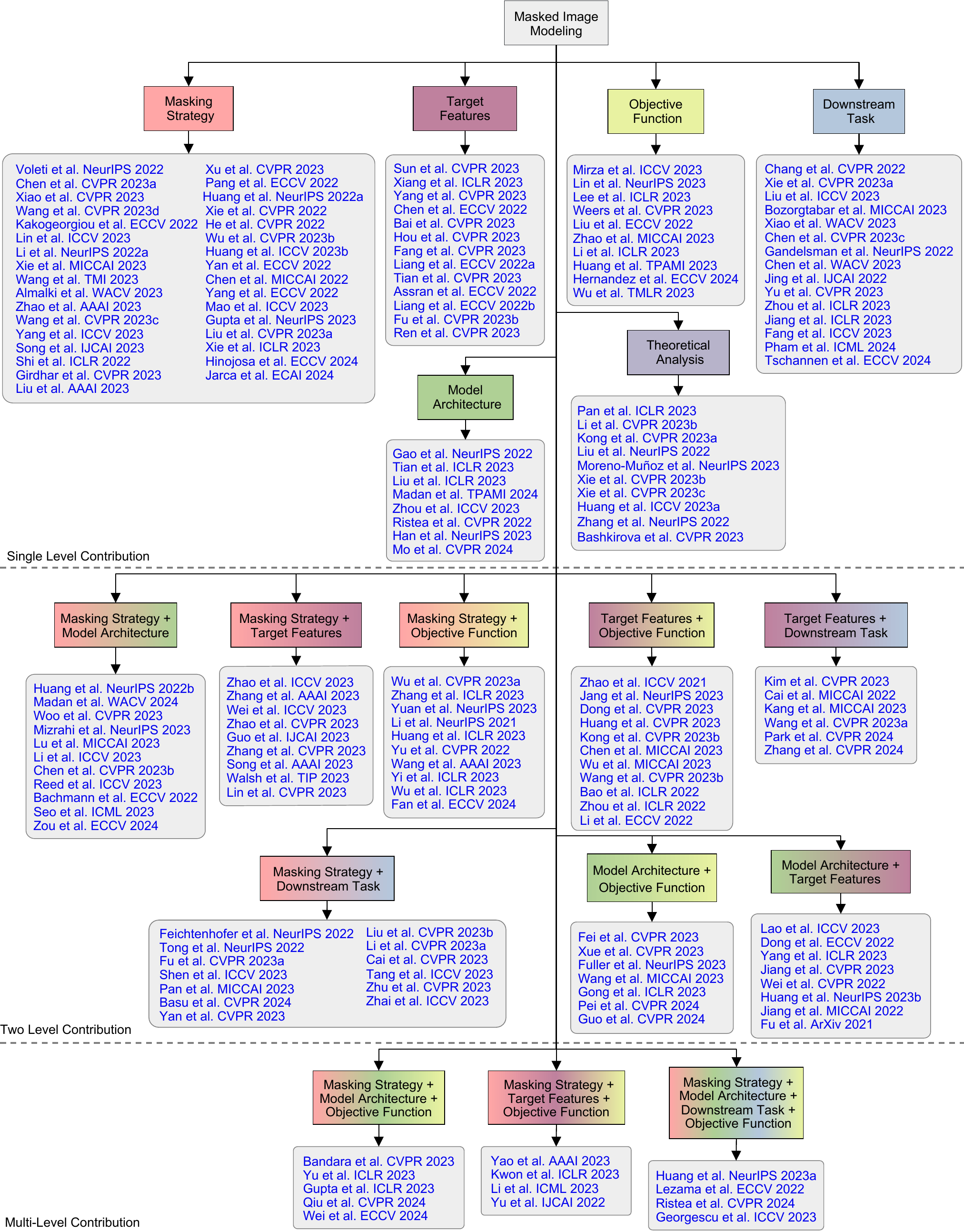}
\caption{A multi-level classification of Masked Image Modeling papers into various categories, based on the research directions studied in the respective papers. References are clickable links to papers. Best viewed in color.}\label{fig_taxonomy}
\vspace{-0.2cm}
\end{figure*}

\subsection{Relation to Other SSL methods}
\noindent
\textbf{Contrastive learning.} The standard contrastive learning paradigm \citep{Chen-ICML-2020, Grill-NeurIPS-2020, He-CVPR-2020} is to learn representations by pulling different augmented views of the same image closer in the embedding space, while pushing views of different images apart. In Section~\ref{section_contrastive}, we show that MIM and contrastive learning are not mutually exclusive. However, if we compare the standard contrastive learning with MIM, we can identify a few different properties.
First, MIM is naturally well-suited to transformer architectures, as it draws inspiration from masked language modeling, which was originally designed for such models. Second, unlike contrastive learning, MIM supports both recognition and generation tasks, as discussed in Section~\ref{section_rec}. In terms of representation learning, MIM encourages the model to focus on spatially meaningful features, because it reconstructs masked regions of the input. Lastly, contrastive learning relies on negative examples, which increases the batch size and often requires hard sample mining, a procedure that introduces additional computational overhead and complexity during training \citep{Ghita-PAKDD-2024}.

\noindent
\textbf{Non-contrastive methods.} Early self-supervised learning approaches primarily focused on creating pretext tasks, e.g. solving jigsaw puzzles \citep{noroozi-eccv-2016,Chen-CVPR-2021}, learning the arrow of time \citep{Wei-CVPR-2018}, or colorizing images \citep{zhang-eccv-2016}. While conceptually simple, these methods often exhibit limited generalization, because the objectives of the pretext tasks are not always aligned with those of downstream tasks. More recent studies~\citep{caron-iccv-2021, Caron-NeurIPS-2020, Bardes-ICLR-2022} leverage semantic similarities between samples. The core idea is to employ a dual-network architecture, where both networks are trained to generate similar embeddings for different augmented views of the same image. Notably, this approach differs from contrastive learning, because it eliminates the need for negative examples.

Compared with MIM, the aforementioned self-supervised approaches differ in several key aspects. First, they can be less stable during training and often rely on additional techniques, such as momentum encoders, to ensure convergence. Second, unlike alternative methods, MIM inherently supports image generation. Lastly, while MIM focuses on learning spatial and contextual features through reconstruction, other self-supervised learning methods prioritize strong global representations by enforcing consistency across different views of the same image.

\section{Taxonomy and Overview}

In Figure~\ref{fig_taxonomy}, we present a manually-generated taxonomy of the most promising MIM papers, organizing them according to their main contributions. In constructing the taxonomy, we consider six main research directions related to: the masking strategy, the type of masked signals, the neural architecture, the objective function, the type of downstream tasks, and theoretical results. Nonetheless, since many papers focused on more than one of the above aspects, we grouped the respective works into distinct categories representing composed contributions. We now continue by presenting the aforementioned papers, divided into sub-sections according to our taxonomy.


\subsection{Masking Strategy}
\begin{figure*}[t]
    \centering
    \includegraphics[width=0.8\linewidth]{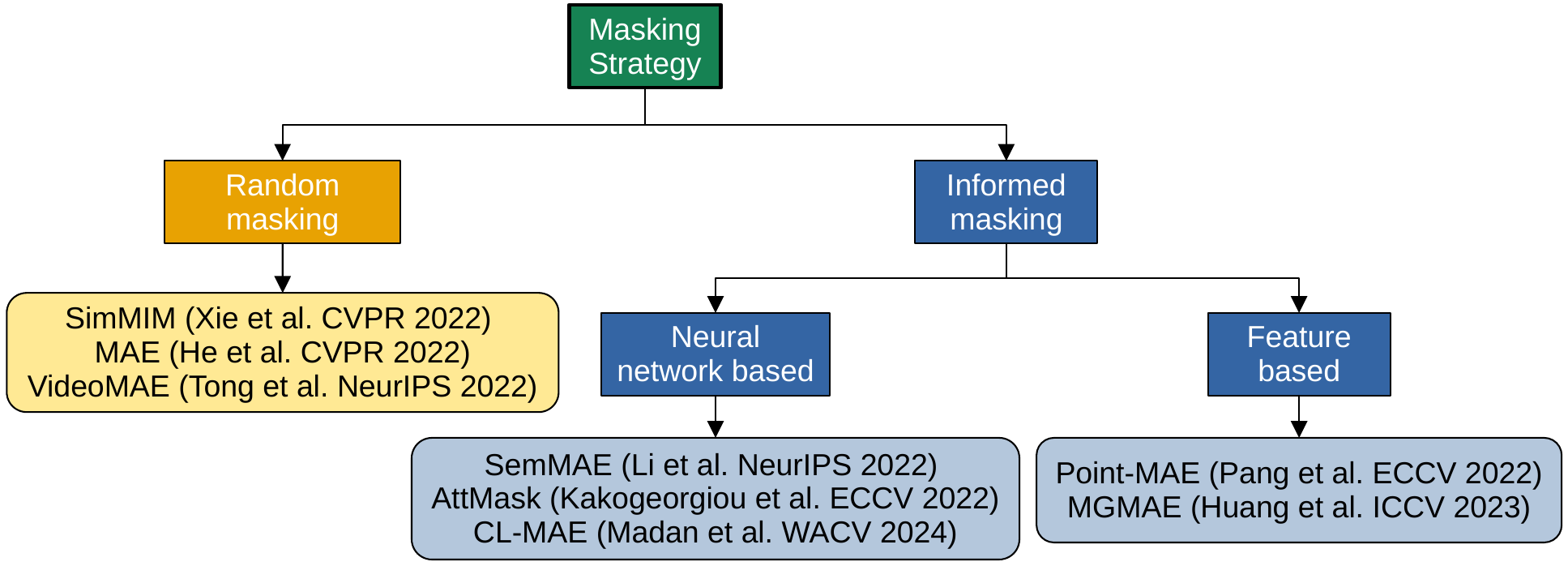}
    \caption{Types of masking strategies employed in MIM pipelines. The masking strategies can either rely on random patch selection or leverage some additional information to selectively mask the salient patches. Additional information can come from auxiliary neural networks or from features derived through classical computer vision algorithms.}
    \label{fig:masking_strategy}
\end{figure*}

In Figure~\ref{fig:masking_strategy}, we categorize the main masking strategies employed in MIM pipelines. A common practice is to mask the content randomly \citep{xie-cvpr-2022, he-cvpr-2022, tong-neurips-2022, shi-iclr-2022,chen-cvpr-2023a, girdhar-cvpr-2023, liu-cvpr-2023a}. Nevertheless, several recent works advocate for a more informed masking \citep{pang-eccv-2022, kakogeorgiou-eccv-2022, li-neurips-2022a, voleti-neurips-2022, huang-neurips-2022a, chen-miccai-2022, xiao-cvpr-2023, wang-cvpr-2023d, xu-cvpr-2023, wu-cvpr-2023b, lin-iccv-2023}, guided by meaningful information, whether it stems from a neural network~\citep{kakogeorgiou-eccv-2022, li-neurips-2022a, madan-wacv-2024} or from classical features~\citep{pang-eccv-2022,  voleti-neurips-2022, huang-neurips-2022a,jarca-ECAI-2024}. By masking the salient patches or semantically relevant regions, these informed approaches enhance the capacity of the model to learn robust representations.
\begin{figure}
    \centering
    \includegraphics[width=1.\linewidth]{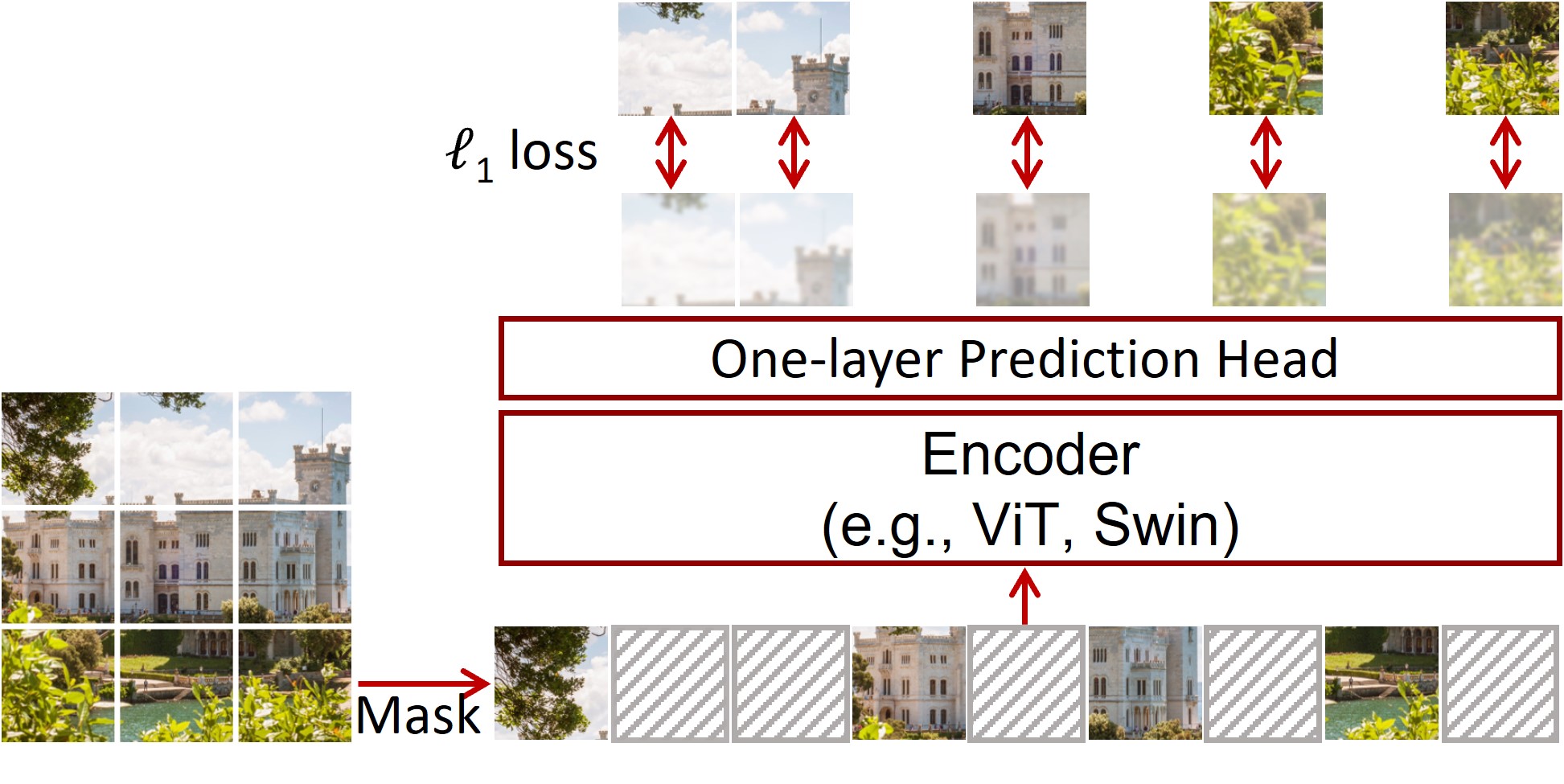}
    \caption{SimMIM pipeline, courtesy of \cite{xie-cvpr-2022} (image licensed under CC BY 4.0). SimMIM is a reconstruction-based method (belonging to the generic pipeline in Figure \ref{reconstruction}), as it employs an ${L}^1$ loss between the masked patches and their counterparts from the original image.}
    \label{fig:simMIM}
\end{figure}

\citet{xie-cvpr-2022} introduced SimMIM, a self-supervised pre-training framework that reconstructs pixel values of randomly masked images. An overview of the method is depicted in Figure~\ref{fig:simMIM}. The model is a simple encoder (ViT) receiving as input an image with masked patches replaced by a learnable token. Additionally, the study motivates the choice of random masking and the raw pixels as target features through extensive ablation studies.

\begin{figure}
    \centering
    \includegraphics[width=1.\linewidth]{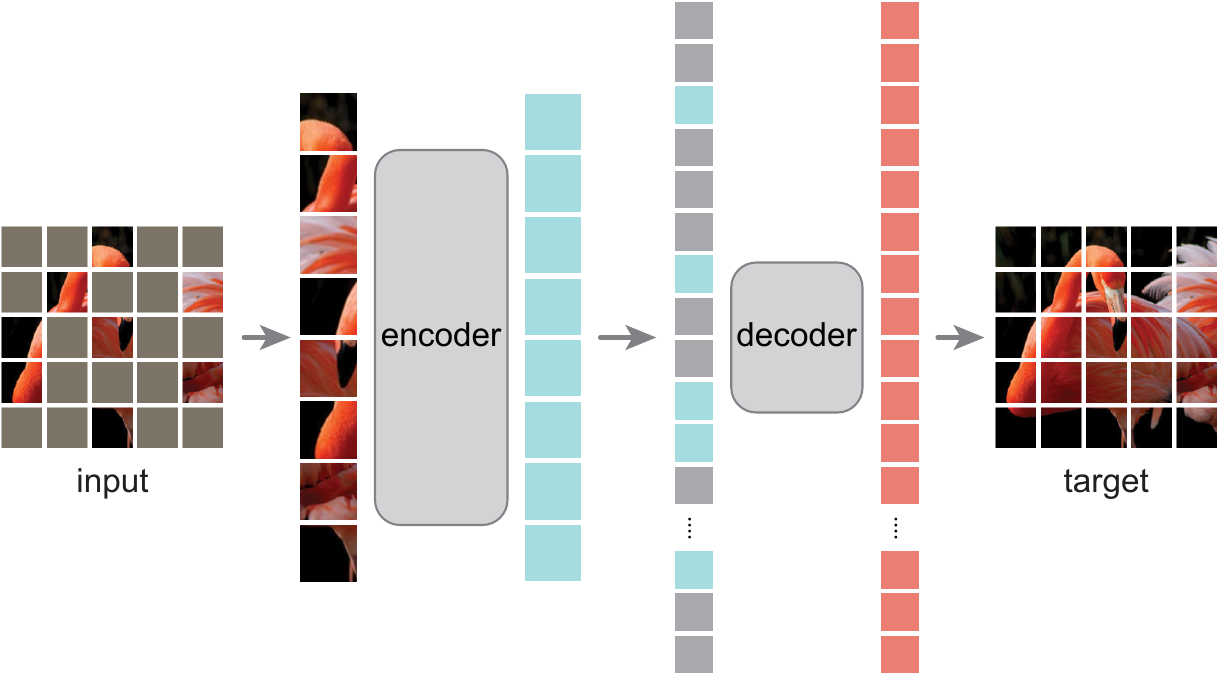}
    \caption{MAE pipeline, courtesy of \cite{he-cvpr-2022} (image licensed under CC BY 4.0). MAE is a reconstruction-based method (belonging to the generic pipeline in Figure \ref{reconstruction}), as it employs an ${L}^2$ loss between the reconstructed patches and their counterparts from the original image.}
    \label{fig:MAE}
\end{figure}

\cite{he-cvpr-2022} presented the masked autoencoder (MAE) framework, where the main contribution is to exclude the masked tokens from the encoder's input and use a very high masking ratio (75\%). These changes imply a more efficient framework compared with other works, such as SimMIM~\citep{xie-cvpr-2022}. However, as a negative effect, having an encoder that operates only on visible tokens makes the framework incompatible by default with the Hierarchical Vision Transformer~\citep{Liu-ICCV-2021} architecture, which usually performs better than a standard ViT. The MAE pipeline is presented in Figure~\ref{fig:MAE}. Aiming to learn more robust representations, 
OmniMAE, introduced by \cite{girdhar-cvpr-2023}, is similar to the standard MAE framework, except that the model is trained with both video and image data. 
Another extension is MixMAE \citep{liu-cvpr-2023a}, a pre-training approach that leverages elements of both MAE \citep{he-cvpr-2022} and SimMIM \citep{xie-cvpr-2022}. This method selects two distinct images from a training dataset and extracts random tokens from each. The tokens are then amalgamated to form a new, hybrid image. This composite image undergoes processing by an encoder. Further, a decoder is employed, aiming to reconstruct the original two images from which the mixed image was derived. During the decoding phase, the input provided to the decoder is unmixed and the patches corresponding to the missing tokens in the hybrid image are filled with masked tokens.

\cite{pang-eccv-2022} adapted the masked pre-training strategy to 3D Point Clouds. The method groups the points into patches and masks some of them. Then, it embeds and encodes only the unmasked patches. Next, the visible patches get concatenated with the masked tokens and passed through the decoder, with the objective to reconstruct the latter. The authors state that passing the masked patches earlier leaks spatial information which simplifies the task. Additionally, 
\cite{xu-cvpr-2023} showed that masked image modeling is more effective on 3D scenes when certain points are excluded from masking, the so-called \emph{Informative Points}. Keeping these points unchanged will help to preserve the geometric structure of the scene after the masking process.

A number of studies used low-level information to improve the masking strategy.
\cite{jarca-ECAI-2024} formulated the masking strategy as as curriculum learning problem by gradually increasing the masking ratio during training. Additionally, they introduced a patch selection strategy that emphasizes masking patches with higher gradient magnitudes. The masking strategy proposed by~\cite{hinojosa-ECCV-2024} is inspired by the color noise used in image processing. Specifically, their approach introduces four filters that operate on the uniform noise typically generated for random masking. This strategy avoids the extra computational overhead often required by other informed masking methods, while still outperforming standard random masking. 
Rather than imposing a masking policy, several studies tried to learn the optimal masking strategy. 
SemMAE~\citep{li-neurips-2022a} deployed an additional stage before masking. This stage, called \emph{semantic part learning}, is responsible for learning attention maps that correspond to meaningful semantic components in the image. The training of this stage is performed by embedding the class token provided by a ViT encoder into part embeddings. The resulting embeddings together with the patch embeddings provided by the same encoder are used in an attention layer. Finally, the obtained attention maps are processed by a decoder to reconstruct the original image. In the second stage, the attention maps are used for part segmentation. The masking varies from masking patches in each part to masking entire parts randomly. In a similar fashion, \cite{kakogeorgiou-eccv-2022} proposed a framework in which the masking strategy is learned. A teacher network, seeing the intact image, generates a mask that is used to pre-train the student. The objective is to reconstruct the feature representation of the teacher for the masked tokens, as well as the class token \emph{[CLS]} that is used in generating the mask. The teacher's parameters are updated using an exponential moving average of the student's weights. The experiments demonstrated the superiority of the method over other masking strategies. Instead of relying on an additional teacher network, 
\cite{wang-cvpr-2023d} developed a masking strategy based on the direct feedback provided by the model being pre-trained. The strategy computes the reconstruction loss for each patch and then selectively masks the patches that are more challenging to reconstruct, as indicated by their higher loss values. This approach ensures that the model focuses on the most difficult aspects of the data during training. Similarly, but leveraging attention maps, 
\cite{liu-aaai-2023} proposed a guided masking strategy for MAE, rather than a random policy. Periodically, all images are passed through the encoder and their latent feature representation of the \emph{[CLS]} token from the last attention layer is extracted in order to compute an importance map of all patches. Using this, the masked patches are sampled (higher chance for more salient patches) and then estimated via reconstruction, while a portion of the least important patches are completely put aside (i.e.~not fed to the encoder).

\cite{huang-neurips-2022a} added masked image modeling in the training pipeline of GANs. The masking is based on two methods. The first one is called \emph{shifted spatial masking} and constitutes random masking of the image. The second one is called \emph{balanced spectral masking} and randomly masks some spectral bands of the image decomposed in the spectral space. The authors observed that these two strategies are orthogonal and help the adversarial training process to become more stable. The frequency domain is also leveraged by 
\cite{xie-iclr-2023}, who proposed to apply the masks in the frequency domain. The images are converted with fast Fourier transform, then either low or high frequencies are masked, and mapped back into the pixel space. Using an encoder-decoder model, the original image is estimated and transformed into the frequency domain. The objective is to identify the frequencies that were masked. 

The work of \cite{tong-neurips-2022} pioneered the application of Masked Autoencoders (VideoMAE) to video data, adapting the principles of MAE used in image processing to the temporal domain. In this approach, sequences of frames are masked with a masking ratio of approximately 90\%. This notably high ratio is strategically chosen to effectively minimize the issue of information leakage that occurs between closely spaced frames. \cite{wang-cvpr-2023c} extended this work and introduced VideoMAEv2, a method for scaling up the original VideoMAE. Other studies~\citep{huang-iccv-2023b, wu-cvpr-2023b, gupta-neurips-2023} focused on designing more suitable masking strategies for video data. 
\cite{huang-iccv-2023b} argued that, for the successful application of MAE on video, it is crucial to consistently mask video segments across time. Without this consistency, there is a risk of temporal information leakage, rendering the learning task overly simplistic. To address this challenge, they introduced a novel solution that leverages optical flow techniques to generate time-coherent masking volumes. These volumes are then utilized to selectively sample visible tokens, ensuring that the masking process maintains temporal integrity and effectively prevents information leakage. 
\cite{wu-cvpr-2023b} proposed DropMAE, an adaptation of MAE for videos. The main observation of \cite{wu-cvpr-2023b} is that it is not enough to mask and reconstruct video patches in order to learn spatio-temporal features. The proposed solution is to drop spatial-attention tokens to guide the model towards looking at the temporal information. Driven by the same goal, 
\cite{gupta-neurips-2023} extended the MAE framework to learn temporal feature representations between the frames of a videoclip. Two frames are randomly sampled, the earlier one being split into patches, while the future one is both split and masked. Both images are then separately encoded using Siamese encoders. The resulting embeddings are passed through a decoder with cross-attention (the queries consisting of the future frame's visible token embeddings and mask tokens) in order to reconstruct the future frame. To learn video-language representations, 
\cite{lin-iccv-2023} proposed SMAUG, which stands out as an efficient framework for video-language pre-training, surpassing previous methodologies. In its pre-training process, SMAUG employed a strategy of masking a significant portion of both frame patches and text tokens, enabling simultaneous masked video modeling and masked language modeling. Targeting video generation, 
\cite{voleti-neurips-2022} introduced Masked Conditional Video Diffusion (MCVD). MCVD leverages different frame masking strategies to train a video diffusion model for unconditional video synthesis and video interpolation. At training time, the method chooses randomly and independently to mask the future or past frames of a given video. This simple strategy allows the diffusion model to perform four different tasks during inference: future and past frame prediction, unconditional generation, and frame interpolation.

 Starting from the idea that video and speech data are strongly related, \cite{shi-iclr-2022} proposed a joint masked pre-training framework for both modalities. Firstly, each data type is encoded into an intermediate latent representation using ResNet for images and a linear layer for audio samples. At each timestep, frames from both modalities are masked by replacing them with an arbitrary sampled frame from the same sequence. These are then concatenated and the information is fused through a transformer-based encoder. Inspired by \cite{hsu-arxiv-2021}, the objective is to predict which cluster every frame belongs to, the clusters being repeatedly computed by applying a discrete latent variable algorithm to the features extracted from the audio sequence.


\cite{xiao-cvpr-2023} introduced a method to fine-tune models, with the goal of achieving a more balanced trade-off between in-distribution (ID) and out-of-distribution (OOD) performance. The authors argue that fine-tuning models on a specific dataset tends to enhance their performance on ID data, but this diminishes their performance on OOD data. Therefore, they proposed an approach that involves masking certain patches within an image and replacing them with content from another image. This modified image is then used to train the model under the supervision of the pre-trained model to recognize the masked image features. Enhancing the robustness of models is also the primary objective of
\cite{chen-cvpr-2023a}, who observed that existing image denoising methods often overfit on the type of noise seen during training. Their approach focused on improving the generalization capabilities of this type of models by leveraging masked image modeling. To this end, the proposed method masks randomly chosen pixels from the input image and tokens in the self-attention layers of the transformer architecture. The use of token masking in self-attention layers effectively mimics the unreliability observed in tokens during inference, when the data is compromised by various types of noise. This simulation helps to prepare the model for real-world scenarios, where data quality may be inconsistent.

Multiple studies have proposed masking strategies that are more meaningful for medical data than random masking.
\cite{yang-iccv-2023} introduced masked relation modeling to improve self-supervised pre-training on medical images. This masking strategy uses the self-attention mechanism to identify strong dependent regions in the input image and breaks such relations by masking the most important patches for a given patch. On the same note, the cross-attention is applied between images and genome features, aiming to capture the correspondence between these two modalities. A similar approach that leverages strong dependent regions is presented by 
\cite{wang-tmi-2023}. They proposed a masking strategy that is applied to superpixels (contiguous groups of pixels with similar properties), demonstrating its capability on medical image segmentation of skin lesions. Nevertheless, after masking a proportion of superpixels, the same MIM methodology is followed: reconstructing the masked superpixels based on the visible ones. Initially, a base policy \citep{achanta-tmi-2012} is adopted to generate and mask superpixels, after which, the policy is optimized in a self-supervised manner, the model being further pre-trained with the new policy. Leveraging medical reports, 
\cite{chen-miccai-2022} presented a masked autoencoder pre-training scheme for vision and language medical data. This model consists of two encoders: one based on ViT for images and one based on BERT for texts. The embeddings of the two modalities are jointly processed with a cross-attention module in order to fuse the information. Finally, each input is reconstructed with a separate decoder: a transformer model for images and a simple multi-layer feed-forward network for texts. While the language decoder receives the output of the last layer of the multimodal module, the vision decoder uses an intermediate latent representation. Another important aspect considered by \cite{chen-miccai-2022} is that different masking ratios are used for different input types. 
Working on a similar task, \cite{xie-miccai-2023} introduced MedIM, a method that guides the masking based on radiology reports. Firstly, both inputs (the medical image and the corresponding report) are encoded using separate encoders. Then, the text embeddings are split into two subsets, according to the original word categories: MeSH and Sentence tokens. Each subset of text embeddings, together with the visual embeddings, are used to generate a mask that hides different information. The loss adds the reconstruction errors between the two resulting decoded masked embeddings (with separate heads) and the original image. In a more specific use case of dental panoramic radiographs, 
\cite{almalki-wacv-2023} adopted a masked image modeling strategy (either SimMIM or UM-MAE) for pre-training a SwinViT in order to ameliorate its need for a large training dataset. Then, the encoder of the pre-trained SwinViT is taken and used as a backbone in the detection and image segmentation downstream tasks.

\cite{mao-iccv-2023} extended the MAE approach to temporal skeleton sequences. Rather than opting for the straightforward method of reconstructing the skeleton sequence directly, this research reconstructed the temporal motion embedded within the sequence, using the masked skeleton sequences as input. In addition, the same motion is used to guide the masking of the skeletons.

\cite{song-ijcai-2023} presented a self-supervised masking-based method for the multi-agent reinforcement learning (RL) setting. For a given timestamp, the observations of all agents are taken. Some of them are masked and replaced by the values of the previous timestamp, eventually being encoded in a latent space. Then, a reconstruction model is used to generate the original agents' feature representations from the masked sequence. The authors assert that the resulting feature space is stimulated to learn more about the interaction between agents, while being more temporally aware.   

\begin{figure*}[t]
    \centering
    \includegraphics[width=\linewidth]{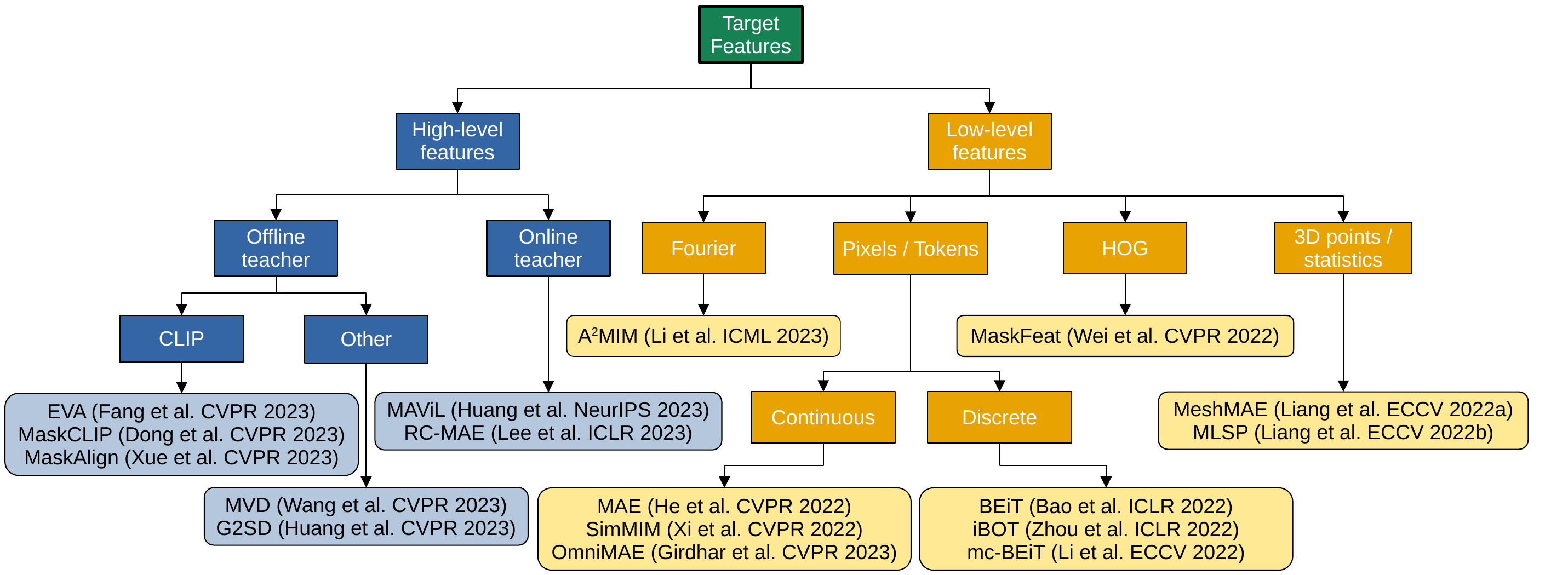}
    \caption{Types of target features employed in MIM pipelines. These features can be either low-level or high-level features. In most cases, low-level features are directly represented by the pipeline’s input. However, in certain cases, they are derived through a transformation of the input. The high-level features are extracted by neural networks that may  be either frozen or updated during the pre-training process.}
    \label{fig:target_features}
\end{figure*}

To detect malicious network traffic, \cite{zhao-aaai-2023} introduced a transformer model, solving the scant data problem by adopting MAE. During pre-training, the raw traffic is processed into a compact 2D matrix consisting of 5 packet levels, dividing it into patches and masking some of them. Then, a transformer encodes the visible tokens and tries to reconstruct the matrix with a decoder. However, during the fine-tuning step, two different encoders are created from the earlier pre-trained one, each having distinctive attention layers (compared with the global attention from the previous stage). On the one hand, an encoder operates only on the tokens within the same packet level. On the other hand, the other encoder performs attention between the patches from all packet levels.

\subsection{Target Features}

In Figure~\ref{fig:target_features}, we illustrate the main categories of target features employed across various MIM models. These features can be categorized in two primary groups, low-level and high-level features. The low-level category encompasses representations such as pixels~\citep{he-cvpr-2022, xie-cvpr-2022}, discrete tokens \citep{bao-ICLR-2022, zhou-ICLR-2022}, 3D~\citep{liang-eccv-2022a, sun-cvpr-2023, hou-cvpr-2023, tian-cvpr-2023}, HOG~\citep{wei-cvpr-2022}, and spectral features~\citep{li-icml-2022}. Methods based on low-level features often employ high masking ratios to encourage the learning of sufficiently abstract representations that are necessary to enhance performance on downstream tasks. The methods that use high-level features as target features can be split into two groups, namely those that rely on a pre-trained teacher model~\citep{fang-cvpr-2023, xue-cvpr-2023, yang-eccv-2022, bai-cvpr-2023, fu-cvpr-2023b}, and those in which the teacher and student are trained simultaneously \citep{huang-neurips-2023a, kakogeorgiou-eccv-2022, chen-eccv-2022, assran-eccv-2022masked, yang-cvpr-2023, zhao-miccai-2023}. In the remainder of this section, we briefly describe the MIM pipelines that primarily contribute through the types of target features they employ.



\begin{figure*}[t]
    \centering
    \includegraphics[width=0.8\linewidth]{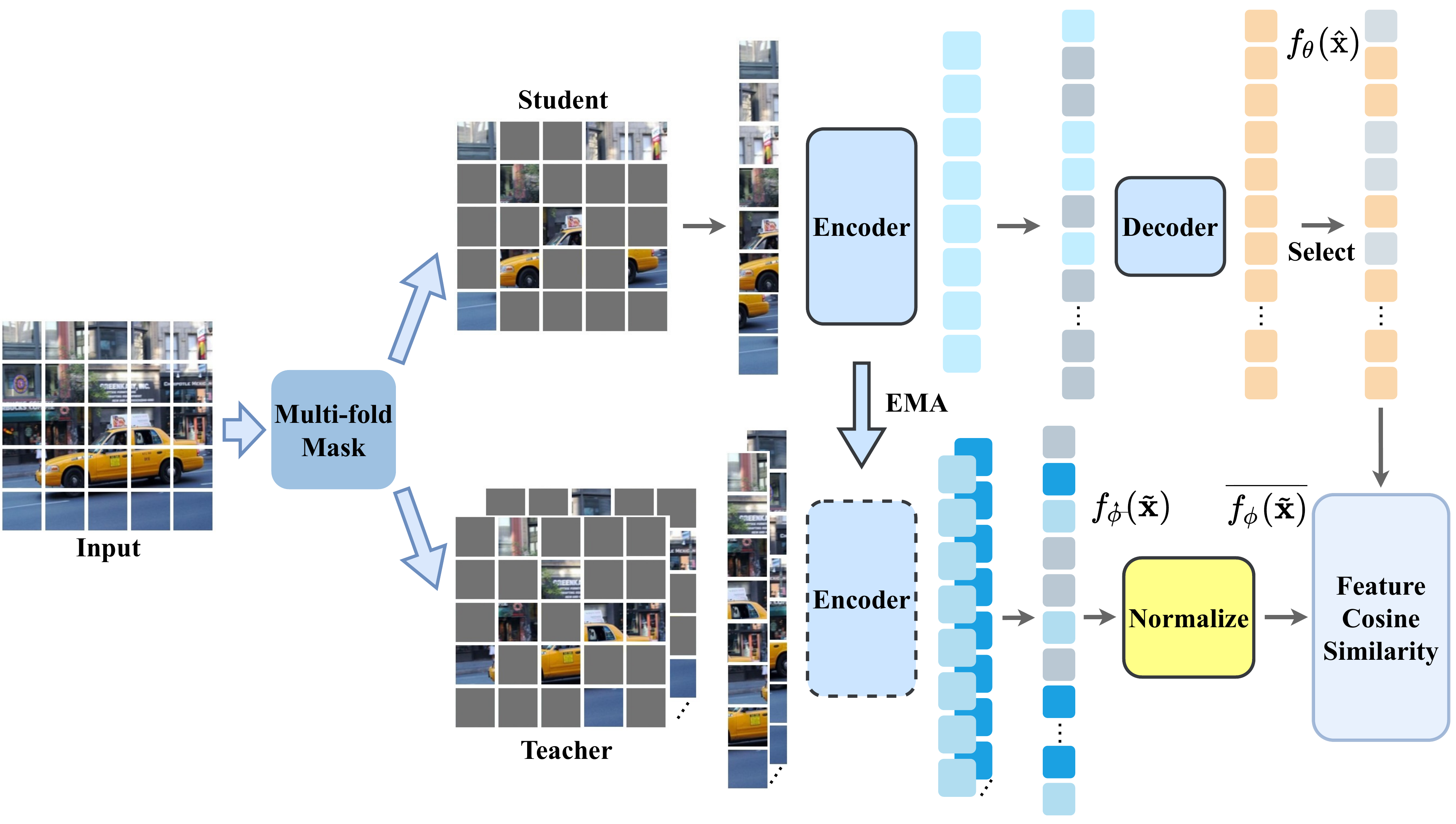}
    \caption{SdAE pipeline, courtesy of \cite{chen-eccv-2022} (image licensed under CC BY 4.0). SdAE is classified as a contrastive-based approach (belonging to the generic pipeline in Figure \ref{contrastive}), as it uses the cosine distance between the student and online teacher features, which are extracted from distinct masked variants of the same input image.}
    \label{fig:sdae}
\end{figure*}

\cite{chen-eccv-2022} argued that reconstructing the image pixel space does not force the network to learn an ideal representation of the data. Thus, they introduced a self-distillation framework, shown in Figure~\ref{fig:sdae}, in which the student branch follows the original MAE flow, while the teacher network (which is not updated by gradient descent, but rather from the weights of the student) only takes the masked patches. The objective is to match the high-level features between the two networks. A similar approach is proposed by 
 \cite{assran-eccv-2022masked}. The authors presented a self-supervised masking pre-training strategy that involves Siamese networks. Given a source image and applying transformations on it, two different views (anchor and target) are generated. Then, only the anchor's tokenized patches are masked, and using the Siamese networks (which follow the ViT architecture), both views are encoded. The latent representations are compared with a set of learnable prototypes in order to generate a distribution, the final goal being that of obtaining matching distributions (predictions of anchor and target). While the anchor's model is updated using gradient descent, the target's network parameters are computed as an exponential moving average from the anchor network's weights. The authors attest that the method greatly improves the performance in few-shot scenarios.

A handful of studies used 3D data as target features within their proposed MIM pipeline.
\cite{liang-eccv-2022a} proposed the masked image modeling pre-training for 3D meshes. Inspired by ViT, they begin by grouping faces together (each face being represented by a 10D vector) into a patch. Then, the authors adopt the transformer architecture, using the 3D location of the center of each patch to compute the positional embedding. The algorithm follows MAE: patches are randomly masked with a high masking ratio, the visible patches are passed through the encoder, then the resulting latent embeddings are concatenated with masked tokens (associated with the masked patches), and subsequently decoded. The objective is not only to reconstruct the masked patches (by predicting the coordinates of the vertices), but also the faces of the patch (the 10D representations). Aiming to learn more robust representations with 2D vision models, Mask3D~\citep{hou-cvpr-2023} enhanced the 2D feature representations of the ViT backbone \citep{dosovitskiy-arXiv-2020} by integrating 3D priors into the training pipeline. Mask3D utilized RGB-D data within a self-supervised framework, where both color images and corresponding dense depth maps are masked and processed through dual encoders. These encoders project the data into a higher-dimensional space, enabling a decoder to accurately reconstruct the dense depth map. This method enriches the capability of ViT to handle spatial depth alongside traditional 2D data. Focusing on point clouds, 
GeoMAE~\citep{tian-cvpr-2023} is an adaptation of the MAE framework to point clouds. This method changed the usual reconstruction objective and replaced it with centroid prediction, normal estimation and curvature prediction. This change showed significant improvements in downstream tasks such as object detection, segmentation and multi-object tracking. 
 In a semi-supervised setting, 
given a labeled (source) and an unlabeled (target) 3D point cloud dataset, the aim of \cite{liang-eccv-2022b} is to transfer the knowledge from the latter to the former by embedding information about common features in an encoder. This model is trained simultaneously on both datasets, but with different objectives. On the one hand, training on the source dataset is performed in a supervised setting on a specific task. On the other hand, points from the target dataset are randomly masked in arbitrary areas and the objective is to estimate their cardinality (number of points in the neighborhood), position and normal vectors. Therefore, the model benefits from unlabeled data by encapsulating information about the structure of objects.

To boost the performance of any knowledge distillation framework based on feature learning, \cite{yang-eccv-2022} proposed an auxiliary task. A proportion of the latent feature maps of the student is masked, and a projection layer tries to recover the masked feature maps and match them with those of the teacher. This results in higher performance for a wide range of computer vision tasks. In a similar manner, but focusing on efficiency, 
\cite{bai-cvpr-2023} introduced a method for knowledge distillation using the MAE framework. Their approach reconstructs masked patches, while training the student model to replicate the early (low-level) feature maps of the teacher. The efficiency stems from utilizing the MAE framework and from the partial evaluation of the teacher network, as it only requires early feature maps for training the student. This significantly reduces computational cost and training time.

\cite{wei-cvpr-2022} used the HOG features of the masked regions as target values. The masked patches are replaced with learnable tokens and the architecture is based on a single encoder followed by a linear head for predicting the HOG features. Low-level target features were also used by \cite{ren-cvpr-2023}. 
 They introduced a novel approach for boosting the performance of ViT. A number of patches from the original image are shuffled and have their positional encodings masked. Besides the original loss of the downstream task, another objective that tries to predict the masked positional encodings is employed.


A few studies have attempted to employ target features better suited for video data.
\cite{sun-cvpr-2023} proposed a video representation learning method based on masking, which reconstructs trajectory features rather than static information (like frames). These target features are carefully designed to capture long-term motion changes (object shape and position). 
\cite{fu-cvpr-2023b} conducted an empirical analysis on optimal target features for video-language pre-training. They found that spatially-focused image features, extracted using a Swin-B transformer \citep{Liu-ICCV-2021}, yield the best results. Additionally, their research incorporates the usual tasks employed in video-language pre-training, such as masked video modeling, masked language modeling and video-text matching. 
 For their video compression method, \cite{xiang-iclr-2023} employed a transformer-based entropy that is pre-trained using masked image modeling. Some tokens from the current frame are randomly masked, and the model tries to estimate their probability mass functions. More prior information about the last decoded frames is supplied as keys and values. At inference, the prediction is performed as an iterative process.
 
\cite{yang-cvpr-2023} combined image-text contrastive learning (CLIP) \citep{Radford-ICML-2021} and MIM. Their main contribution over the naive approach for combining the two methods consists in using the language space as target space for the reconstruction objective. This is motivated by the intuition that the language features serve as rich semantic representations of the visual signal. The CLIP representation space is also utilized by
EVA~\citep{fang-cvpr-2023}, a ViT model trained to reconstruct CLIP features conditioned on masked image tokens. \cite{fang-cvpr-2023} showed that this self-supervised task is suitable for large scale representation learning. The EVA model scales to one billion parameters using tens of millions of samples, showcasing its potential for handling extensive datasets and complex learning tasks.

\cite{zhao-miccai-2023} introduced a masked-based auxiliary objective to train a model for semantic segmentation of Laparoscopic images. Due to the scarcity of labeled data, the authors proposed to use a labeled proxy source dataset (with simulated images) and an unlabeled target dataset (with real images) to transfer the knowledge. The former dataset was used to compute the supervised loss for segmentation with a student model. Each image from the second dataset has its higher frequencies masked (by first applying the Fourier transform and then the inverse), and its segmentation map is predicted using the student model. The resulting output is compared with the prediction of the intact image given by a teacher model (i.e.~the exponential moving average of the student), the objective being that of minimizing the distance between the two.

\subsection{Objective Function}
\begin{figure*}[t]
    \centering
    \includegraphics[width=0.88\linewidth]{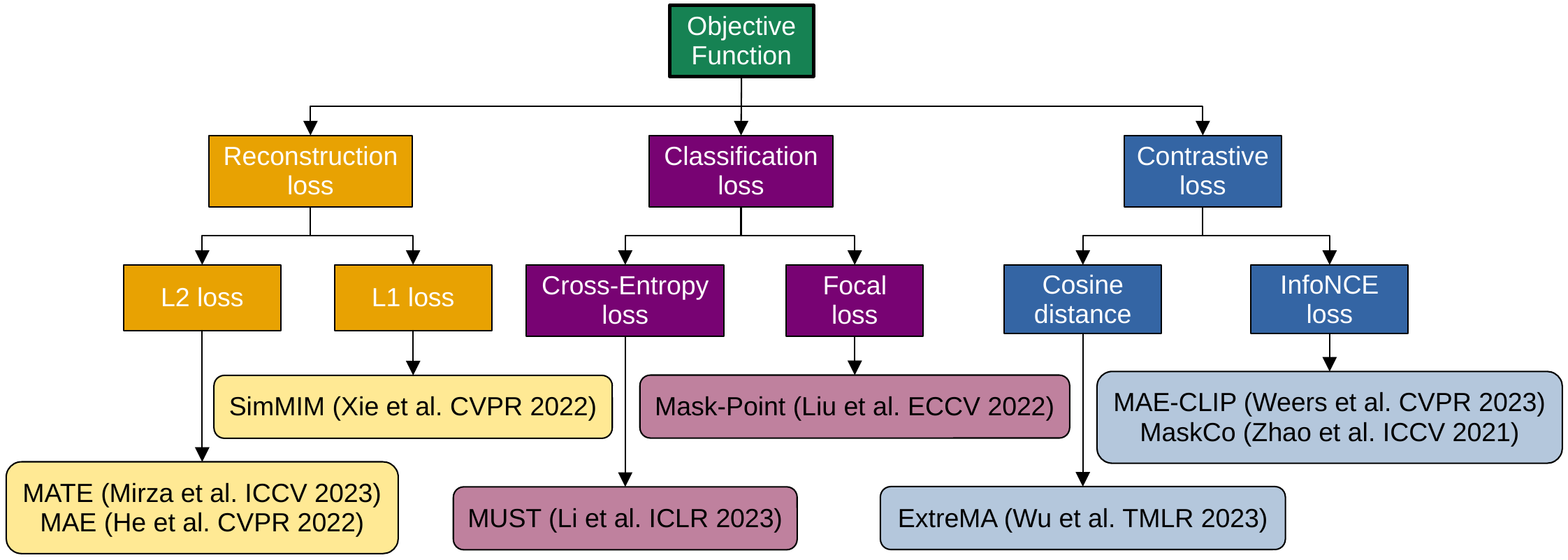}
    \caption{Types of objective functions employed in MIM pipelines. Reconstruction losses, particularly those based on $L^1$ and $L^2$ distances, constitute the most commonly employed objective functions in MIM pipelines. These are followed by the contrastive loss objectives, where the most used instance is the InfoNCE loss. Classification losses are less frequent, but they are present in pipelines that employ pseudo-labels or conduct masking within a discrete space.}
    \label{fig:objective_functions}
\end{figure*}

We next present the works that have as a main contribution the loss function employed in their MIM pipeline. In Figure~\ref{fig:objective_functions}, we provide an overview of the main objective functions used across MIM approaches, distinguishing between reconstruction~\citep{he-cvpr-2022, xie-cvpr-2022, mirza-iccv-2023, huang-tpami-2023}, contrastive~\citep{weers-cvpr-2023, lee-iclr-2023, wu-TMLR-2023, huang-tpami-2023}, and classification losses~\citep{liu-eccv-2022, li-iclr-2023}. Reconstruction losses remain the most frequent option in MIM pipelines. In this section, we highlight methods that employed them in less typical ways. For instance \cite{mirza-iccv-2023} used a reconstruction loss at test time, while \cite{huang-tpami-2023} combined reconstruction and contrastive losses. Although classification losses are less common, they appear in scenarios where pseudo-labels can be generated~\citep{liu-eccv-2022, li-iclr-2023}.

Inspired by MAE, \cite{liu-eccv-2022} proposed an analogous pre-training framework for 3D point clouds, but substituted the reconstruction task with discrimination. A small proportion of points are left unmasked, and subsequently encoded. Then, a subset of the masked ones is sampled (real), along with some random 3D points from the space (fake), and the decoder's objective is to discriminate between the two. The encoded unmasked points are used in the cross-attention blocks of the decoder. Experiments are conducted for various downstream tasks, in which the method shows considerable performance improvements. A classification loss is also used by \cite{li-iclr-2023}. 
To boost the performance of models that have zero-shot classifying capabilities (such as CLIP), \cite{li-iclr-2023} proposed a framework composed of three tasks. Besides the reconstruction loss of the masked patches, the second objective is to minimize the distance between the resulting embeddings of the masked tokens and the embedding of the prepended \emph{[CLS]} token in a shared projected space. The third objective employs a classification loss in which the labels are provided by an EMA teacher applied on the unmasked image.

\cite{weers-cvpr-2023} studied the effectiveness of combining MAE and CLIP in a single framework. The conclusion is that the combination brings some benefit when training on a smaller dataset (tens of millions of samples), but this benefit is marginal when the experiments are carried out on a much larger dataset (1 billion samples).

\cite{mirza-iccv-2023} introduced an application of the MAE self-supervised learning framework during test phases, followed by executing predictions with the refined weights. This method was evaluated in the context of point cloud classification, demonstrating enhanced performance across a variety of standard perturbations affecting 3D point clouds. The findings suggest that updating model weights with MAE at test time significantly improves the robustness and accuracy of point cloud classification tasks.

\begin{figure*}[t]
    \centering
    \includegraphics[width=0.92\linewidth]{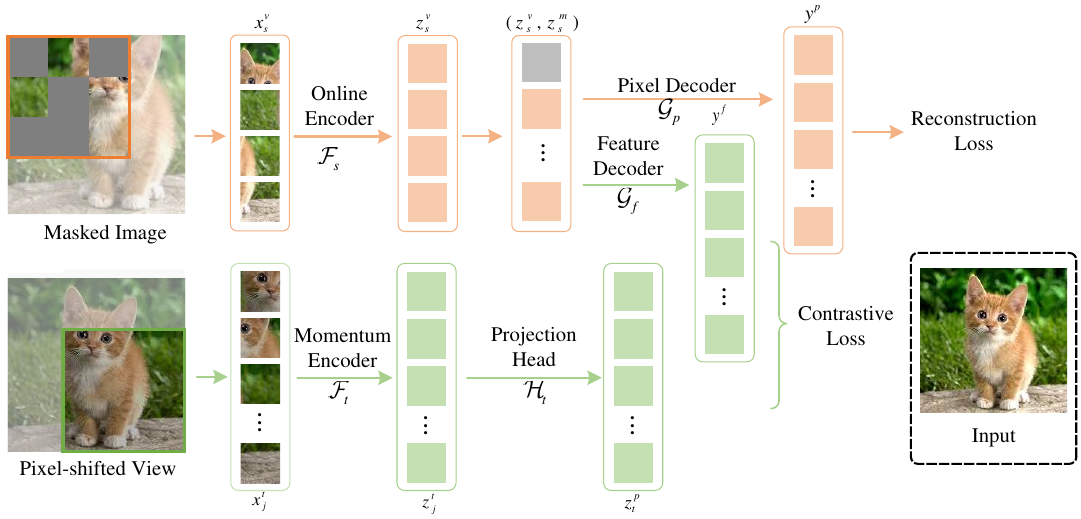}
    \caption{CMAE pipeline, courtesy of \cite{huang-tpami-2023} (image licensed under CC BY 4.0). CMAE is a hybrid of the generic pipelines illustrated in Figures \ref{reconstruction} and \ref{contrastive}, combining a pixel-level reconstruction loss with a contrastive loss. The latter ensures alignment between features extracted from a masked image and those derived from a pixel-shifted version of the same image, extracted with a momentum encoder.}
    \label{fig:cmae}
\end{figure*}

\cite{lee-iclr-2023} formally introduced the self-distillation masked autoencoder framework. An input image is divided into patches, some of which are randomly masked. Two encoder-decoder networks (teacher and student) are used to reconstruct the original image. A new objective is employed to minimize the distance between the predictions of the teacher and the student. While the student is trained using gradient descent, the teacher's weights are computed as an exponential moving average of the student's. \cite{wu-TMLR-2023} proposed a similar method based on Siamese networks. One network, which is fed with a highly masked input image, tries to predict the tokens from the other network, which is given the complete image. 
The work of \cite{huang-tpami-2023} is built on the same idea, but combines the reconstruction objective with a contrastive loss.  As presented in Figure~\ref{fig:cmae}, the authors achieve this by using two branches, one for each strategy. The first one uses an encoder and a pixel decoder, which are updated at every step. The second one employs a distinct encoder, which is updated as an exponential moving average from the other encoder. It also uses a projection layer and a feature decoder that have the same output vector space. Two views that have a different shift are created from the input image, one being passed through the first branch (as well as being masked), while the other is fed into the second branch (unaltered). The first branch follows the original MAE framework, with the reconstruction objective. The second branch applies a contrastive loss between the projected embedded features (with the latter encoder) of the second view and the decoded embedded features (i.e.~from the former encoder) of the first view. A similar hybrid approach is used by \cite{hernandez-eccv-2024}. The authors apply a reconstruction loss to masked images and a contrastive loss to masked video frames, enabling their method to be effective in both image and video tasks.

\cite{lin-neurips-2023} integrated a masked autoencoder model into a reinforcement learning setting in order to obtain a reward model for exploration. The autoencoder is trained by masking some states from a trajectory and then estimating them. Given a sampled trajectory, for each timestamp, a fixed number of previous states are kept and encoded. Some of the resulting embeddings are then masked, while the rest are passed through the decoder to predict the missing states. In the end, the exploration reward is given by the prediction error.


\subsection{Downstream Task}
\begin{figure*}[t]
    \centering
    \includegraphics[width=1.0\linewidth]{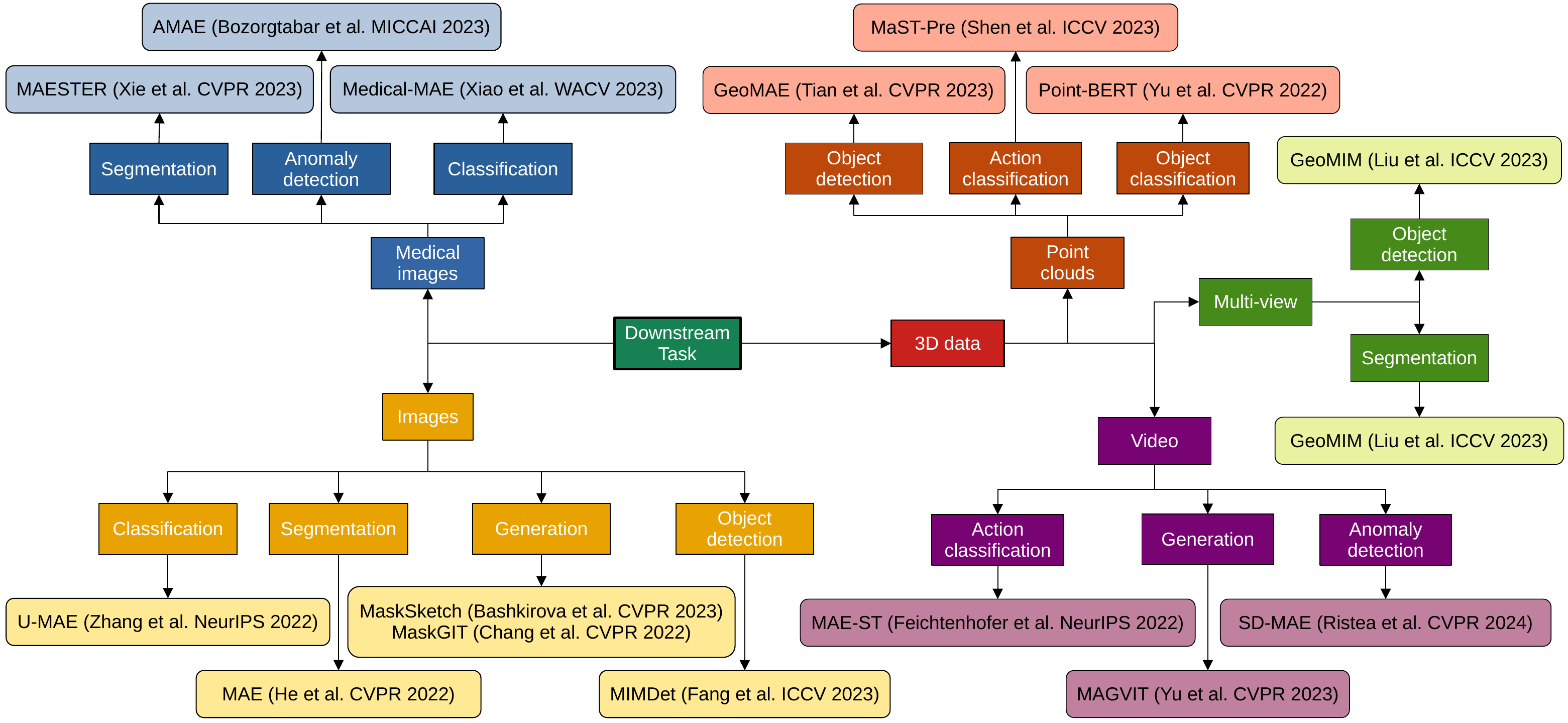}
    \caption{Downstream tasks on which MIM pipelines are typically applied. The downstream tasks can be grouped by the input modality. Most MIM research focuses on natural images, evaluating the learned representations through downstream tasks such as classification, segmentation, generation, and object detection. Similar downstream tasks are used for medical images, multi-view images and point clouds. In the case of videos, the primary task is action classification, but there are also some studies that investigate the effectiveness of MIM in video generation and anomaly detection.}
    \label{fig:downstream_task}
\end{figure*}

In Figure~\ref{fig:downstream_task}, we outline the downstream tasks where MIM pipelines have been successfully applied, grouping them by modality. In the case of images, pre-training strategies are typically evaluated on classification~\citep{he-cvpr-2022, xie-cvpr-2022, gandelsman-neurips-2022}, object detection~\citep{fang-iccv-2023}, segmentation~\citep{he-cvpr-2022}, and generation~\citep{chang-cvpr-2022, chen-cvpr-2023c}. For medical images, segmentation~\citep{chen-wacv-2023} and classification~\citep{xiao-wacv-2023} remain prevalent, with anomaly detection~\citep{bozorgtabar-miccai-2023} also being investigated. When it comes to videos, MIM pipelines are employed for generation~\citep{yu-cvpr-2023}, action classification~\citep{feichtenhofer-neurips-2022}, and abnormal event detection~\citep{ristea-cvpr-2024}. In multi-view images, object detection~\citep{liu-iccv-2023} and segmentation~\citep{liu-iccv-2023} are common applications, while for point clouds, researchers have explored object detection~\citep{tian-cvpr-2023}, classification~\citep{yu-cvpr-2022}, and action classification~\citep{shen-iccv-2023}.

MaskGIT \citep{chang-cvpr-2022} trains a generative transformer to reconstruct randomly masked image patches. At inference time, it starts by predicting all the patches and keeps the most confident ones for the next iteration, when the rest of the patches are again masked and regenerated. The process continues for a few iterations. Overall, the pipeline has actually two stages, the first one encodes the patches into visual tokens with a VQ-Encoder, and the second stage (decoder) receives masked tokens for reconstruction. Inspired by this, \cite{Tschannen-ECCV-2024} adopted the token masking strategy in their work, but replaced the values with zero (rather than a special token) and the token embeddings are obtained with a single dense layer. Another work that studies the application of MIM in image synthesis is MDT~\citep{Gao-ICCV-2023}. In their work, \cite{Gao-ICCV-2023} integrated latent masking in the training pipeline of a latent diffusion model.

\cite{feichtenhofer-neurips-2022} proposed to use MAE on videos. The video is split into equally-sized patches that do not overlap along any dimension (i.e.~including time), as well as consisting only of two timesteps. As hypothesized and demonstrated by the authors, a high masking ratio (about 90\% of the whole input video, unaware of any axis) is used due to redundant data when decoding. Positional (i.e.~height and width) and temporal (i.e.~time) embeddings are added to the input tokens. The architectures of both the encoder and the decoder are based on ViT. The method follows the same logic as MAE: encoding only the visible tokens, then decoding the complete set of tokens. Targeting video generation, 
MAGVIT~\citep{yu-cvpr-2023} trained a transformer-based model  for conditional video generation, handling tasks like frame prediction, interpolation, and central outpainting. The model processes videos by selecting a task, and creating conditional tokens from the preprocessed video. The conditional tokens, along with masked and original tokens, form the input used to train the model, which is optimized with three objectives: refining conditional tokens, predicting masked tokens, and reconstructing original tokens. Inference is conducted autoregressively, reducing the masking ratio progressively.

\begin{figure*}[t]
    \centering
    \includegraphics[width=0.8\linewidth]{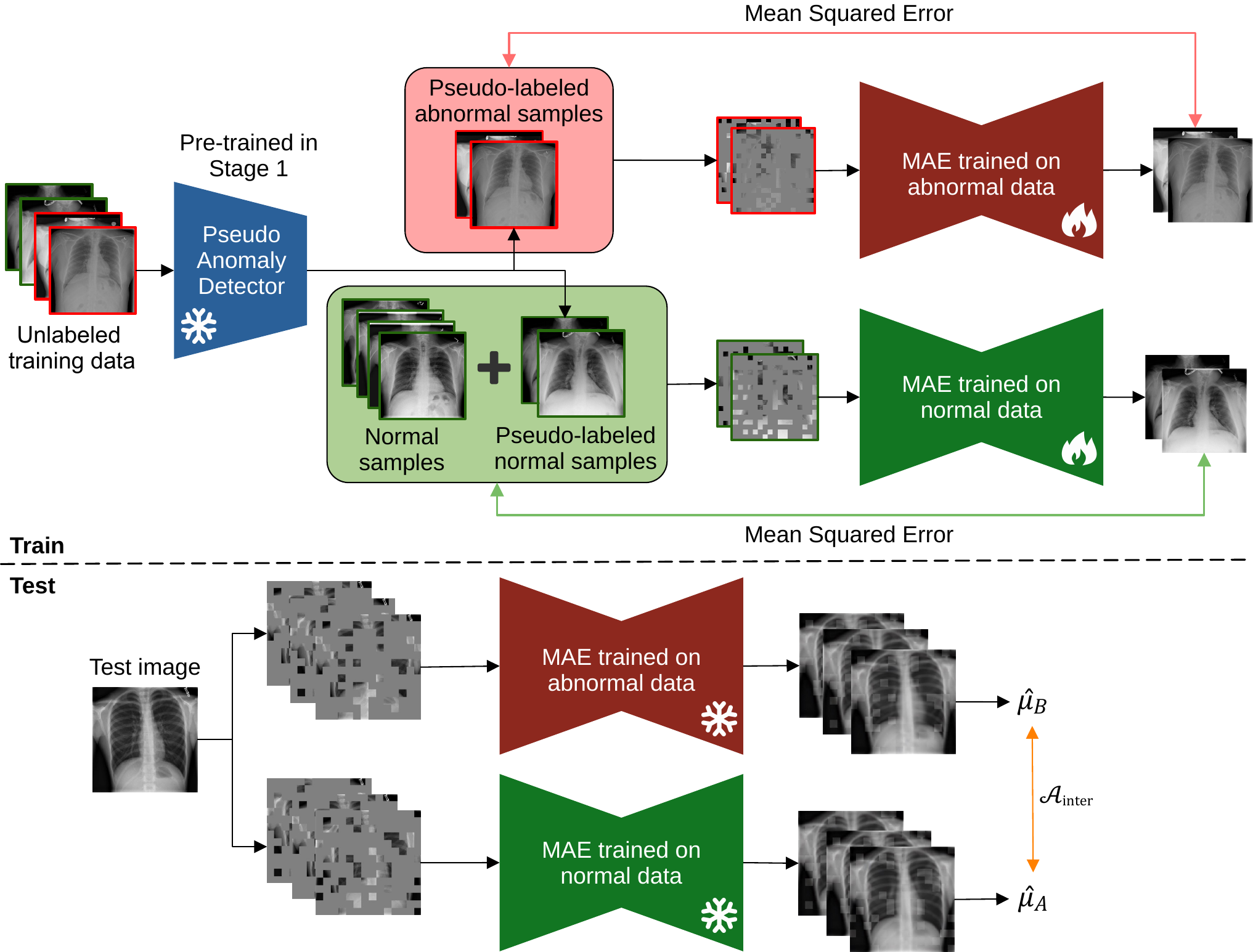}
    \caption{AMAE \citep{bozorgtabar-miccai-2023} belongs to the family of reconstruction-based methods (it falls under the generic pipeline presented in Figure \ref{reconstruction}), as it employs an ${L}^2$ loss for masked patch reconstruction, similar to MAE.}
    \label{fig:amae_step2}
\end{figure*}

\cite{gandelsman-neurips-2022} adapted MAE for test-time training. Employing a pre-trained MAE whose weights are frozen, a classification head, represented by a ViT, is attached to the encoder and fined-tuned on the supervised dataset. At inference time, each sample is initially used to train the network on the reconstruction objective in multiple steps, thus modifying the resulting encoded latent representation, while the head does not change, and then classifying the input. The encoder and the decoder are always reset with their original weights after each prediction. An increase in performance comes at the cost of inference speed.



\cite{yan-eccv-2022} applied a pre-training masking strategy to the less studied task of panoramic depth completion. A pair consisting of an RGB image and the associated sparse depth map are jointly masked. Both are then passed through a guided convolutional network \citep{tang-tip-2020} to generate the sparse depth map, and the reconstruction loss is only computed for the masked depths. Carrying out experiments on the dense depth map prediction downstream task, the proposed method, called \textit{$M^3PT$}, shows better qualitative and quantitative results than previous state of the art.


Several works have focused on various tasks in the medical domain.
Masked Autoencoder Guided Segmentation at Pixel Resolution (MAESTER)~\citep{xie-cvpr-2023a} is a masked image modeling approach for the segmentation of cellular images. MAESTER incorporates the masked autoencoder in the training pipeline of a visual transformer to learn token representations relevant to the sub-cellular structure segmentation task (i.e.~texture) and performs the segmentation at inference time by clustering the learned representations.
\cite{chen-cvpr-2023c} used masked image modeling to learn a latent space from fMRI inputs. After this stage, the authors used the learned latent representations to condition a diffusion model that is able to generate visualizations of the initial visual stimuli. 
\cite{bozorgtabar-miccai-2023} leveraged an MAE for detecting anomalies in chest X-rays. The first stage of their pipeline 
consists of initially pre-training a masked autoencoder. A classification head is then attached to the encoder (whose weights are frozen) to classify normal and abnormal samples, the latter being artificially created. During the second stage, unlabeled data is classified using the model from the previous stage, and the examples that have high confidence are kept. The last step is to employ two different copies of the pre-trained autoencoder, one for each class, and train them separately on the masked reconstruction task. At inference, multiple reconstructions are generated from both autoencoders, an anomalous prediction being detected by a large difference between the mean reconstructed images of the two modules. The steps in the second stage are presented in Figure~\ref{fig:amae_step2}. 
In an effort to boost the performance of computer vision models in the medical domain, \cite{xiao-wacv-2023} used visual transformers for multi-label classification of chest X-rays. The large quantity of data required for training a ViT was overcome by pre-training the model on the masked modeling task. Furthermore, the authors proved the superior performance of their method compared with CNN-based architectures.
Similar to \cite{xiao-wacv-2023}, \cite{chen-wacv-2023} demonstrated the benefits of masked image modeling on 3D medical images. They adopted two strategies (original MAE and SimMIM) with various masking hyperparameters, showcasing improved results on two different segmentation tasks. 
\cite{kraus-cvpr-2024} adopted the MAE pre-training strategy for microscopy images. Their objective was to learn a rich feature representation of the cellular morphology of the data. The extensive experiments demonstrated excellent results for predicting biological relationships. Leveraging both images and text, 
 \cite{zhou-iclr-2023} introduced masked modeling as a pre-training strategy for medical radiography tasks by using the associated radiology reports. While the images are downsampled by half and their unmasked patches are encoded, the text reports are tokenized, randomly masked and then embedded through a look-up table. A global average pooling over the resulting visual embeddings is added to the unmasked text embeddings, and all of them are decoded to obtain the intact report.  The image tokens are fed as well through another decoder in order to reconstruct the radiography at its original resolution.

\cite{fang-iccv-2023} explored the effectiveness of a pre-trained Vision Transformer (ViT) on masked image modeling tasks within the context of object detection.
Geometry Enhanced Masked Image Modeling (GeoMIM)~\citep{liu-iccv-2023} tackled the problem of 3D detection using multi-view cameras. This method involves pre-training a student network by utilizing masked inputs from multiple camera views. The objective is for the student network to reconstruct the bird's-eye-view features, leveraging the guidance of a pre-trained LiDAR-based model. This strategy bridges the gap between multi-view camera inputs and LiDAR precision, enhancing the student network's ability to accurately interpret and reconstruct 3D environments.

 In order to combine masked image modeling with contrastive learning during pre-training, \cite{jiang-iclr-2023} proposed to apply the former on the initial layers, while the latter is used on the last layers. This is achieved in an iterative manner, by firstly pre-training on the reconstruction task, freezing the respective layers, and then training on the contrastive task. 
 

\cite{Pham-ICML-2024} integrated a secondary masked image modeling task in their pose-guided human image generation training pipeline. The authors opted for a transformer-based denoising architecture, rather than a U-Net. As a result, the masked tokens of the target image are reconstructed not only through the self-attention layers, but also using cross-attention modules with the aggregated tokens of the target pose and the source image.

\subsection{Theoretical Analysis}

Besides the applied contributions of MIM, some pieces of work take another approach: they theorize about various aspects of MIM and dive deeper into its fundamentals. The papers from this category address aspects such as overall understanding \citep{kong-cvpr-2023a, pan-iclr-2023, xie-cvpr-2023c, liu-neurips-2022, li-cvpr-2023b, moreno-neurips-2023}, the connection between different MIM strategies \citep{zhang-neurips-2022}, or present certain drawbacks and how to overcome them \citep{huang-iccv-2023a, xie-cvpr-2023b}. 


\cite{liu-neurips-2022} evaluated the performance of self-supervised learning (SSL) by determining whether the trained model represents enough information to obtain the data distribution, given additional information about the distribution family. The SSL task chosen for this assessment was the masked prediction task.

\cite{li-cvpr-2023b} explored the effectiveness of MIM on out-of-distribution (OOD) detection. Their results showed that MIM improves the performance in multiple settings, such as one-class, multi-class or near-distribution OOD.

\begin{figure*}
    \centering
    \includegraphics[width=0.92\linewidth]{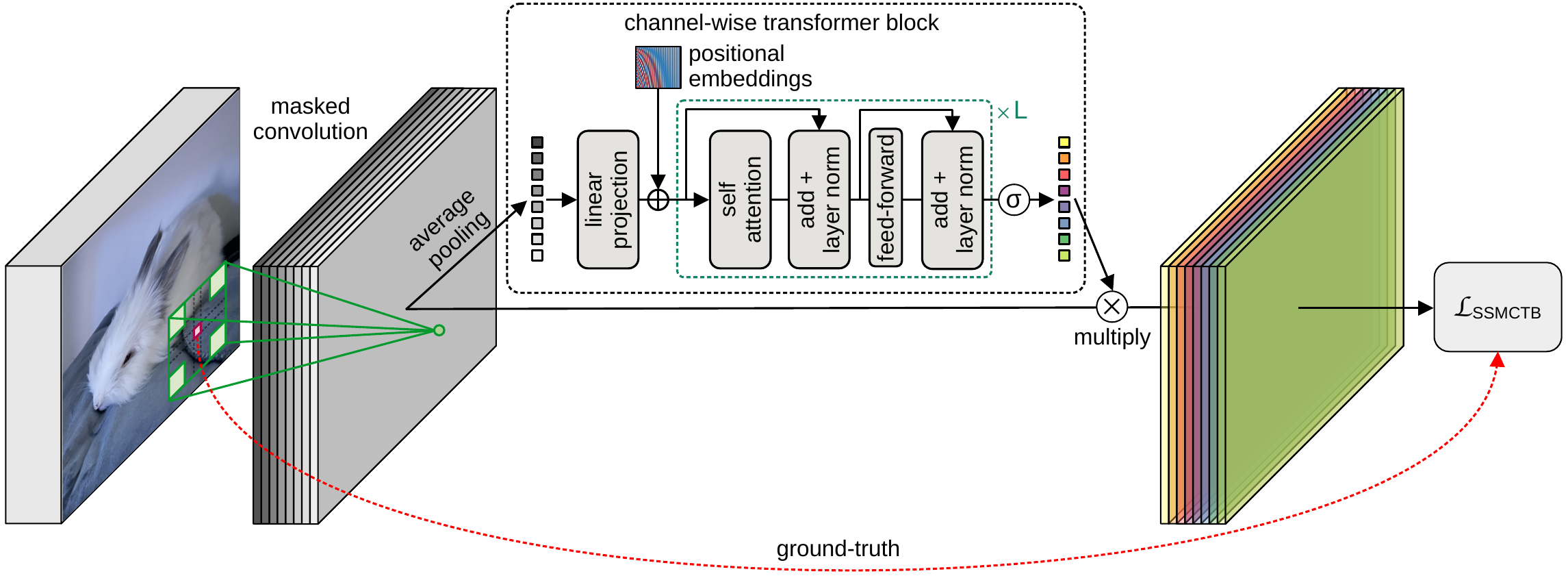}
    \caption{SSMCTB architecture, courtesy of \cite{madan-tpami-2024} (image available under the ArXiv perpetual license agreement). SSMCTB falls into the category of reconstruction-based methods depicted in Figure \ref{reconstruction}, as it employs the Huber loss between the reconstructed output and the input.}
    \label{fig:ssmctb}
\end{figure*}

\cite{kong-cvpr-2023a} formulated the underlying data generation process as a hierarchical latent variable process.
The authors discovered relationships between the latent variables of the data generation process and the masking parameters (masking ratio and patch size) of the MAE framework. These relationships allow MAEs to recover variables of different semantic levels. 
The authors validated their theoretical discoveries with several experiments, where the main result showed that very large masking ratios have a similar effect as low ratios, namely that of learning low-level information.

\cite{xie-cvpr-2023b} investigated the data scaling capabilities of masked image modeling. Their experiments use datasets of various sizes, ranging from 100.000 samples to 14 million samples. The authors verified their observations against two masked image modeling approaches, SimMIM~\citep{xie-cvpr-2022} and MAE~\citep{he-cvpr-2022}. The conclusions of their study state that MIM still necessitates data scaling to effectively facilitate model scaling. The study also noted that, in non-overfitting scenarios, simply increasing the number of unique samples does not necessarily enhance performance. 

\cite{xie-cvpr-2023c} explored the differences in representations learned by deep models through MIM versus traditional supervised training. They observed that MIM encourages models to focus on local patterns across all layers, whereas supervised training emphasizes these patterns only in the initial layers. Additionally, MIM results in a greater diversity among the attention heads compared with supervised methods, suggesting a more nuanced feature recognition within the model. In addition, 
\cite{huang-iccv-2023a} studied the adversarial robustness of the transformers pre-trained with MIM. Their first observation is that MAE, in particular, has a lower robustness compared with other methods. Moreover, they found that the robustness is related to the reconstruction target. For example, a model trained to reproduce the pixels of an image is prone to adversarial attacks because its focus is on medium and high-frequency features. To ameliorate this issue, the authors proposed a test-time solution based on visual prompts optimized on the frequency domain. These prompts are then included in the input images through prototype-based prompt selection.

\cite{pan-iclr-2023} theorized about the mechanisms behind reconstructing the masked input, the benefits of this pre-training strategy and why it learns valuable feature representations. The main finding is that discriminative features are learned during pre-training, and thus, when applied to a downstream task, these are further enhanced, which has a great advantage over randomly initialized weights.


\begin{figure*}
    \centering
    \includegraphics[width=0.94\linewidth]{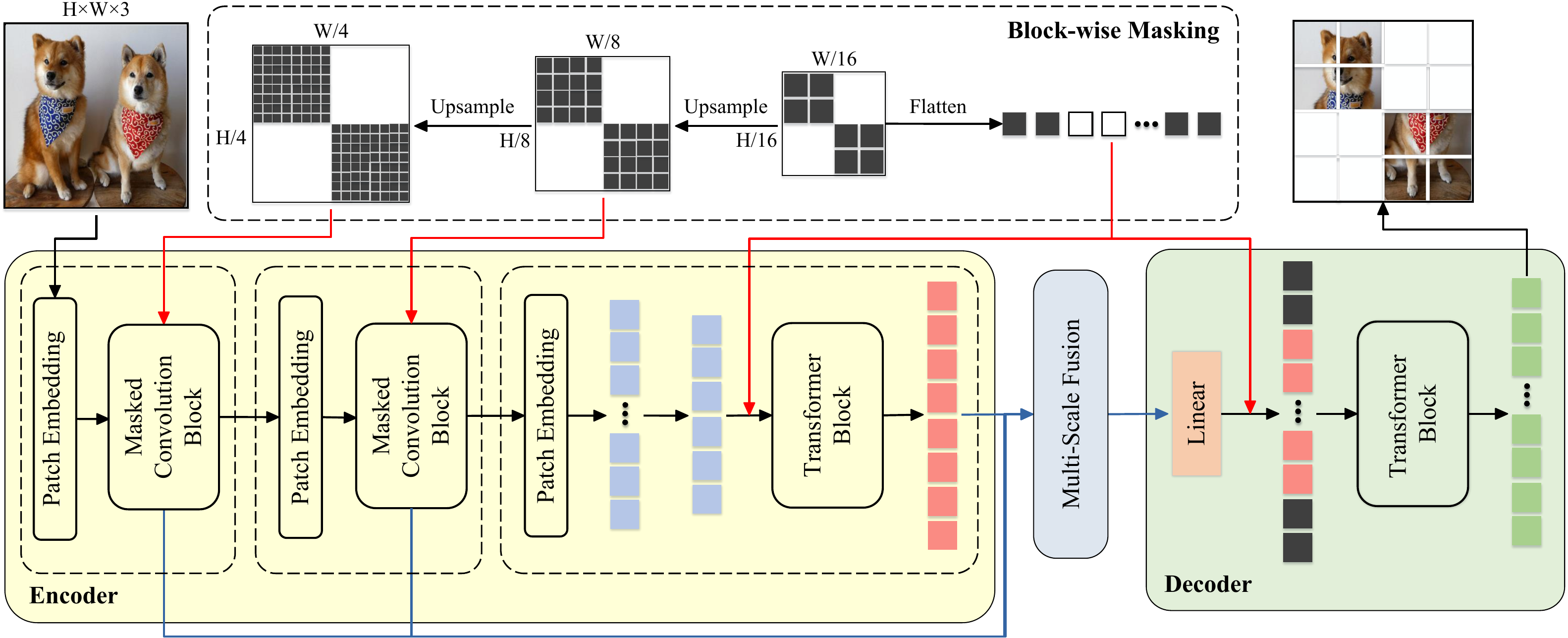}
    \caption{MCMAE~\citep{gao-neurips-2022} falls into the category of reconstruction-based methods illustrated in Figure \ref{reconstruction}, as it leverages an ${L}^2$ loss between the masked patches and the corresponding patches from the input image.}
    \label{fig:mcmae}
\end{figure*}

\subsection{Model Architecture}

While the architecture employed throughout MIM research was consistent (a transformer-based encoder and a shallow decoder), there have been some important contributions that further enhance the performance of the pre-training task through architectural modifications \citep{liu-iclr-2023}. Some of these advances increase the benefit of MIM pre-training on compact transformers~\citep{zhou-iccv-2023, liu-iclr-2023}, while others integrate multiple modalities within the architecture~\citep{mo-cvpr-2024}. Furthermore, a few attempts have tried to distinguish themselves from the usual ViT-based models, either by using CNNs \citep{tian-iclr-2023, woo-cvpr-2023, gao-neurips-2022} or by integrating MIM into the convolution operation \citep{madan-tpami-2024, ristea-cvpr-2022}.


\cite{ristea-cvpr-2022} presented the self-supervised predictive convolutional 
attentive block (SSPCAB), a novel block comprising a masked convolutional layer and a Squeeze-and-Excitation (SE) module. The filters of the masked convolutional layer contain learnable parameters in the corner regions of the receptive field and the masked region is located in the center. This novel block is trained using a self-supervised reconstruction loss, being integrated in anomaly detection networks. Later, \cite{madan-tpami-2024} introduced the self-supervised masked convolutional transformer block (SSMCTB) for anomaly detection. SSMCTB is an extension of SSPCAB, being trained via a self-supervised reconstruction loss and comprising a masked convolutional layer. In contrast to \cite{ristea-cvpr-2022}, \cite{madan-tpami-2024} employed a channel-wise transformer block instead of the SE module, as shown in Figure \ref{fig:ssmctb}.

Motivated by leveraging the masked pre-training strategy of autoencoders with convolutional layers, \cite{gao-neurips-2022} presented an architecture that combines them, as illustrated in Figure~\ref{fig:mcmae}. The encoder consists of three stages, where the first two are convolutional and the last one is transformer-based. First, a mask is sampled to determine which tokens are visible. The mask is then upsampled at the resolutions of the first two stages, in order to be used by masked convolutional blocks. Information from the first two stages is added to the resulting tokens of the third stage, and then, they are linearly projected. Finally, all tokens (predicted and masked) are decoded into the original pixel space. The authors state that the advantage of this method is represented by the multi-scale features learned by the encoder. Having the same goal of adopting convolutional layers in MIM pipelines, 
\cite{tian-iclr-2023} presented a MIM method that leverages CNNs. As in the MAE framework, the input is split into non-overlapping patches and some of them are masked. The encoder is composed of sparse convolutional layers, thus preserving the mask pattern intact along the feature maps. The decoder, used for reconstructing the original image, is composed of upsampling blocks that receive the previous layer, as well as the encoder's features on the same level. The masking regions are replaced with a mask embedding.

SparseMAE~\citep{zhou-iccv-2023} offered a novel solution to the problem that small transformers face in not benefiting significantly from MAE pre-training. It did this by concurrently training a full-scale transformer alongside a smaller sparse network, which resides inside the full transformer. This smaller network is tasked with reconstructing the masked patches of input images. Uniquely, SparseMAE independently manages two sets of weights, one for the sparse network and another for the encompassing larger transformer. Despite this separation, both networks aim to achieve the same objective: the accurate reconstruction of the masked image patches. After pre-training, only the sparse network is used for fine-tuning. A similar approach based on smaller networks is introduced by \cite{liu-iclr-2023}. 
Inspired by TokenMoE \citep{riquelme-neurips-2021}, \cite{liu-iclr-2023} proposed a pre-training method that is more robust for all downstream tasks. The first step is to obtain the features representations of a pre-trained MAE and cluster them (obtaining the centroids). The architecture of the model is adapted to contain multiple experts (i.e.~groups of heads in the transformer layers), each expert being associated with a cluster. The tokens are routed through the expert which the image was assigned to. When applied to a downstream task, the method selects only the most used experts by the dataset for fine-tuning.

\cite{han-neurips-2023} proposed to integrate RevCol \citep{cai-arxiv-2022} in the MAE framework. Besides the bottom-up columns, RevCol is extended to include top-down columns as well, thus resembling an encoder-decoder. The network contains reversible connections, helping the model to learn disentangled representations. In this way, the decoder is no longer dropped during fine-tuning, as it contains salient information.

\cite{mo-cvpr-2024} proposed to combine audio and visual modalities at the earliest stage in a multimodal architecture. Aside from separate modality transformer blocks, another block is used to process both the audio and visual tokens, further fusing the resulting information into the other blocks. The experiments demonstrated strong performance on a wide range of downstream tasks, such as visually-guided sound source separation, audio-visual segmentation or classification.

\subsection{Features and Objective}

We next discuss studies that make contributions to both the target features and the objective function. While most approaches use a deep encoder, either frozen \citep{huang-cvpr-2023, wu-miccai-2023} or updated online \citep{zhao-iccv-2021, zhu-tpami-2022, dong-cvpr-2023, kong-cvpr-2023b, chen-miccai-2023} during training, to extract the target features, there are also methods that extract their features directly from the raw input signal~\citep{wang-cvpr-2023b}, eliminating the need for a neural network. In terms of objective functions, most of the works focus on integrating a contrastive loss component \citep{zhao-iccv-2021, dong-cvpr-2023, kong-cvpr-2023b}, but there are also a few methods that leverage a multi-scale reconstruction loss~\citep{wang-cvpr-2023b}. Another important direction is closely following the strategy in BERT \citep{devlin-naacl-2019}, using discrete tokens and predicting the id from a learned vocabulary \citep{bao-ICLR-2022, zhou-ICLR-2022}. These tokens are obtained using a variational autoencoder (VAE), the pixels being mapped to a discrete latent space \citep{Ramesh-ICLR-2021}. With a similar goal in mind, other works map each token into the nearest entry from a codebook of learned embeddings \citep{van-NeurIPS-2017, Esser-CVPR-2021}.

MaskCo~\citep{zhao-iccv-2021} is a region-level contrastive learning framework. It begins by augmenting images to create two distinct views of the same image, with one view partially masked. The model is then trained using contrastive learning to align the features of the masked regions with those of the corresponding regions in the unmasked view.
\cite{kong-cvpr-2023b} reinterpreted MIM via the lens of contrastive learning. They found a formulation showing that the classic MIM approach is equivalent to a setting with Siamese networks, where one network reconstructs the masked tokens and its counterpart focuses on the unmasked tokens. The goal is to closely align the outputs of these models.
MaskCLIP~\citep{dong-cvpr-2023} combines masked image modeling and the CLIP contrastive loss between images and text in a single framework. Compared with vanilla CLIP, MaskCLIP has an additional loss for reconstructing a masked image and the two losses share the same visual encoder.

\cite{zhu-tpami-2022} proposed to integrate an additional masked contrastive objective into an RL pipeline dedicated to video games. Sequences from the video clips are sampled and then each frame is encoded using a CNN-based network. Then, besides the main RL policy network, an auxiliary branch is added that employs the student and teacher (exponential moving average of the student) framework. While the former receives masked input features, the latter is fed with the intact latent representation. The objective is to maximize the similarity between the two resulting embeddings.

Aside from studying the contrastive objective, additional components added to the loss function were sometimes explored. Interestingly, such studies used multimodal data.
To learn better visual representations during MIM pre-training within the medical domain, \cite{chen-miccai-2023} leveraged an additional modality (text). Two different encoders are used (one for each modality), however, the text encoder is fed with the average global vision embedding as well. The method consists of two reconstruction objectives, one for the image and one for the text, each with its own decoder. Furthermore, the authors employed a third contrastive objective between the fully visible encoded text and a decoded representation (using an additional decoder) of all the image embeddings (masked and unmasked tokens).
\cite{jang-neurips-2023} improved the MAE pre-training strategy on multimodal data by optimizing the latent encoding space. Some neighboring tokens of the unmasked tokens are sampled as well, in order to compute the reconstruction loss and use the resulting gradient to produce a more explicit latent representation. This is further used to compute the final reconstruction loss. Moreover, an additional contrastive loss is employed to maximize the similarity of the two resulting latent representations of a task, while minimizing it for different tasks.

\cite{wang-cvpr-2023b} showed that, in the previous masked image modeling approaches, it is difficult for the lower layers to learn inter-patch relations. To alleviate this issue, the authors introduced LocalMIM. They addressed the problem by using a loss function based on the weighted sum of the local patches losses. Their pipeline also includes a multi-scale reconstruction task, in which the model is supervised with different feature descriptors.

\cite{huang-cvpr-2023} presented a two stage distillation framework. In the first stage, the student network distills the task-agnostic knowledge of the teacher, and the authors chose MIM as a proxy task for this goal. Thus, the student learns to align its visible and masked patches representations with those of the teacher. In the second stage, the classic task-oriented knowledge distillation takes place.

\cite{wu-miccai-2023} presented a modified MAE framework for Whole Slide Images (WSI). After removing the background, the images are split into patches (some of which are masked), and for each patch, a feature vector is computed using DINO \citep{caron-iccv-2021}. Following \cite{zheng-miccai-2022}, trainable anchor vectors are employed, which are then used to calculate the distance and polar angle between the features and the anchor vectors. All these representations computed only for the visible patches are encoded, appended with mask tokens, and finally decoded to reconstruct all WSI representations. The cross-attention units between the patches and anchors from both the encoder and the decoder are bidirectional.

\cite{bao-ICLR-2022} adapted the pre-training strategy from BERT to images. First, the image patches are mapped to discrete visual tokens using a discrete Variational Autoencoder (dVAE) \citep{Ramesh-ICLR-2021}. Then, some tokenized patches are randomly masked and replaced with a special mask token. These are then passed through the encoder together with the visible tokens. The final objective is to predict the id of the masked tokens from the learned vocabulary of visual tokens. Instead of a decoder, this method uses only a classification head, which is dropped during fine-tuning. \cite{li-ECCV-2022} extended BEiT by employing multiple tokens for each patch. Inspired by the aforementioned work, \cite{zhou-ICLR-2022} proposed to learn both the tokenizer and the encoder. Their method adopts the teacher-student framework, such that the student tries to reconstruct the corresponding masked tokenized patches from the teacher and to match the \emph{[CLS]} token.

\subsection{Architecture and Objective}

A number of papers contributed to both the model architecture and the objective function. An important subset of these studies made their architectural changes as a result of integrating the new objective~\citep{fei-cvpr-2023, xue-cvpr-2023, wang-miccai-2023, liang-arxiv-2022}. For example, the introduction of adversarial loss by \cite{fei-cvpr-2023} requires an additional discriminator, and the multi-scale reconstruction loss used by \cite{xue-cvpr-2023} implies additional adapters to ensure compatibility between features. Another direction of research that requires both architectural and objective function modifications is the integration of multiple modalities~\citep{gong-iclr-2023, guo-cvpr-2024}. In such scenarios, each modality typically requires its own dedicated modules, leading to architectural adjustments, while an additional loss term is introduced to ensure alignment between the features of the two modalities.

Rather than following the standard approach in MIM with a reconstruction objective, \cite{xue-cvpr-2023} proposed a unique loss function that aligns features extracted from visible tokens with those extracted by a teacher model across various architectural levels. To facilitate this alignment, this study introduced a novel module named Dynamic Alignment. This module is specifically designed to ensure compatibility between the two sets of features, enabling more effective feature alignment.

For an improved objective function, some methods were augmented with additional loss components.
\cite{liang-arxiv-2022} demonstrated that an additional auxiliary supervised classification task helps the MAE pre-training framework. Besides the main reconstruction loss, the authors integrated another branch that takes only a subset of the visible encoded tokens as input. Further, the branch applies a global average pooling operation on the visible tokens, and finally predicts the class through a multi-layer perceptron. During the fine-tuning stage, all tokens are used.
\cite{fei-cvpr-2023} enhanced the standard MAE pipeline by integrating a discriminator for adversarial training. Notably, this discriminator, which shares parameters with the MAE's encoder, is trained to distinguish between synthesized and real patches. This enhancement is an addition to the existing pipeline, with the typical reconstruction loss still being present in the training process.

Inspired by masked autoencoders that process both visual and textual modalities, \cite{fuller-neurips-2023} adapted the masking pre-training framework for optical and radar inputs. Both modalities, after being aligned, tokenized and randomly masked, are encoded using an individual encoder, and then, they are jointly processed by a multimodal encoder. The resulting embedding is decoded to reconstruct both input images. A contrastive loss between the mean embeddings of the individual encoders is adopted to match sensor measurements from the same timestamp, while maximizing the difference between those at different timestamps.

\cite{wang-miccai-2023} demonstrated how MAE can boost the performance of 3D medical image segmentation. The input volume (a 3D scan) is split into equal subvolumes and these are randomly masked. Then, different views (frontal, horizontal, or longitudinal) are obtained, and a further arbitrary rotation is applied. During pre-training, besides the main reconstruction objective for each view, three more losses are utilized: the rotation angle estimation, a contrastive loss, as well as an additional MSE loss between two different reconstructed views (after being normalized). The architecture of the encoder is based on Swin transformer. A cross-attention module, which attends to the features between two views, is integrated before the first level of the decoder. 

In the work of \cite{gong-iclr-2023}, the audio spectrogram and the images are tokenized. The unmasked tokens are encoded with separate encoders, while also adding a corresponding modality embedding. Then, three separate forward passes are performed through a common encoder: one for each modality embedding, as well as the concatenation of the two. The concatenated tokens, together with the masked tokens, are decoded and the reconstruction loss is applied. Furthermore, a contrastive loss is applied between the averaged-pooled encoding of each modality. Rather than relying on global information for learning audio-visual features, \cite{guo-cvpr-2024} took a different approach by focusing on a finer-grained level, as well as improving the linkage between the two modalities. Their pipeline involves two more objectives besides the contrastive loss between the embeddings of the two modalities. One additional objective aims to reconstruct the original signal from the unmasked tokens of both input sources, while the second additional objective aims to reconstruct the embeddings using the counterpart tokens and some learnable queries.

The contribution of \cite{pei-cvpr-2024} is twofold. Firstly, they integrated MAE pre-training for videos by applying a consistency loss between two successive frames that are masked in the same manner. The masked frames are then encoded and decoded with two different networks (one is an exponential moving average of the other). Secondly, they proposed an analogous architecture using sparse convolutional layers instead of ViT, which results in lower computational costs.

\subsection{Masking Strategy and Features}

Several studies have impacted both the masking strategy and the target features. Generally, these studies adjust their target features to reflect changes in masking strategy or in the type of input features that they use~\citep{zhao-cvpr-2023, zhang-cvpr-2023, lin-cvpr-2023, zhao-iccv-2023, zhang-aaai-2023, walsh-tip-2023}. For example, a common scenario is when the model processes pairs of images~\citep{zhao-cvpr-2023, song-aaai-2023}. In such a scenario, the model employs shared decoders and the target image for reconstruction for at least one of the decoders is altered to reflect the relationship between the images comprising the input pair. Additionally, other studies~\citep{zhang-cvpr-2023} use the feedback from an auxiliary model to identify the relevant regions for masking and to define the space in which the reconstruction loss is applied.


A couple of papers in this section were applied to object tracking.
\cite{zhao-cvpr-2023} presented a method for learning representations useful for object tracking. The method is based on MAE, but the authors used two inputs, one is the search region and the other one is the template. The MAE is trained to reconstruct the search region as it is, and to recreate the template in the position found in the search region.
Inspired by the MAE framework, \cite{song-aaai-2023} proposed to boost the performance of a ViT model used for object tracking (given an object in a template image, find the same object in the search image) by applying MIM as an additional concurrent task. Both input images are concatenated and passed through the encoder. Besides the main task head, two more decoders are employed. After a high portion of the embeddings is masked, the two sets (each corresponding to an image) are separately fed through one decoder to reconstruct the two frames. The other decoder only receives as input the token embeddings of the search image and reconstructs the template image.

Other works focused on implementing MIM for point clouds. On the one hand, 
\cite{zhang-cvpr-2023} presented I2P-MAE, a method designed to learn better 3D features by reconstructing masked point clouds. The approach leverages 2D pre-trained models to keep the important point tokens visible while masking. Moreover, the 2D models are used to get the target representations for a semantic reconstruction loss that is applied on the visible tokens, after the decoder.
On the other hand, some works harnessed 2D images in their point-cloud methods.
\cite{guo-ijcai-2023} proposed a pre-training method for 3D point clouds by leveraging the corresponding 2D visual representation. The 3D points and their 2D projections are jointly encoded. The resulting encoding is randomly masked and passed through a two-stage decoder. First, there is a shared decoder, then each representation continues through a separate module to reconstruct both input modalities. 
\cite{zhang-aaai-2023} utilized the mask reconstruction strategy in 3D segmentation due to the lack of supervised data, as well as the domain difference between training and testing data. During training, given a pair composed of a 2D image and a 3D point cloud, patches from one modality are masked and the model tries to estimate them using the other modality. A CNN backbone with a lower masking ratio is employed. Having two different datasets (source-labeled and  target-unlabeled), the MIM is performed on both datasets, while the supervised task is performed only on the former.

The teacher-student framework was utilized in a couple of scenarios.
\cite{lin-cvpr-2023} combined self-supervised knowledge distillation and masked image modeling into a single framework. In the proposed pipeline, the teacher network processes an image from the same class as the student network. The student network processes a masked image, being trained to maximize the similarity between its class token and the teacher's class token. In addition, the student is trained to distill the knowledge of the most similar tokens of the teacher.
The study of \cite{zhao-iccv-2023} integrated the MAE pipeline into a teacher-student setting for domain adaptive object detection. In this setup, the student network has a dual focus: it learns the detection task using labels generated by the teacher network, and concurrently, it undertakes the reconstruction of missing multi-scale features from the target images. This reconstruction aspect is pivotal, especially when the availability of pseudo-labels from the teacher is limited, as it significantly improves the model's adaptability to the target domain, ensuring more robust and accurate object detection performance.

\cite{wei-iccv-2023} evaluated the efficacy of using image generation as a self-supervised pre-training task, finding that it yields only marginal improvements in downstream recognition tasks when applied within a diffusion model framework. In response to this observation, the authors presented a novel strategy that merges MAE with diffusion models for self-supervised pre-training, focusing specifically on an inpainting task. This approach demonstrated competitive performance, aligning closely with state-of-the-art methods in image recognition, thus offering a compelling alternative to enhance pre-training effectiveness. 

Inspired by MAE, \cite{walsh-tip-2023} modified the MAE framework in order to boost performance when applied to downstream classification tasks in few-shot scenarios. Rather than reconstructing the original pixel space, the authors proposed to operate in the latent representation space of a frozen backbone. Two subsets of images, called support and query, are firstly embedded, the latter's embeddings being masked. Then, the support embeddings are encoded, concatenated with mask tokens, and decoded to estimate the corresponding query embeddings, MSE being the computed reconstruction loss. After being pre-trained with a large labeled dataset, the method is further trained on a smaller dataset (containing few examples per class) in order to learn global information about each class.

\subsection{Masking Strategy and Architecture}

A number of studies have influenced both the masking strategy and the employed architecture. A notable proportion of these studies have multimodal inputs \citep{mizrahi-neurips-2023, lu-miccai-2023, li-iccv-2023, chen-cvpr-2023b, bachmann-eccv-2022, seo-icml-2023}, which typically requires distinct modules for each modality, coupled with carefully designed masking strategies to prevent information leakage between modalities~\citep{chen-cvpr-2023b, seo-icml-2023}. Additionally, some studies have adapted the MAE framework to be compatible with other popular architectures in computer vision, such as hierarchical transformers~\citep{huang-neurips-2022b} and CNNs~\citep{woo-cvpr-2023}. Furthermore, there are works that specifically develop architectural modules to directly influence the masking strategy employed during pre-training~\citep{madan-wacv-2024}.

While ViT is the main architecture employed in MIM, convolutional layers were still found to be beneficial. \cite{woo-cvpr-2023} implemented the MAE framework for convolutional networks. One of the changes was to create the masks based on the deep feature maps of the encoder and resize them to the resolution of the input images. The second change was also in the encoder, where the authors used sparse convolutional layers to preserve the speed improvements brought by the masking.

\cite{huang-neurips-2022b} introduced a set of changes required by the Hierarchical Vision Transformer architectures in order to be compatible with the MAE framework, where the masked tokens are ignored from the input sequence. There are two problems when applied directly, one is the window attention with non-overlapping windows, and the other is the usage of convolutional and pooling layers. For the first issue, the authors’ solution is to group the tokens from uneven windows with a novel Optimal Grouping algorithm and then apply the mask attention. For the second issue, they opted for using sparse convolutions.
\cite{seo-icml-2023} made several contributions to adapt the MAE pre-training strategy to their scenario. Firstly, the linear projection layer of the ViT architecture is substituted with CNN layers. However, positional, viewpoint and timestep embeddings are still added. Secondly, the method operates on sequences of images that have multiple viewpoints, requiring a novel masking strategy, on the one hand, to fully mask one viewpoint per video frame, and on the other hand, to mask the latent feature maps of the intact frames. In order to facilitate the reconstruction task, the encoder is fed with tokens from different viewpoints, as well as from adjacent frames. This pre-training framework allows the authors to train a world model for visual robotic manipulation.

\begin{figure*}
    \centering
\includegraphics[width=0.98\linewidth]{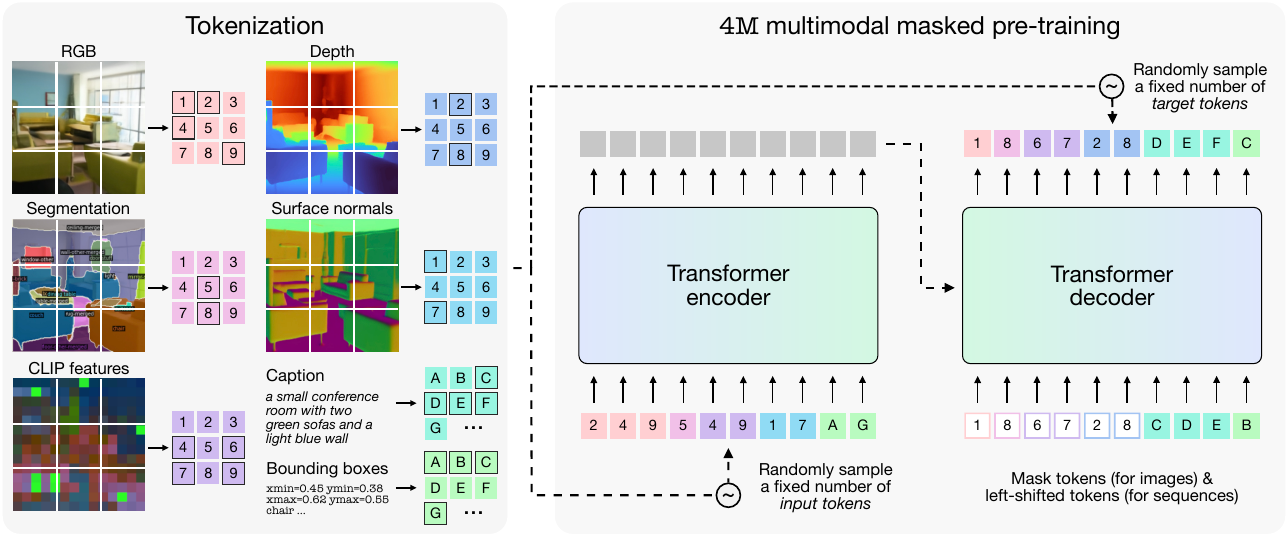}
    \caption{The 4M framework for multiple modalities, courtesy of \cite{mizrahi-neurips-2023} (image available under the ArXiv perpetual license agreement). 4M belongs to the category of reconstruction-based methods shown in Figure \ref{reconstruction}, as it tries to recover masked discrete tokens using the cross-entropy loss.}
    \label{fig:4M}
\end{figure*}

Harnessing multiple sources of information during MIM pre-training was demonstrated to bring great benefits, especially for multimodal downstream tasks. \cite{mizrahi-neurips-2023} proposed an MAE framework that can be used with multiple modalities. Given a set of modalities, each is tokenized into a common representation form. Rather than using all tokens, only two subsets from each modality are sampled: one that is encoded and the other one that is masked and then reconstructed. A joint encoder is employed for all input types. Still, a modality embedding is added to each token to indicate its source. In the cross-attention layers of the decoder, the embeddings corresponding to one modality are masked from the rest, while using all resulting tokens from the encoder. Besides demonstrating good results on downstream tasks, the method shows promising cross-modal generative capabilities.
To leverage the information from multiple input data types and learn richer feature representations, \cite{bachmann-eccv-2022} proposed to pre-train a network with multiple modalities. Their self-supervised method involved three types of data: an RGB image, a depth map and a semantic segmentation map. Using the ViT architecture, the modalities are split into patches and projected into tokens (with separate layers for each modality). Then, a great proportion of tokens from each modality is masked, while the rest (unmasked) are encoded (using a joint encoder) and concatenated with the masked tokens. A separate decoder for each data type is used to reconstruct the corresponding input. In the first stage of the decoder composed of a cross-attention layer, the tokens associated with the respective modality are given as queries, while the given keys and values are from all tokens. Experiments carried out on downstream tasks for all three data types indicate that the proposed strategy shows competitive results.
PiMAE \citep{chen-cvpr-2023b} is a self-supervised framework based on MAE, which learns representations that capture interactions between point clouds and images. Overall, the approach is based on the usual reconstruction objective for each modality. However, in contrast to MAE, the masking strategy in this case is designed to be complementary between the two modalities. In terms of architectural changes, the encoder and decoder include some common blocks between the two modalities, but they also have modality specific layers.
\cite{lu-miccai-2023} combined two sources of information (Hematoxylin and Eosin and Immunohistochemical staining images) to detect breast cancer, adopting MAE as the base for their method. Both images are split into patches, some of which are randomly masked, while the remaining visible patches are fed together through a ViT-based model. The resulting embeddings, together with the learnable mask embeddings, are further processed by two self-attention modules (one specific for each modality), as well as a cross-attention module (that processes all embeddings). Finally, two separate decoders reconstruct the original images, each receiving the modality-specific embeddings, as well as the inter-modal ones. More recently, \cite{Zou-ECCV-2024} employed the MAE framework to learn robust cross-modal representations. Their primary contribution is the MultiModal 3D Interaction module, which takes as input the concatenation of masked 3D volume features from both video frames and LiDAR sensors. Built upon self-attention layers and feed-forward networks, this module outputs a representation that is then split along the channel dimension and fed into modality-specific decoders to reconstruct the original inputs.

\cite{li-iccv-2023} proposed a novel two-stage pre-training framework for video foundation models. The initial stage focuses on training the model to align the features extracted from masked frames with those derived from an image foundation model on unmasked frames. In the subsequent stage, the authors introduced a text encoder and a cross-modality decoder to further train the model for video-text matching and masked language modeling, while maintaining the training objective employed in the first stage.

Scale-MAE \citep{reed-iccv-2023} introduced a pre-training method suitable for scale-dependent domains, specifically testing it on remote sensing data. The method is similar to the MAE framework, but it has some key changes. First, the positional embeddings that are added to the token embeddings depend on the Earth area covered in the image. Second, the authors changed the decoder to use a three-stage architecture. The first stage is a transformer-based block. The second stage performs upsampling with deconvolutional layers. The last stage employs Laplacian blocks to reconstruct the high and low frequency features.

\cite{madan-wacv-2024} introduced a learnable masking module that facilitates curriculum learning within the MAE framework. Initially, the new module creates masks that are easy to reconstruct. The training objective is changed over time to adopt an adversarial role, progressively creating more challenging masks for the MAE to reconstruct. This dynamic adjustment of the training masks enhances the MAE's ability to handle increasingly complex reconstruction tasks, thereby improving its learning efficiency and robustness.

\subsection{Masking Strategy and Objective}

A body of works have contributed to both the masking strategy and the objective function. These works either introduced a novel masking strategy that excels in hiding the salient information \citep{li-neurips-2021, yuan-neurips-2023, zhang-iclr-2023}, or adapted the masking strategy to a different input type \citep{yu-cvpr-2022, wang-aaai-2023}. Further, to maximize the benefit of using a new masking strategy, the objective function is altered to be more suitable for recovering the original input source~\citep{yu-cvpr-2022, wu-cvpr-2023a, zhang-iclr-2023, wu-iclr-2023}. For example, \cite{yu-cvpr-2022} performed the masking in the discrete space of a dVAE and, due to this change, the reconstruction loss is the cross-entropy. Another example is the work of \cite{wang-aaai-2023}, in which the contrastive loss is employed to align the token representations of two different masked versions of the same image.

\begin{figure*}
    \centering
    \includegraphics[width=0.98\linewidth]{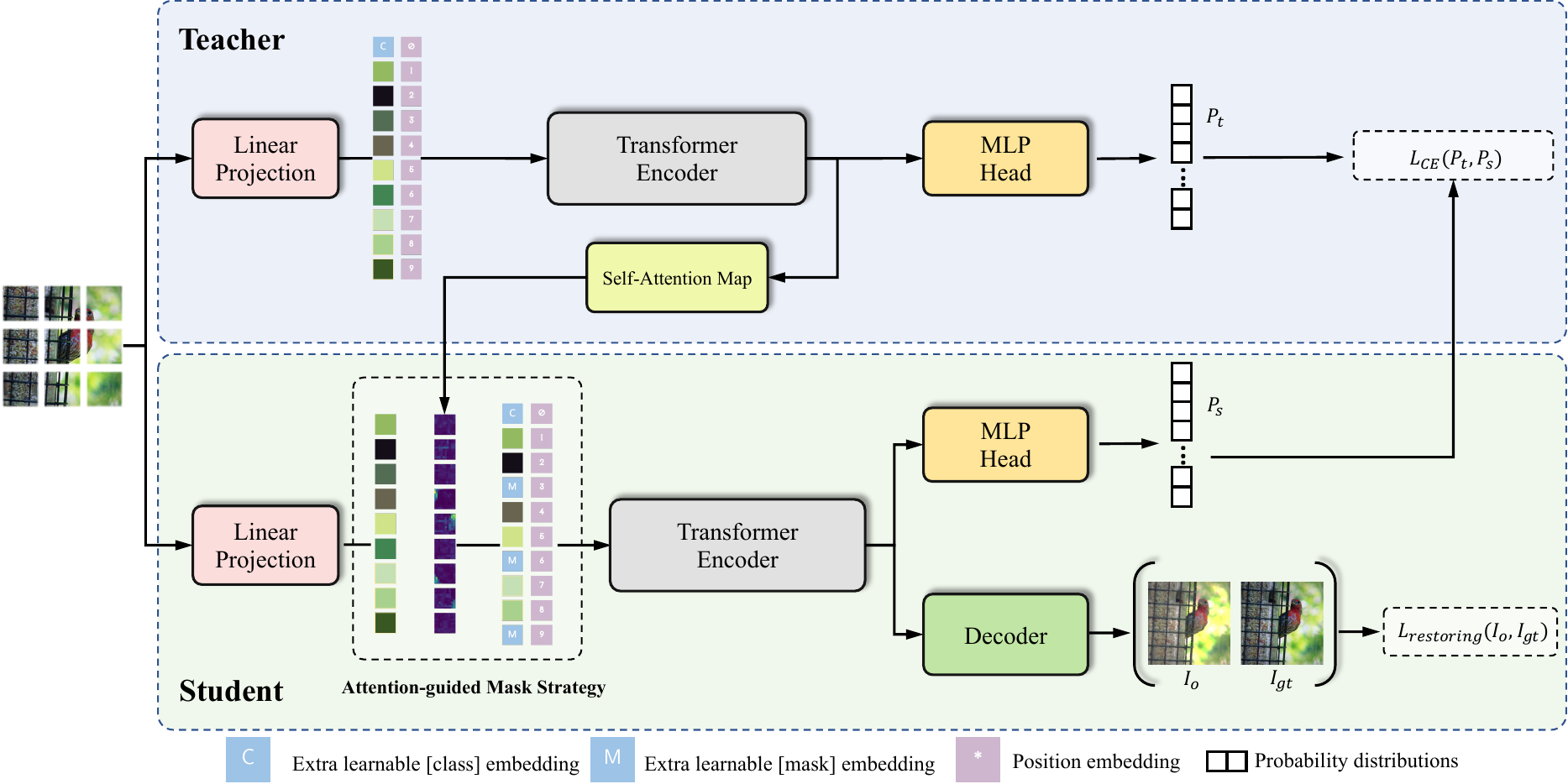}
    \caption{MST pipeline, courtesy of \cite{li-neurips-2021} (image licensed under CC BY 4.0). MST leverages an ${L}^1$ loss to reconstruct the patches that are masked according to the attention maps provided by the teacher, thus falling into the category of reconstruction-based methods depicted in Figure \ref{reconstruction}. The cross-entropy loss is only used to synchronize the teacher and the student.}
    \label{fig:MST}
\end{figure*}

A consistent number of papers in this section employed an objective function composed of multiple losses.
Masked Scene Contrast~\citep{wu-cvpr-2023a} is a framework for 3D representation learning that utilizes contrastive learning. This framework generates input pairs through a series of data augmentation techniques and applies complementary masking. The contrastive learning objective is then employed between the features of the unmasked and masked patches. Additionally, the framework incorporates a reconstruction loss to enhance learning efficacy.
\cite{wang-aaai-2023} integrated MAE in their method for self-supervised video hashing. A video clip is first downsampled using a CNN, and then, a token is extracted for each frame. Two different subsets of token frames are sampled, each being then fed into an encoder and hashed. After some of the hashed embeddings are masked, a decoder is used to reconstruct the original frame tokens. A contrastive loss is added to maximize the similarity of the mean hash embeddings between the two subsets.

\cite{wu-iclr-2023} slightly modified the MAE model by adding Gaussian noise to all pixels in the input image. With this change, besides reconstructing the masked regions, the objective is further extended to decode all denoised patches. The experiments demonstrated the superiority of this method over the original MAE.
\cite{zhang-iclr-2023} introduced some improvements to the teacher-student MIM based on contrastive pre-training. The first addition is a ranking component that divides the patches into two subsets: the ones that contain salient information and the meaningless ones. The former subset is passed to the student, while the latter is given to the teacher, both being partially masked. The first objective is to reconstruct the masked patches of the student, as they are harder to predict due to containing more salient information. A second loss is used to align the globally encoded representations of the two subsets by maximizing the similarity between the embeddings of the \emph{[CLS]} token. While the same encoder is used for both models, the gradients do not flow through the teacher.

Aside from multi-component losses, informed masking strategies were also employed in some studies.
The Masked Self-Supervised Transformer (MST)~\citep{li-neurips-2021} selects patches for masking based on low responses as determined by attention maps from a teacher network, which is an EMA of the student network. The selected patches are replaced with a special token. The training of MST includes a dual-objective approach: a reconstruction loss to rebuild the masked inputs, and a cross-entropy loss designed to synchronize the teacher and student networks, particularly in class differentiation. A detailed overview of the pipeline is depicted in Figure~\ref{fig:MST}.
\cite{yuan-neurips-2023} tailored the MIM pre-training strategy for human perception tasks. They began by detecting the human parts and masking the patches corresponding to these parts, the objective being that of reconstructing the masked tokens from the visible ones. The authors also generated another masked view of the input image (sampling other human parts), aiming to align the global representations of both views (by applying a contrastive loss on the \emph{[CLS]} tokens).

\cite{Fan-ECCV-2024} introduced a masking strategy that selects important patches according to their similarity with a given text, computed in the CLIP embedding space. They further employed a contrastive loss to align text features with the features extracted from the unmasked patches.

\cite{yu-cvpr-2022} presented a method inspired by BERT to pre-train transformers from point cloud data. First, the points are partitioned into sub-clouds, and a PointNet is applied to extract point embeddings. With the resulting embeddings, the authors trained a dVAE, where the encoder is a tokenizer, mapping the continuous embeddings into discrete tokens. After this stage, the transformer pre-training is performed, in which the model receives a masked sequence of point embeddings and learns to output the discrete tokens. The masked tokens are replaced by a learnable token.

A couple of papers opted for a contrastive objective between two different views of an image obtained by distinct augmentations.
Taking a different masking SSL approach, \cite{huang-iclr-2023} proposed a method whose objective is to be more robust for downstream tasks. On the input image, two different sets of augmentations are applied. The augmented images are encoded, and a projection layer transforms the encoded representations into multivariate Gaussian distributions. Finally, a series of learnable masks are applied. The difference between both masked probabilistic embeddings is computed, the objective being that of minimizing this difference.
\cite{yi-iclr-2023} presented a pre-training framework that combines MIM and contrastive learning. Different subsets (a more aggressive one and a lighter one) of augmentations are applied to the input image to create two views, the latter having a portion of its tokens masked. Then, two vision transformer models (where the one for the unmasked image is an EMA of the other model) are used to encode the tokenized patches. The contrastive loss is applied only between the corresponding corrupted patches.

\subsection{Masking Strategy and Task}


The research highlighted in this subsection focuses on developing MIM pre-training methods that are effective on specific downstream tasks. These studies accomplish this by creating custom masking strategies~\citep{cai-cvpr-2023, fu-cvpr-2023a, shen-iccv-2023, basu-cvpr-2024} or by transforming the input signal into a specialized feature space~\citep{zhu-cvpr-2023, liu-cvpr-2023b, li-cvpr-2023a, yan-cvpr-2023, pan-miccai-2023}. The custom masking strategies are created such that the features of interest are masked in the pre-training phase. For example, \cite{cai-cvpr-2023} learn representations that are meaningful for facial recognition by masking the regions that contain the eyes, nose and mouth, during pre-training.

To improve the masking strategy, some papers proposed an informed masking policy. 
\cite{cai-cvpr-2023} used masked autoencoders to learn rich, generic, transferable and robust facial representations from face videos. The masking prioritizes specific tokens (those containing eyes, nose, mouth and hair). The learned representations are then tested on downstream tasks, such as facial attribute recognition, facial expression recognition, DeepFake detection, and lip synchronization.
Aiming to improve the performance on WSI classification, \cite{tang-iccv-2023} studied several masking strategies to create a hard mining method useful for multiple instance learning. They observed that the best candidates for masking are the salient patches. To identify this type of patches during training, the authors proposed a pipeline in which the attention scores provided by a teacher network serve as indicators of patch saliency.
\cite{basu-cvpr-2024} addressed the shortcomings of gallbladder cancer detection in static images, proposing the use of video sequences instead. They adopted MAE as a pretext task, but presented an improved masking strategy that is able to hide the malignant regions more consistently, and thus learn a better representation of the disease. The masking strategy involved a Region Selection Network that generates a probability for each token, which is then used to sample the visible tokens.

A few papers adopted MIM to increase the quality of images.
The Saturated Mask AutoEncoder (SMAE), introduced by \cite{yan-cvpr-2023}, is a two-stage approach for few-shot HDR imaging. The first stage focuses on representation learning, which is performed via masked image modeling. In this stage, the method creates two additional images from the original frame using exposure adjustment. Next, all three images are randomly masked with a high masking ratio before passing them to the model.
LEMaRT \citep{liu-cvpr-2023b} is an effective pre-training framework when applied on image harmonization as a downstream task. In this approach, the masked patches are replaced with the patches taken from a perturbed version of the original image. The authors also investigated what is the best strategy for creating the masks, concluding that random masking works best for image harmonization.
Magnetic Resonance scans are generated in k-space and then transformed in the image domain with the inverse Fourier transform. To obtain an image with high quality and fidelity, the k-space needs to be fully sampled, but this is not realistic in most scenarios. To this end, \cite{pan-miccai-2023} proposed to leverage the MAE framework by masking data in the k-space, represented in 3D: height, width and time, along the first dimension. By reconstructing the missing tokens via the \textit{$L^1$} loss, the adopted ViT models learn a rich feature representation that is able to estimate the unsampled data from k-space at inference. The estimated k-space is further refined using three sequential transformer-based decoders (one for each pair of dimensions), and employing the High-Dynamic Range loss after each decoder.

\cite{li-cvpr-2023a} presented a framework for representation learning and image generation. The applied pre-training method is similar to MAE \citep{he-cvpr-2022}, but the tokens are given by a VQ-GAN tokenizer and the masking ratio is variable.

\cite{Zhai-ICCV-2023} demonstrated that MAEs can be used in class incremental learning as a rehearsal-based method. The efficiency of MAEs, which require only a few patches, allows for the storage of more examples from previous tasks. In addition, the authors designed a two-branch MAE architecture to ensure higher quality reconstructions, one branch being responsible with inserting details into the image.

\cite{fu-cvpr-2023a} introduced the TVC (text-guided video completion) task: the model needs to generate videos based on a subset of frames, while respecting some text instructions. Depending on the subset of frames, the task requires either prediction (future), rewinding (past) or infilling (between two moments). The authors proposed a training strategy that is based on masking frames, which addresses all three possible TVC subtasks.

\cite{shen-iccv-2023} introduced a novel self-supervised pre-training technique designed for point cloud videos, employing a unique approach that involves masking point tubes. This method focused on training the model to accurately reconstruct these masked tubes. Simultaneously, the model is engaged in training on a temporal cardinality difference task. This dual training strategy enhances the model's ability to understand both the spatial structure and temporal dynamics inherent in point cloud videos.

\cite{zhu-cvpr-2023} argued that MIM alone is not sufficient for downstream tasks such as geometric matching. Therefore, they proposed paired MIM, which reconstructs pairs of masked images instead of a single masked image. Their study demonstrated that this pre-training task is more effective for geometric matching, because such tasks require the correlation between two images.

\subsection{Features and Task}


Several studies aiming to develop MIM methods optimized for specific downstream tasks achieved their goal by exploring various types of target features for masking. Most of these works are in the medical domain~\citep{cai-miccai-2022, kang-miccai-2023, zhang-cvpr-2024}. The remaining two are unrelated, MAGVLT~\citep{kim-cvpr-2023} being a visual-language transformer, and MVD~\citep{wang-cvpr-2023a} being a distillation framework.

The adoption of MIM in the medical domain is highly motivated by the scarcity of annotated datasets.
\cite{cai-miccai-2022} proposed a method to pre-train a model using MIM, which is able to process both 2D and 3D ophthalmic images. The authors developed a new module, called Unified Patch Embedding, consisting of two branches, one for each data type. The module divides the inputs into equal patches and then masks most of them. Then, a common encoder computes the latent representations of the visible patches. Finally, two decoders are employed: one that reconstructs the patches, and another one that estimates their gradient maps (composed of horizontal and vertical edge maps). The experiments showed that the method yields state-of-the-art performance in ophthalmic image classification.
To boost the performance gains of an MAE in the ultrasound imaging domain, \cite{kang-miccai-2023} introduced an additional task during the pre-training stage. Due to the high noise-to-signal ratios in such images, they are initially blurred, so that masked patches are reconstructed, while the visible patches are deblurred. Thus, in contrast to the original MAE, all patches are passed through the encoder. The experiments on the downstream task of thyroid ultrasound image classification demonstrated leading results. 
Different from the previous methods, \cite{zhang-cvpr-2024} introduced a novel unsupervised domain adaptation framework based on MAE, in which the authors employed a convolutional architecture. The masked reconstruction task is applied to two input signals that are created from two volumetric scans: a local sub-volume and a global downsampled scan. When applied to the segmentation downstream task, the method uses a teacher-student framework. The teacher (an EMA of the student) generates a pseudo-label segmentation mask of a target domain image, and then, the student is trained using the resulting pair along with a sample from the source domain. 


MAGVLT~\citep{kim-cvpr-2023} is a non-autoregressive generative visual-language transformer trained jointly for image-to-text, text-to-image and image-text-to-image-text. The training objectives are three mask prediction losses, one for each task.

\begin{figure*}
    \centering
    \includegraphics[width=0.85\linewidth]{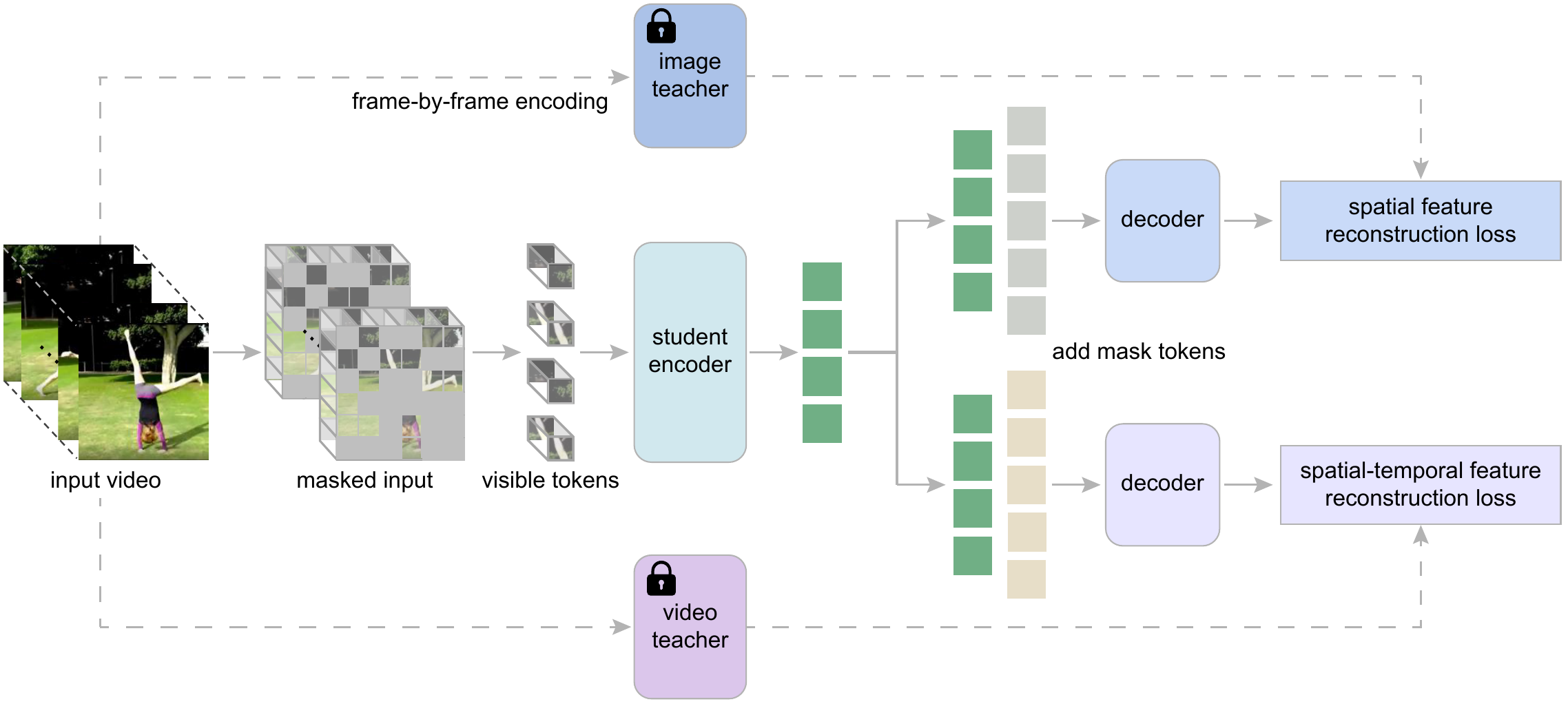}
    \caption{The MVD knowledge distillation pipeline, courtesy of \cite{wang-cvpr-2023a} (image available under the ArXiv perpetual license agreement). MVD is classified as a reconstruction-based approach (belongs to the generic pipeline illustrated in Figure \ref{reconstruction}), as it uses an $L^2$ loss between the output features of its two decoders and the features provided by two teacher networks, one for video and one for images.}
    \label{fig:mvd}
\end{figure*}

\cite{wang-cvpr-2023a} introduced Masked Video Distillation (MVD), a new method for self-supervised video representation learning, depicted in Figure~\ref{fig:mvd}.
This approach has two stages. The first stage is to train masked image models and masked video models as teachers.
The second stage is to train a student with the representations learned by the teachers as target vectors. The masking is also used in this latter stage.

\subsection{Architecture and Features}


Some studies contributed in terms of both model architecture and target features. The modifications proposed by these works are tightly coupled. On the one hand, some works proposed new target features and consequently implemented architectural adjustments to support these features \citep{jiang-cvpr-2023, huang-neurips-2023b}. On the other hand, other works extended the existing MIM frameworks \citep{dong-eccv-2022, lao-iccv-2023}, or even introduced new pre-training frameworks based on MIM~\citep{jiang-miccai-2022, yang-iclr-2023}.

Rather than using a single decoder, some works introduced multiple decoders in their pipeline. \cite{dong-eccv-2022} proposed a couple of modifications to the original MAE pre-training. They used two decoders: one for image reconstruction (the original task) and one for feature representation estimation. The latter one tries to predict the feature representation of the masked patches, the ground-truth coming from an EMA replica of the MAE encoder. For both decoders, some information about the visible patches from the encoder is injected with the cross-attention layers: the former receives low-level context, while the latter is given high-level features.
Observing that masked pre-training negatively affects the final layers of a deep ViT model, \cite{huang-neurips-2023b} proposed Masked Image Residual Learning (MIRL). The framework consists of dividing a ViT along the depth into an even number of stages. A decoder is added for each stage, which is fed with the corresponding intermediate embeddings, as well as the masked tokens. While the decoders in the first half reconstruct the main components of the input image, the rest estimate the residuals, i.e.~the differences between the target and the prediction.

\cite{jiang-miccai-2022} introduced a complex self-supervised self-distilled pre-training framework, demonstrating its capability on the 3D medical image segmentation downstream task. The first step is to generate two views of the same 3D image, split them into equally-sized patches, and then randomly mask them. Two encoder networks, teacher and student, are used, the former being updated as an exponential moving average with momentum of the latter. While the student processes only the masked patches of one view, the teacher fully encodes the same uncorrupted view, as well as the masked patches of the other view. Three independent single linear layers are then used to densely reconstruct the image from each embedded patch, estimate the masked patches and generate a global embedding of the image. Consequently, three losses are computed: the reconstruction loss between the predicted masked tokens of the student and the corresponding tokens encoded by the teacher, the cross-entropy between the global teacher's and student's embeddings of the two masked views, but also the reconstruction loss of the view predicted by the student.

Masked Shape Prediction (MSP)~\citep{jiang-cvpr-2023} uses geometric shape as prediction target for masked points. This target includes explicit shape context and deep shape features. Additionally, the architecture of MSP modifies cross-attention and self-attention layers to avoid possible masked shape leakage caused by including the masked part positions in the token embeddings. These modifications are designed to restrict the interaction among masked points.

\cite{lao-iccv-2023} presented a knowledge distillation technique suitable for object detection, based on the MAE framework. The student network receives a masked image and learns to recover the missing multi-scale features provided by the teacher network. The student and the teacher networks are based on convolutional layers. Thus, in the design of the student, the authors used masked convolutions to avoid information leakage.

\cite{yang-iclr-2023} introduced a masking pre-training framework that tackles the difference in training and test data distribution, the main idea being that of reconstructing images from other domains. The framework consists of one encoder and multiple decoders (one for each domain). First, the style-mixed image is obtained from the input image. The patches of the style-mixed image are masked, the visible tokens being passed through the encoder and all decoders to estimated the input image in all styles. The reconstruction loss is first employed for the predicted style corresponding to the input. The second stage of the framework involves taking the other estimated images, randomly masking their patches, and passing them through the autoencoder (with the decoder of the input style) to estimate the input image. The second objective is the reconstruction of the input style from all other inputs.

\cite{fu-arxiv-2021} proposed to integrate masked modeling for pre-training video-language tasks. A common cross-modal transformer is employed for the tokens of both modalities, with the objective of reconstructing them. While the text tokens follow the approach in BERT, the video frames are tokenized into discrete values with a dVAE, the masking being achieved at the pixel level by replacing image patches with zero. Their experiments attest the benefits of reconstructing discrete visual tokens.

\subsection{Objective, Task and Theoretical Analysis}

Two studies had an impact on the objective function and the downstream task, while also conducting a theoretical analysis. These works~\citep{zhang-neurips-2022, bashkirova-cvpr-2023} begin by conducting a deep analysis about different MIM aspects, and then, they provide solutions in order to improve them~\citep{zhang-neurips-2022} or exploit the properties induced by MIM to implement a particular downstream task~\citep{bashkirova-cvpr-2023}.

\cite{zhang-neurips-2022} unveiled several theoretical insights of the MAE paradigm. Initially, they uncovered a link between the MAE framework and contrastive learning principles, revealing that the reconstruction loss in MAE is analogous to the alignment loss found in contrastive learning. Further exploration provided theoretical assurances regarding the downstream efficacy of MAE models.
The connection with contrastive learning also implies the presence of feature collapse, a common challenge in contrastive learning, where aligning solely positive samples diminishes model effectiveness.
The researchers introduced Uniformity-enhanced MAE as a solution for the feature collapse problem. This adaptation modifies the loss objective to integrate a novel loss function, specifically designed to reduce feature similarity across unmasked views, thereby preserving feature diversity and enhancing model robustness.

MaskSketch~\citep{bashkirova-cvpr-2023} is a method to generate images from sketches using a masked generative transformer. In general, a masked generative transformer synthesizes new examples by accepting new tokens in consecutive iterations, and the accepted tokens are the ones above a certain threshold. In this case, the authors modified this threshold to depend on a distance computed between the self-attention maps of a sketch and the image that is being generated. Hence, the main observation is that the self-attention maps provided by a masked transformer are domain-invariant, preserving a similar structure for both sketches and natural images.

\subsection{Masking Strategy, Architecture and Objective}


A handful of works contributed to advancements in the masking strategy, the model architecture and the objective function. Some of these studies leveraged their three-fold contributions to tackle unique pre-training scenarios \citep{yu-iclr-2023, gupta-iclr-2023}. For example, \cite{yu-iclr-2023} presented a custom solution for images that contain text. Additionally, other studies discuss about the suboptimal choices made in previous MIM pipelines~\citep{bandara-cvpr-2023, Wei-ECCV-2024} and try to improve them. For example, \cite{Wei-ECCV-2024} imposed certain properties on the masked patches to avoid the patch correlation issue.

Adaptive Masking (AdaMAE)~\citep{bandara-cvpr-2023} employs an adaptive masking strategy for MAE performed by an additional neural network that assigns greater masking probabilities to the patches containing spatio-temporal information (a.k.a.~foreground). The new neural network gives a vector of probabilities from which the sampling is performed. Thus, it cannot be trained with the reconstruction loss. The solution for this problem is to use an additional loss function based on the reconstruction one, where the terms are weighted with the probability vectors given by the adaptive network.

\cite{yu-iclr-2023} presented a masking pre-training method specially designed for images with text. Random patches that contain text are masked, then passed through a CNN-based model to extract a feature representation. The feature maps are tokenized and then fed into a transformer. Following the Feature Pyramid Network (FPN) architecture \citep{lin-cvpr-2017}, the resulting embedding is upsampled and combined with feature maps from multiple levels. The first objective is to predict the masked words, while the second goal is to reconstruct the corrupted pixels (with the help of the predicted word tokens). A ROI-alignment unit is used to associate the feature maps with the masked regions.

In order to pre-train a ViT for videos, \cite{gupta-iclr-2023} proposed to first transform the input video into the latent space of a VQ-VAE. Then, each frame is transformed into a set of tokens, and a varying high ratio of the tokens from the whole video sequence is masked. The attention units of the model are modified such that every token has access either to only the surrounding tokens in the same frame, or to a small neighboring region of tokens along all dimensions (i.e.~including time). The objective is to estimate the masked tokens by minimizing their negative log-likelihood.

In their work, \cite{qiu-cvpr-2024} introduced a general framework to learn various computer vision tasks with transformers. They framed the input and output of each task (e.g.~detection and segmentation) as a sequence, and used an encoder-decoder transformer with bidirectional attention masks. To capture a rich context for each task, they employed MAE pre-training by reconstructing the sequence of tokens.

\cite{Wei-ECCV-2024} analyzed the drawbacks in previous latent-space MIM applications \citep{yi-iclr-2023}, showing how a conventional reconstruction loss restricts diverse latent learning. To counter this, the authors introduced a patch discrimination objective to increase similarity between the predicted and the corresponding target latents. Additionally, \cite{Wei-ECCV-2024} tackled the issue of patch correlation by changing the masking strategy. They used a high masking ratio (90\%), a gap between adjacent patches, and a similarity constraint on visible and target patch sets. Further, the last contribution was an improved decoder architecture, suitable for latent representation prediction and incorporating self-attention, cross-attention layers, and visual cues from visible patches.

\subsection{Masking Strategy, Features and Objective}

Some papers had an impact on the masking strategy, the target features, as well as the objective function. The contributions to the masking strategies are highly varied: integrating new target features \citep{yu-ijcai-2022, li-icml-2022}, improving the masking policy based on guidance \citep{yao-aaai-2023}, and even leveraging multimodal data \citep{kwon-iclr-2023}. The objective function of these works is adapted as a result of the employed features.

Within the context of few-shot learning, \cite{yu-ijcai-2022} employed a masked autoencoder for reconstructing the latent embeddings rather than the original image. Additionally, each instance acts as a patch, and thus, the input consists of more images from the same class. After encoding the input, a high proportion of the input is masked and the decoder tries to reconstruct the masked embedding representation (given some identification variables). In this way, the backbone, i.e.~the encoder, learns more discriminative features and has a better few-shot performance.

In their work, \cite{li-icml-2022} introduced a pre-training method that is applicable to standard convolutional neural networks. The authors begin by removing patches from the image and substituting them with the mean value of the pixels. Different from previous methods, the masking tokens, corresponding to the previously erased patches, are introduced in the intermediate layers of the network. Aside from the reconstruction objective, another loss is added, which takes into account the difference between the discrete Fourier transforms of the original and reconstructed images. The role of the additional loss is to enhance the representations learned due to the interactions between patches at an intermediate level rather than at the lower levels.

\cite{yao-aaai-2023} proposed to use MAE for reconstructing a normal estimation of the masked input image and compare it with the original, in order to detect anomalies. The pre-training of the model follows the original MAE, only optimizing the masking strategy for contiguous blocks of patches. As the aim is to reconstruct the abnormal regions, during inference, a proposal masking unit is employed, estimating the likely locations of the image patches that contain anomalies in order to mask them. The unit is composed of a feature extraction model that obtains the latent representations of the input image patches and some prototype normal images (one example from each normal class), as well as a normalizing flow model for computing the likelihood.

\cite{kwon-iclr-2023} focused on implementing a masking pre-training strategy for both visual and textual data. The main idea is to mask one modality and reconstruct the missing data using the other input type. First, each modality is encoded with its own encoder, and further processed by its specific cross-attention encoder (which is also fed with the other modality embedding). Finally, the masked tokens are decoded using a transformer for images, or a linear classification layer for text. Besides the reconstruction objective, two more losses are employed to align the embeddings of the modalities.

\subsection{Masking Strategy, Architecture, Task and Objective}

Few papers have contributed in four directions, having an extensive scope: the masking strategy, the model architecture, the downstream task and the objective function. Among these, one study stands out for its integration within a reinforcement learning pipeline~\citep{lezama-eccv-2022}, illustrating how MIM principles can be extended beyond conventional computer vision tasks. The remaining works delve into multimodal data \citep{huang-neurips-2023a, ristea-cvpr-2024, georgescu-iccv-2023}, proposing innovative ways of combining diverse input sources, while simultaneously refining the pre-training methodology and the architectural design.

\cite{lezama-eccv-2022} proposed a Token-Critic algorithm to guide the synthesis of a non-autoregressive image generation model by predicting which tokens need to be sampled or not. To train the model, the following procedure is employed. First, a tokenized image (through a Vector-Quantized Autoencoder) is randomly masked. Then, using the transformer-based generative model, the image is reconstructed, while the critic must distinguish between the sampled and the original tokens. In this way, a completely masked tokenized image is gradually unmasked, using the Token-Critic to select which tokens to sample, eventually generating a new synthesized image.

\begin{figure*}[!t]
\centering
\includegraphics[width=1.0\linewidth]{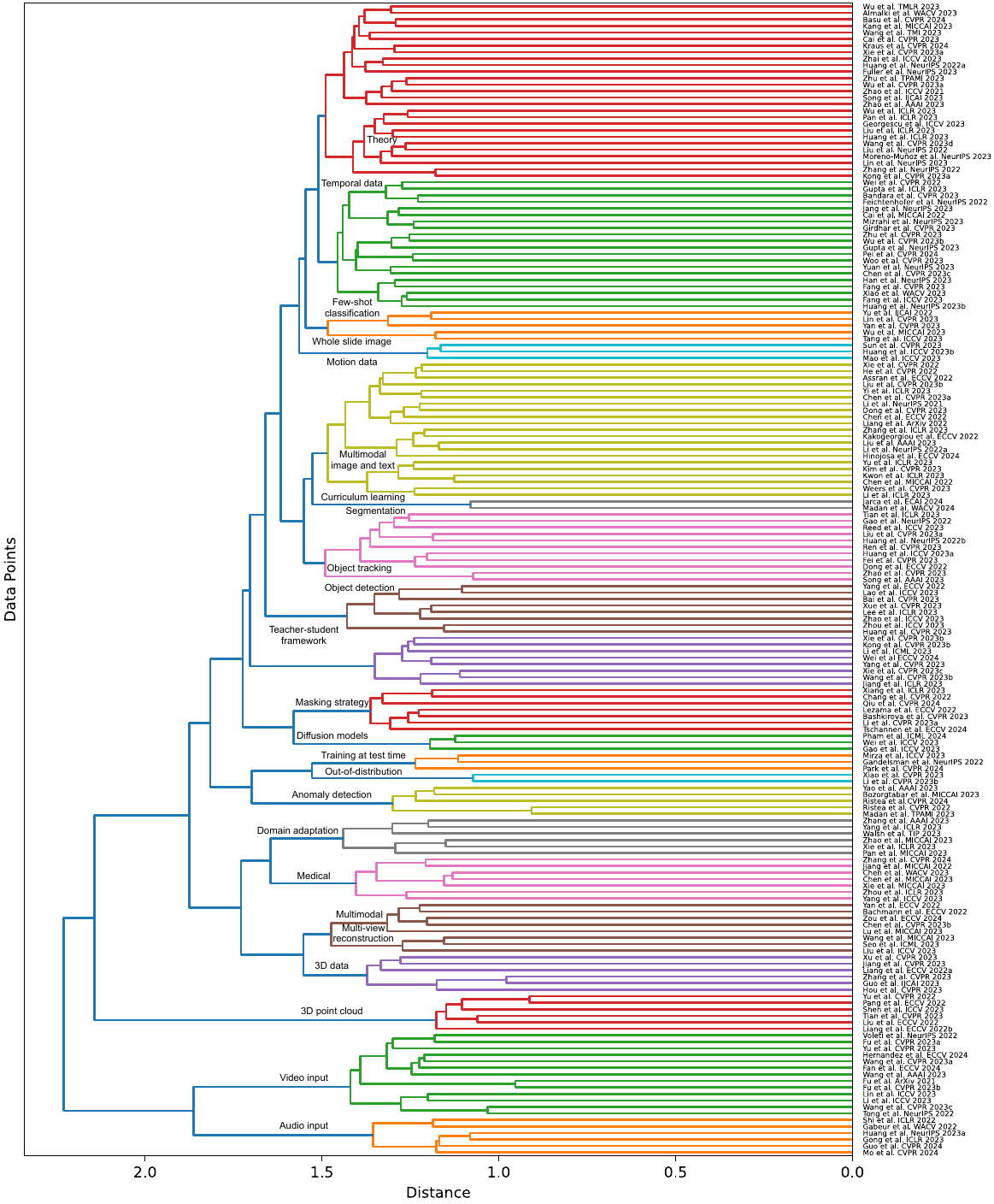}
\vspace{-0.3cm}
\caption{The dendrogram generated through hierarchical clustering (Ward linkage) applied on TF-IDF vectors derived from the titles and abstracts of the papers. Zoom in supported for the electronic version. Best viewed in color.}\label{dendrogram}
\vspace{-0.1cm}
\end{figure*}

\cite{huang-neurips-2023a} harnessed MAE to pre-train an audio-video model by integrating all prominent objectives of MIM into a two-stage framework. The first stage consists of reconstructing the original inputs, while the second stage adopts the teacher-student distillation scheme, where the objective of the student is to reconstruct the predictions of the teacher. The teacher receives the full visible inputs, while the student is fed with the masked modalities. During both training stages, two different masked views of the same modality are generated, encoded, and two contrastive losses are computed: between embeddings of the same modality, as well as across modalities. Then, a joint encoder fuses the two modality embeddings, in the end being decoded with separate decoders. \cite{georgescu-iccv-2023} also harnessed audiovisual data to enhance self-supervised representation learning. They proposed various pre-training architectures and objectives within a masked autoencoding framework to improve performance on audiovisual downstream classification tasks. The framework also supports multiple unimodal downstream tasks, using a single audiovisual pre-trained model.

\cite{ristea-cvpr-2024} developed a lightweight MAE tailored for video anomaly detection. This MAE model leveraged weights derived from motion gradients to emphasize foreground objects in the reconstruction loss. Additionally, the authors enhanced the training procedure by introducing synthetic anomalies, while using normal frames as the target for the reconstruction loss. Later stages of training involve a student decoder, which learns to mimic the output of the main (teacher) decoder, further refining the detection process.

\section{Automatic Clustering}

To complement the manual taxonomy, we generate a dendrogram by executing a hierarchical clustering algorithm on TF-IDF vectors, which are computed by concatenating the titles and abstracts of the surveyed papers. We employ TF-IDF vectors to diminish the influence of stop words and increase the importance of content words. The hierarchical clustering is based on the Ward linkage, which aims to minimize the total within-cluster variance. Each TF-IDF vector starts as its own individual cluster. At each step, the two clusters that result in the smallest increase in total within-cluster variance when merged are combined. The process of merging is repeated until all TF-IDF vectors are combined into a single cluster, thus generating a dendrogram. We opted for Ward linkage in detriment of other alternatives to ensure that the resulting clusters are more homogeneous. We illustrate the resulting dendrogram in Figure~\ref{dendrogram}.

By analyzing the dendrogram, we identify several relevant clusters, which are annotated in Figure~\ref{dendrogram}. The first observed category of clusters is related to the input data type: temporal data, 3D data or 3D point clouds, video, audio, and even multimodal. Other identified clusters are represented by the domain in which the MAE framework was used, or by the downstream task it was applied on: medical imaging, anomaly detection, image classification in few-shot scenarios, object detection, object tracking and semantic segmentation. The clustering algorithm was also able to capture more complex concepts: training at test time, multi-view masked reconstruction, and domain or out-of-distribution adaptation. Probably one of the most notable clusters is formed by the papers that adopted the teacher-student MAE framework based on a contrastive objective. Another category consists of methods that employed diffusion models. Finally, two clusters related to the input masking strategy and theoretical analysis overlap with our manual taxonomy.

\begin{figure}[t]
 \begin{center}
    \includegraphics[width=1.\linewidth]{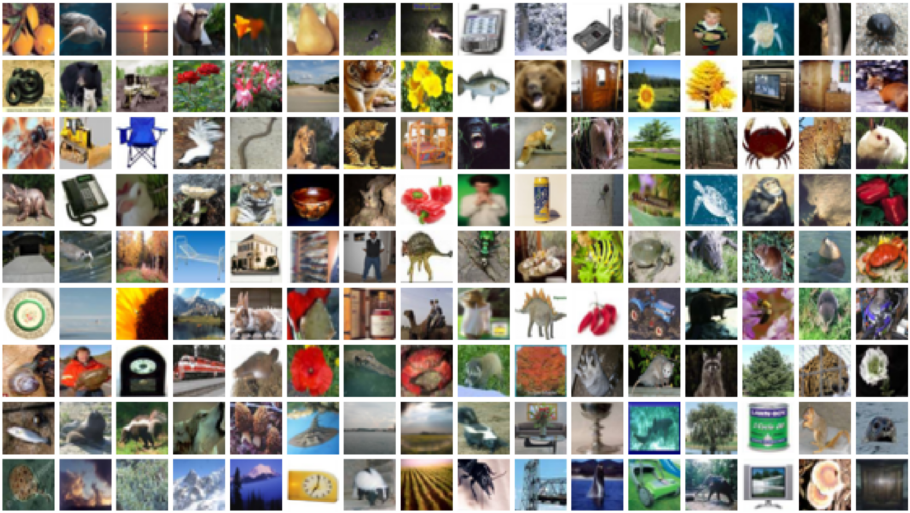}
\end{center}
  \caption{Sample images from CIFAR-100 dataset.}
\label{fig:cifar100}   
\end{figure}

When compared with the manual taxonomy, we consider that the automatically generated clustering provides a distinct yet equally-useful categorization of the papers.

\section{Datasets}

Various datasets have been used by different masked image modeling frameworks. Some of the most representative datasets are: CIFAR-100, ImageNet-1K, MS COCO, UCF101, ShapeNet, CC3M, FFHQ, LAION-400M, and Visual Genome. A brief description of these datasets are provided in the following part.

\begin{figure}[t]
 \begin{center}
    \includegraphics[width=1.\linewidth]{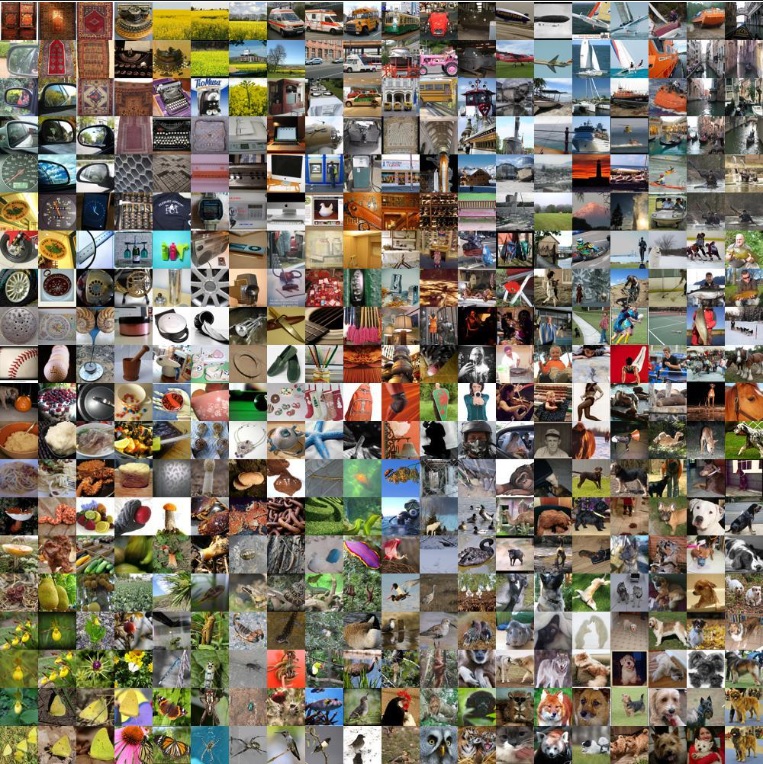}
\end{center}
  \caption{Sample images from ImageNet-1k dataset. Courtesy of \citep{Deng-CVPR-2009}.}
\label{fig:imagenet1k}   
\end{figure}

\begin{figure}[t]
 \begin{center}
\includegraphics[width=1.\linewidth]{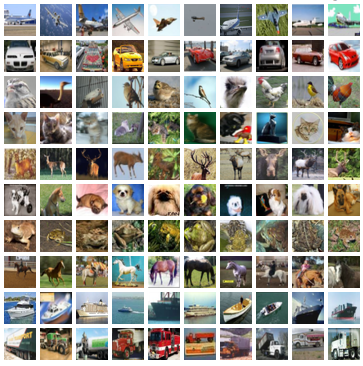}
\end{center}
  \caption{Sample images from MS COCO dataset. Courtesy of \cite{Lin-ECCV-2014}.}
\label{fig:mscoco}   
\end{figure}

\begin{figure}[t]
 \begin{center}
    \includegraphics[width=1.\linewidth]{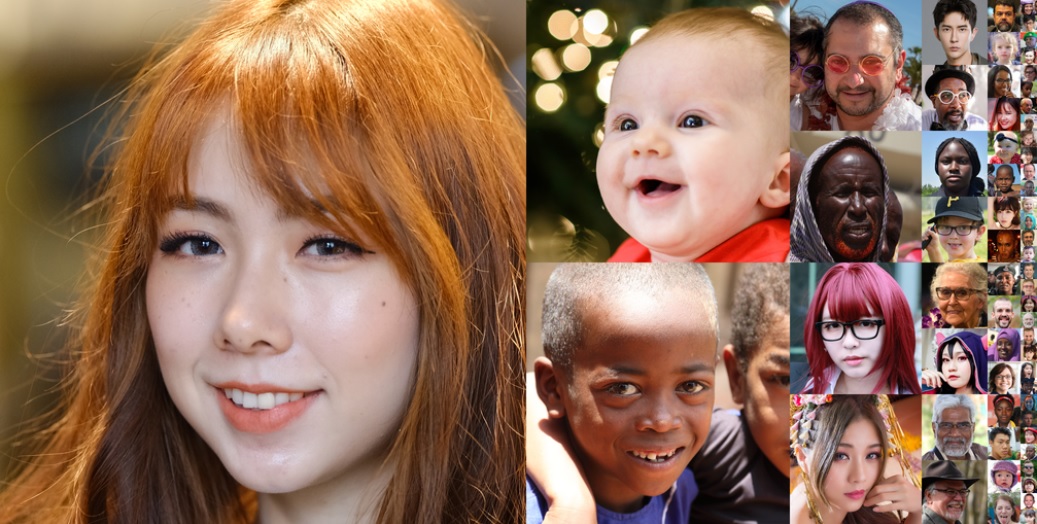}
\end{center}
  \caption{Sample images from FFHQ dataset. Courtesy of \cite{Karras-CVPR-2019}.}
\label{fig:ffhq}   
\end{figure}

\begin{figure*}[t]
 \begin{center}
    \includegraphics[width=1.\linewidth]{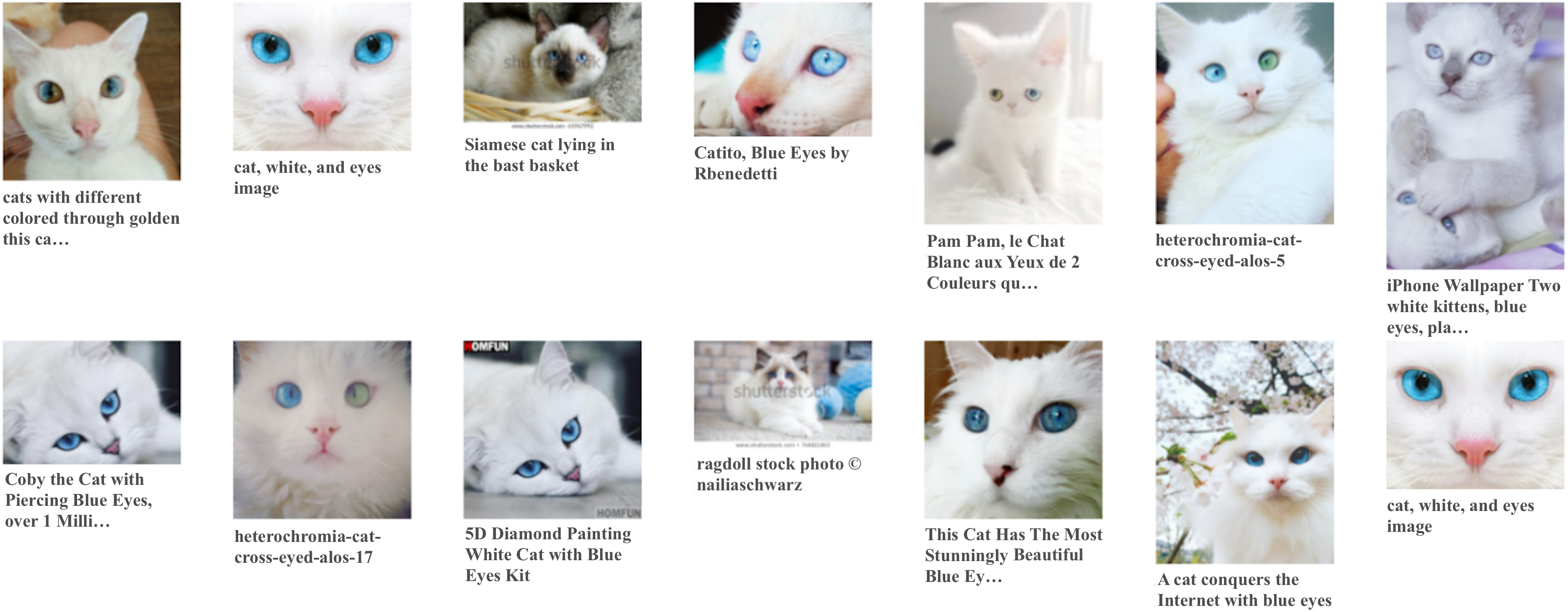}
\end{center}
  \caption{Sample images with the the queries ``blue cat'' or ``cat with blue eyes'' from LAION-400M dataset. Courtesy of \cite{Schuhmann-arxiv-2021}.}
\label{fig:laion400m}   
\end{figure*}

\textbf{CIFAR-100}~\citep{Krizhevsky-techrep-2009} contains low-resolution images (32$\times$32 pixels), with 600 images for each of the 100 available classes. The classes are grouped into 20 categories, called super-classes. The labels range from animals to humans and objects. CIFAR-100 is typically used to demonstrate MIM methods on downstream tasks. Some sample images from this dataset are shown in Figure \ref{fig:cifar100}. 

\textbf{ImageNet-1K}~\citep{Deng-CVPR-2009} is a subset of one of the most popular datasets, namely ImageNet-21K. It consists of approximately 1.45 million images divided into 1000 object classes. ImageNet-1K provides a diverse range of high-resolution images organized according to the WordNet hierarchy. ImageNet-1K is generally used in the pre-training stage. Some examples from this dataset are shown in Figure \ref{fig:imagenet1k}.

\textbf{MS COCO}~\citep{Lin-ECCV-2014} contains around 330k images, in which over 1.5 million object instances are annotated across 80 object categories. The dataset includes annotations for a wide variety of tasks: object detection, segmentation, and captioning.
Some sample images from this dataset are shown in Figure \ref{fig:mscoco}.

\textbf{UCF101}~\citep{Soomro-CoRR-2012} is a dataset of 13,320 videos spanning 101 action categories. The videos cover a wide range of activities including human actions, sports, and daily activities. This dataset is typically used by MIM frameworks in the video domain.

\textbf{Kinetics 400}~\citep{Kay-arxiv-2017} contains 400 human action classes, with at least 400 video clips for each action. Each clip lasts around 10 seconds and is sampled from a different YouTube video. The actions are focused on humans, covering a broad range of classes based on human-object interactions, such as playing instruments, as well as human-human interactions, such as shaking hands. Kinetics 400 is often employed in the pre-training stage of video-based MIM.

\textbf{ShapeNet}~\citep{Chang-TechRep-2015} is a dataset that contains 3D shapes, covering over 55 categories with about 51,300 unique 3D models. It includes both geometric and semantic annotations.

\textbf{CC3M}~\citep{Sharma-ACL-2018} consists of approximately 3.3 million images and their associated captions, describing the visual content of the images.

\textbf{FFHQ}~\citep{Karras-CVPR-2019} is a dataset composed of 70,000 high-quality (1024$\times$1024) images of diverse human faces, varying in age, ethnicity, and background. Some of the sample images from this dataset are shown in Figure \ref{fig:ffhq}.

\textbf{LAION-400M}~\citep{Schuhmann-arxiv-2021} is one of the primary datasets used for generative text-to-image models, consisting of 400 million image-text pairs. Some of the sample images from this dataset are shown in Figure \ref{fig:laion400m}.

\textbf{Visual Genome}~\citep{Krishna-IJCV-2017} contains over 100,000 images annotated with objects, attributes, and relationships. It includes region descriptions, object metadata, and dense annotations to facilitate scene understanding. It is mainly used within the visual question-answering task.

\begin{table*}[t!]
    \centering
    \caption{Statistics of the datasets that are commonly used in MIM literature.}
    \setlength\tabcolsep{4.6pt}
    \begin{tabular}{llcrr}
    \toprule
         Modality & Dataset & Type & \#Train Samples & \#Test Samples \\
         \midrule
         \multirow{7}{*}{Images} & ImageNet-1K~[\citenum{Deng-CVPR-2009}] & Pre-training \& Downstream & $1,281,167$ & $100,000$ \\
         & ImageNet-21K~[\citenum{Ridnik-NeurIPS-2021}] & Pre-training & $14,197,122$ & - \\
         & LAION-400M~[\citenum{Schuhmann-arxiv-2021}] & Pre-training & $400,000,000$ & -  \\
         & CIFAR-100~[\citenum{Krizhevsky-techrep-2009}] & Downstream & $60,000$  & $10,000$ \\
         & Food-101~[\citenum{Bossard-ECCV-2014}] & Downstream & $75,750$ & $25,250$ \\
         & MS COCO~[\citenum{Lin-ECCV-2014}] & Downstream & $165,482$ & $81,434$ \\
         & ADE20K~[\citenum{Bolei-CVPR-2017}] & Downstream & $25,574$ & $2,000$ \\
         & FFHQ~[\citenum{Karras-CVPR-2019}] & Downstream & $60,000$ & $10,000$ \\
         
        \midrule 
        \multirow{5}{*}{\begin{tabular}{l} Language \\
                                            \& Images
                \end{tabular}} & CC3M~[\citenum{Sharma-ACL-2018}] & Pre-training & $3,318,333$ & -\\
         &  CC12M~[\citenum{Changpinyo-CVPR-2021}] & Pre-training & $12,423,374$ & -\\
         & MS COCO~[\citenum{Lin-ECCV-2014}] & Pre-training & $165,482$ & - \\
        & Visual Genome~[\citenum{Krishna-IJCV-2017}] & Pre-training & $108,000$ & - \\
        & SBU Captions~[\citenum{Ordonez-NIPS-2011}] & Pre-training & $860,000$ & - \\
        \midrule
        \multirow{9}{*}{Video} & SSv2~[\citenum{Raghav-ICCV-2017}] & Pre-training \& Downstream & $169,000$ 
        & $25,000$\\
        & Kinetics-400~[\citenum{Kay-arxiv-2017}] & Pre-training \& Downstream & $240,000$ 
        & $20,000$\\
         & Kinetics-600~[\citenum{Carreira-arXiv-2018}] & Pre-training \& Downstream & $390,000$ 
        & $60,000$\\
         & Kinetics-700~[\citenum{Carreia-arXiv-2022}] & Pre-training \& Downstream & $545,317$ 
        & $105,000$\\
        & Kinetics-710~[\citenum{Li-ICCV-2023b}] & Pre-training & $ 660,000$ & - \\
        & WebVid2M~[\citenum{Bain-ICCV-2021}]& Pre-training & $2,500,000$& - \\
        & UnlabeledHybrid~[\citenum{wang-cvpr-2023c}] & Pre-training & $1,350,000$ & -\\
        & UCF101~[\citenum{Soomro-CoRR-2012}] & Downstream & $9,500$ & $3,500$ \\
        & AVA~[\citenum{Li-ArXiv-2020}] & Downstream & $211,000$ & $57,000$ \\
        \midrule
        \multirow{6}{*}{3D Data} & ScanNet~[\citenum{Dai-CVPR-2017}] & Pre-training & $1,513$ & -\\
        & ShapeNet~[\citenum{Chang-TechRep-2015}] & Pre-training & $57,448$ & - \\
        & nuScenes~[\citenum{Caesar-CVPR-2020}] & Pre-training \& Downstream & $750$ & $150$\\
        & ModelNet40~[\citenum{Wu-CVPR-2015}] & Downstream & $9,843$ & $2,468$ \\
        & ScanObjectNN~[\citenum{Uy-ICCV-2019}] & Downstream & $11,416$ & $2,882$ \\
        & ShapeNetPart~[\citenum{Li-ACM-2016}] & Downstream & $14,007$ & $2,874$ \\
        \midrule
         \multirow{10}{*}{\begin{tabular}{l} Medical \\
                                             Images
                \end{tabular}} & UKB~[\citenum{Spitzer-MICCAI-2018}] & Pre-training & $155,238$ & -\\
        & ROCO~[\citenum{Pelka-CVII-2018}] & Pre-training & $81,000$ & - \\
        & MedICaT~[\citenum{Subramanian-ACL-2020}] & Pre-training & $217,000$ & - \\
        & TCIA-COVID19~[\citenum{Harmon-Nature-2020}] & Pre-training & $771$ & - \\
        & BraTS~[\citenum{Simpson-Arxiv-2019}] & Pre-training \& Downstream & $351$ & $191$\\
        & BTCV~[\citenum{Landman-Synapse-2015}] & Pre-training \& Downstream & $24$ & $6$ \\
        
        & APTOS~[\citenum{Kaggle-APTOS-2019}] & Downstream &  $28,100$ & $7,026$ \\
        & RFMiD~[\citenum{Panchal-data-2023}] & Downstream & $2,560$ & $640$ \\
        & VQA-RAD~[\citenum{Lau-sdata-2018}] & Downstream & $3,064$  & $451$  \\
        & VQA-2019~[\citenum{Asma-CLEF-2019}] & Downstream & $12,792$  &  $500$ \\
        \bottomrule
    \end{tabular}
    \label{tab:dataset_stats}
\end{table*}

In Table~\ref{tab:dataset_stats}, we classify the datasets used for MIM based on their application, distinguishing between those employed for self-supervised pre-training and those used for downstream performance evaluation. Additionally, we provide the number of training and test samples for each dataset.

\section{Performance Overview}


We next provide an in-depth analysis of the performance achieved by MIM pre-training methods, when applied on leading computer vision architectures and popular benchmarks. We focus our analysis on three widely used datasets in MIM research: ImageNet-1K, MS COCO, and Kinetics-400.

In Table~\ref{tab:perf_IN1K}, we present the results of different frameworks on ImageNet-1K. 
Most of these works obtained accuracy improvements by introducing novel masking strategies~\citep{wang-cvpr-2023d, he-cvpr-2022, xie-cvpr-2022} or target features \citep{wei-cvpr-2022, wang-cvpr-2023d}. In general, the new masking strategies are based on the intuition that randomly masking patches, as proposed in MAE \citep{he-cvpr-2022} and SimMIM \citep{xie-cvpr-2022}, is suboptimal. Thus, recent methods \citep{wang-cvpr-2023d, li-neurips-2022a, madan-wacv-2024} aim to identify semantically important patches, usually implementing masking strategies in an easy-to-hard pipeline, starting by masking the least important patches and progressively targeting the more semantically significant ones. Other works~\citep{girdhar-cvpr-2023, yang-cvpr-2023} studied approaches to scale up the architectures and the number of samples used during pre-training. In contrast to these studies, some works~\citep{huang-iccv-2023b}  tried to limit the computational requirements of the networks, while preserving the performance of the previous works. Another significant observation that stems out from Table~\ref{tab:perf_IN1K} is that integrating MIM with generative models results in substantial performance enhancements, as presented in some recent studies~\citep{fei-cvpr-2023, wei-iccv-2023}. This finding supports the widely known view that generative tasks yield robust world representations. Moreover, it demonstrates the orthogonality between MIM and generative models, suggesting complementary benefits when combined.

\begin{table}[t!]
    \centering
    \caption{Performance on ImageNet-1K (IN1K) of different MIM pre-training schemes.}
    \setlength\tabcolsep{3.0pt}
    \begin{tabular}{lllc}
    \toprule
    \multirow{2}{*}{Backbone} &\multicolumn{2}{c}{Pre-training}  & \multirow{2}{*}{Acc.} \\
    \cmidrule{2-3}
     & Dataset & Method& \\
    \midrule
    \multirow{20}{*}{ViT-B}& \multirow{15}{*}{IN1K} & MAGE~[\citenum{li-cvpr-2023a}] & 82.5 \\
    & & MAGE-C~[\citenum{li-cvpr-2023a}] & 82.9 \\
    & & U-MAE~[\citenum{zhang-neurips-2022}] & 83.0\\
    & & Prototypical~[\citenum{lin-CVPR-2025}] & 83.3 \\
    & & MAP~[\citenum{liu-CVPR-2025}] & 83.6 \\
    & & MAE~[\citenum{he-cvpr-2022}] & 83.6\\
    & & SimMIM~[\citenum{xie-cvpr-2022}] & 83.8\\
    & & LocalMIM~[\citenum{wang-cvpr-2023b}] & 84.0\\
    & & MaskFeat~[\citenum{wei-cvpr-2022}] & 84.0\\
    & & DMAE-B~[\citenum{bai-cvpr-2023}] & 84.0 \\
    & & HPM~[\citenum{wang-cvpr-2023d}] & 84.2\\
    & & GAN-MAE~[\citenum{fei-cvpr-2023}] & 84.3\\
    & & SemMAE~[\citenum{li-neurips-2022a}] & 84.5 \\
    & & DiffMAE~[\citenum{wei-iccv-2023}] & 84.9 \\
    & & MCMAE~[\citenum{gao-neurips-2022}] & 85.0 \\
    & & CMAE~[\citenum{huang-tpami-2023}] & 85.3 \\
    & & MaskAlign~[\citenum{xue-cvpr-2023}] & 85.4 \\
    \cmidrule{2-4}
    & Laion-50M & RILS~[\citenum{yang-cvpr-2023}] & 83.6 \\
    \cmidrule{2-4}
    & IN1K+SSv2 & OmniMAE~[\citenum{girdhar-cvpr-2023}] &  83.0\\
    \midrule
    \multirow{11}{*}{ViT-L} & \multirow{9}{*}{IN1K} & U-MAE~[\citenum{zhang-neurips-2022}] & 83.2\\
    & & MAGE~[\citenum{li-cvpr-2023a}] & 83.9 \\
    & & MAGE-C~[\citenum{li-cvpr-2023a}] & 84.3 \\
    & &  MaskFeat~[\citenum{wei-cvpr-2022}] & 85.7\\
    & & HPM~[\citenum{wang-cvpr-2023d}] & 85.8\\
    & & LocalMIM~[\citenum{wang-cvpr-2023b}] & 85.8\\
    & & MAE~[\citenum{he-cvpr-2022}] & 85.9\\
    & & GAN-MAE~[\citenum{fei-cvpr-2023}] & 86.1\\
    & & {MAP}~[\citenum{liu-CVPR-2025}] & {86.1}\\
    & & DiffMAE~[\citenum{wei-iccv-2023}] & 86.9 \\
     \cmidrule{2-4}
    & IN1K+SSv2 & OmniMAE~[\citenum{girdhar-cvpr-2023}] &  85.2\\
    \midrule
    \multirow{3}{*}{ViT-H} & \multirow{2}{*}{IN1K}  &  MAE~[\citenum{he-cvpr-2022}] & 86.9\\
    & & DiffMAE~[\citenum{wei-iccv-2023}] & 88.0\\
     \cmidrule{2-4}
    & IN1K+SSv2 & OmniMAE~[\citenum{girdhar-cvpr-2023}] &  86.6\\
    \midrule
    ViT-H$_{448}$ & IN1K & MAE~[\citenum{he-cvpr-2022}]  & 87.8 \\
    \midrule
    \multirow{4}{*}{Swin-B} &  \multirow{4}{*}{IN1K} & GreenMIM~[\citenum{huang-neurips-2022b}]& 83.8\\
     & & SimMIM~[\citenum{xie-cvpr-2022}] & 84.0 \\
     & & LocalMIM~[\citenum{wang-cvpr-2023b}] & 84.1\\
     & & MixMAE~[\citenum{liu-cvpr-2023a}] & 84.6\\
    \midrule
    \multirow{3}{*}{Swin-L} & \multirow{3}{*}{IN1K} & GreenMIM~[\citenum{huang-neurips-2022b}]& 85.1\\
     &  & SimMIM~[\citenum{xie-cvpr-2022}] & 85.4 \\
     & & LocalMIM~[\citenum{wang-cvpr-2023b}] & 85.6\\
    \midrule
    SwinV2-H& \multirow{1}{*}{IN1K} & SimMIM~[\citenum{xie-cvpr-2022}] & 85.7 \\
    \midrule
    SwinV2-G & IN21K & SimMIM~[\citenum{xie-cvpr-2022}] & 90.2 \\
    \midrule
    ConvNeXt V2-B & IN1K & FCMAE~[\citenum{woo-cvpr-2023}] & 84.9\\
    \midrule
    ConvNeXt V2-L & IN1K & FCMAE~[\citenum{woo-cvpr-2023}] & 85.8\\
    \midrule
    ConvNeXt V2-H & IN1K & FCMAE~[\citenum{woo-cvpr-2023}] & 86.3\\
    \bottomrule
    \end{tabular}
    \label{tab:perf_IN1K}
\end{table}

In Table~\ref{tab:perf_COCO}, we present a performance overview of different MIM frameworks on the MS COCO dataset. For object detection, we report the mean Average Precision for bounding boxes (mAP$_{box}$), and for segmentation tasks, we provide the mean Average Precision for masks (mAP$_{mask}$). EVA~\citep{fang-cvpr-2023} stands out as the most effective method, concentrating on the expansion of training examples and network parameters. According to Table~\ref{tab:perf_COCO}, its pre-training dataset integrates four distinct datasets, collectively summing over 29 million images. The closest result to EVA is obtained by SimMIM~\citep{xie-cvpr-2022}, when applied to a similar size architecture (SwinV2-G), but with much fewer images in the pre-training phase. 
As demonstrated in Table~\ref{tab:perf_COCO}, the MAE strategy exhibits suboptimal performance when compared with alternative approaches applied to ViT-B and ViT-L models. Notably, among the methods outperforming MAE on ViT-B, two incorporate convolutional layers into their frameworks~\citep{fang-iccv-2023, gao-neurips-2022}. This observation suggests that employing a hybrid architecture that combines both transformer and convolutional layers may offer significant improvements in object detection and segmentation.

In terms of training objectives, based on the results reported in Tables~\ref{tab:perf_IN1K} and \ref{tab:perf_COCO}, the reconstruction loss remains the primary driver for high performance in MIM methods across both ImageNet-1K and MS COCO. At the same time, the contrastive loss often serves as a complementary component. For example, one of the top-performing methods, CMAE~\citep{huang-tpami-2023}, leverages both reconstruction and contrastive losses.

\begin{table*}[t!]
    \centering
    \caption{Performance on MS COCO of different MIM pre-training schemes.}
    \begin{tabular}{lllcc}
    \toprule
    \multirow{2}{*}{Backbone} &\multicolumn{2}{c}{Pre-training}  & \multirow{2}{*}{mAP$_{box}$} & \multirow{2}{*}{mAP$_{mask}$} \\
    \cmidrule{2-3}
     & Dataset & Method & & \\
     \midrule 
     \multirow{2}{*}{ViT-Tiny} & \multirow{2}{*}{IN1K} & MAE~[\citenum{he-cvpr-2022}] & 38.9 & 35.1\\
     & & SparseMAE~[\citenum{zhou-iccv-2023}] & 47.1 & 42.0 \\
    \midrule
    \multirow{6}{*}{ViT-B}& Laion-20M & RILS~[\citenum{yang-cvpr-2023}] & 48.5 & 42.6 \\
    \cmidrule{2-5}
    & \multirow{5}{*}{IN1K}& GAN-MAE~[\citenum{fei-cvpr-2023}] &49.0 & 43.8 \\
    &  & MAE~[\citenum{he-cvpr-2022}]  & 50.3  & 44.9  \\
    & & MaskAlign~[\citenum{xue-cvpr-2023}] & 52.1 & 45.7 \\
    & & MCMAE~[\citenum{gao-neurips-2022}] & 52.5 & 46.5 \\
    & & CMAE~[\citenum{huang-tpami-2023}] & 52.9 & 47.0 \\
    \midrule
    \multirow{2}{*}{ViT-L} & \multirow{2}{*}{IN1K} & MAE~[\citenum{he-cvpr-2022}] & 53.3 & 47.2\\
    & & DiffMAE~[\citenum{wei-iccv-2023}] & 55.3 & 49.0 \\
    \midrule
    \multirow{4}{*}{ViT-G} & IN21K& \multirow{4}{*}{EVA~[\citenum{fang-cvpr-2023}]} &\multirow{4}{*} {64.2} & \multirow{4}{*}{55.0}\\
    & CC12M & & &\\
    & CC3M & & &\\
    & COCO & & &\\
    \midrule
    \multirow{1}{*}{Swin-T} & IN1K & MST~[\citenum{li-neurips-2021}] & 42.7 & 38.8 \\
    \midrule
    \multirow{3}{*}{Swin-B} & \multirow{3}{*}{IN1K} & GreenMIM~[\citenum{huang-neurips-2022b}] & 50.0 
    & 44.1 \\
    & & SimMIM~[\citenum{xie-cvpr-2022}] & 50.4 & 44.4 \\
    & & LocalMim~[\citenum{wang-cvpr-2023b}] &  50.7 & 44.9\\
    \midrule
    \multirow{1}{*}{SwinV2-G} & IN22K & SimMIM~[\citenum{xie-cvpr-2022}] & 63.1 & 54.4 \\
  
    \midrule
    MIMDET-Base~[\citenum{fang-iccv-2023}] & \multirow{2}{*}{IN1K} & \multirow{2}{*}{MAE~[\citenum{he-cvpr-2022}]}& 51.7&  46.1\\
    MIMDET-Large~[\citenum{fang-iccv-2023}] & & &54.3 & 48.2 \\
    
    \midrule
    \multirow{1}{*}{{HybridNet-Ti}~[\citenum{liu-CVPR-2025}]} & {COCO} & {MAP}~[\citenum{liu-CVPR-2025}] & {46.4} & {39.8} \\
    
    \bottomrule
    \end{tabular}
    \label{tab:perf_COCO}
\end{table*}

Results on video recognition are included in Table~\ref{tab:perf_K400}. 
We select the Kinetics-400 dataset for this analysis because it is the most frequently used video dataset across the reviewed studies. A common practice among the studies listed in Table~\ref{tab:perf_K400} is their incorporation of large-scale image datasets during the pre-training phase, alongside a video dataset. Additionally, an analysis of the performance outcomes from EVA~\citep{fang-cvpr-2023} and VideoMAEv2~\citep{wang-cvpr-2023c} on the ViT-G architecture reveals that the size of the pre-training dataset plays an important role in determining the final performance of the model. Larger datasets tend to provide more effective and generalized capabilities in complex video processing tasks.

Consistent with what we observed in image classification, the integration of diffusion models and MIM proves beneficial in the video domain as well. Notably, the DiffMAE model~\citep{wei-iccv-2023} achieves results that are competitive with those of EVA~\citep{fang-cvpr-2023}, although DiffMAE is operating with a substantially smaller model size. This finding underscores again the effectiveness of combining generative diffusion models with MIM techniques. However, a significant factor contributing to this performance enhancement is the utilization of a large dataset, WIT400M, during the pre-training phase. When the model is pre-trained solely with the smaller Kinetics-400 dataset, its results fall short of those achieved by VideoMAE~\citep{tong-neurips-2022}. This underscores the critical importance of employing large-scale datasets in the pre-training phase to maximize model effectiveness.

\begin{table}[t!]
    \centering
    \caption{Performance on Kinetics-400 (K400) of different MIM pre-training schemes.}
    \setlength\tabcolsep{3.5pt}
    \begin{tabular}{lllc}
    \toprule
    \multirow{2}{*}{Backbone} &\multicolumn{2}{c}{Pre-training}  & \multirow{2}{*}{Acc.} \\
    \cmidrule{2-3}
     & Dataset & Method& \\
    \midrule
     \multirow{3}{*}{ViT-S} & \multirow{1}{*}{K400} & VideoMAE~[\citenum{tong-neurips-2022}] & 79.0\\
     \cmidrule{2-4}
     & K400 & \multirow{2}{*}{MVD~[\citenum{wang-cvpr-2023a}]}& \multirow{2}{*}{81.0} \\
     & IN1K & & \\
     
     \midrule
     \multirow{10}{*}{ViT-B} & \multirow{5}{*}{K400} & {SMILE}~[\citenum{thoker-CVPR-2025}] & {83.1}\\
     & & VideoMAE~[\citenum{tong-neurips-2022}] & 81.5\\
     & & ST-MAE~[\citenum{feichtenhofer-neurips-2022}] & 81.3\\
     & & MME~[\citenum{sun-cvpr-2023}] & 81.8\\
     & & MGMAE~[\citenum{huang-iccv-2023b}] & 81.8 \\
     & & OmniMAE~[\citenum{girdhar-cvpr-2023}] & 80.8\\
     \cmidrule{2-4}
     & K400 & \multirow{2}{*}{MVD~[\citenum{wang-cvpr-2023a}]}& \multirow{2}{*}{83.4} \\
     & IN1K &  & \\
     \cmidrule{2-4}
     & \multirow{1}{*}{UnlabeledHybrid} & VideoMAEv2~[\citenum{wang-cvpr-2023c}]& 81.5\\
     
    \midrule
     \multirow{14}{*}{ViT-L} & \multirow{4}{*}{K400} & {UT}~[\citenum{li-iccv-2023}] & {90.9} \\
     &  & VideoMAE~[\citenum{tong-neurips-2022}] & 85.2\\
     & & ST-MAE~[\citenum{feichtenhofer-neurips-2022}] & 84.8 \\
     & & DiffMAE~[\citenum{wei-iccv-2023}] & 84.5 \\
     \cmidrule{2-4}
     & K400 & \multirow{2}{*}{MVD~[\citenum{wang-cvpr-2023a}]} & \multirow{2}{*}{86.4}\\
     & IN1K & & \\
     \cmidrule{2-4}
     & K400& \multirow{3}{*}{OmniMAE~[\citenum{girdhar-cvpr-2023}]} & \multirow{3}{*}{84.0} \\
     & SSv2 & & \\
     & IN1K & & \\
     \cmidrule{2-4}
     & K400 & \multirow{2}{*}{DiffMAE~[\citenum{wei-iccv-2023}]} & \multirow{2}{*}{88.1} \\
     & WIT400M & & \\
     \cmidrule{2-4}
     & \multirow{1}{*}{UnlabeledHybrid} 
     & VideoMAEv2~[\citenum{wang-cvpr-2023c}]& 85.4\\
     
     \midrule
      \multirow{10}{*}{ViT-H}& \multirow{3}{*}{K400} & {VideoMAEv2}~[\citenum{wang-cvpr-2023c}] & {88.6}\\
     & & VideoMAE~[\citenum{tong-neurips-2022}] & 86.6\\
     & & ST-MAE~[\citenum{feichtenhofer-neurips-2022}] & 85.1 \\
     \cmidrule{2-4}
     & K400& \multirow{2}{*}{MVD~[\citenum{wang-cvpr-2023a}]} & \multirow{2}{*}{87.3}\\
     & IN1K & & \\
     \cmidrule{2-4}
     & K400 & \multirow{3}{*}{OmniMAE~[\citenum{girdhar-cvpr-2023}]} & \multirow{3}{*}{84.8} \\
     & SSv2 & & \\
     & IN1K & & \\
     \cmidrule{2-4}
      & \multirow{1}{*}{UnlabeledHybrid} & VideoMAEv2~[\citenum{wang-cvpr-2023c}]& 86.9\\
     
     \midrule
      \multirow{7}{*}{ViT-G}& \multirow{1}{*}{UnlabeledHybrid} & VideoMAEv2~[\citenum{wang-cvpr-2023c}] & 87.2\\
      \cmidrule{2-4} &
      \multirow{1}{*}{K400} & {VideoMAEv2}~[\citenum{wang-cvpr-2023c}] & {90.0}\\
      \cmidrule{2-4}
       & IN21K& \multirow{4}{*}{EVA~[\citenum{fang-cvpr-2023}]} &\multirow{4}{*} {89.7} \\
        & CC12M &  &\\
        & CC3M &  &\\
        & COCO &  &\\
    \midrule
    \multirow{1}{*}{MViTv2-S} &K400& MaskFeat~[\citenum{wei-cvpr-2022}] & 82.2\\
    \midrule
    
    \multirow{1}{*}{MViTv2-L} &K400& MaskFeat~[\citenum{wei-cvpr-2022}] & 86.7\\
  \bottomrule   
    \end{tabular}
    \label{tab:perf_K400}
\end{table}

Other significant observations, based on the results presented in Table \ref{tab:perf_K400}, refer to the masking strategy and the types of features that are reconstructed. First, in videos, the temporal correlation between frames~\citep{tong-neurips-2022} can introduce a bias in the model, allowing it to reconstruct masked patches by leveraging information from adjacent frames. To alleviate this issue, OmniMAE~\citep{girdhar-cvpr-2023} randomly masks patches from videos, similar to MAE~\citep{he-cvpr-2022}, but uses a higher masking ratio, $95\%$ instead of $75\%$. However, this masking strategy performs poorly compared with those used in VideoMAE~\citep{tong-neurips-2022} and VideoMAEv2~\citep{wang-cvpr-2023c}, which are designed to mask temporal tubes instead of spatial patches. The latter approach is more effective in preventing information leakage from adjacent frames. Second, the top performing methods, such as MVD~\citep{wang-cvpr-2023a} and EVA~\citep{fang-cvpr-2023}, benefit from the use of high-level features as target features for reconstruction. These features help address the tendency of pixel-level MIM models to focus on spatial and structural information, leading to improved performance on downstream tasks.

\section{Closing Remarks and Future Directions}
In this paper, we highlighted two strategies for applying masked image modeling, one based on reconstruction and one based on contrastive learning. Moreover, we presented how both methods are great pre-training approaches for feature learning. Although their objectives are different, the theoretical analysis shows that they are equivalent. Furthermore, we provided a review of the most recent research advancements in masked image modeling, and explained how this pre-training strategy was implemented for various tasks. 

Through our work, we aimed to give a better overview of masked image modeling, simplifying the effort needed by the research community and the industry to analyze the literature. We believe that both the manual taxonomy and the hierarchical clustering dendrogram are great resources for all individuals interested in learning more about masked image modeling, or how to apply this technique to their specific use case.

As previously mentioned, masked pre-training started from natural language processing, and it was later adopted in vision. Over time, masked image modeling was integrated into multiple downstream tasks and this is perhaps the main research direction that will continue, especially in domains or tasks with low quantities of annotated data, such as the medical domain.

While masked image modeling (MIM) has achieved remarkable progress in self-supervised visual representation learning, several promising directions remain open for future exploration. One important area is the extension of MIM to multi-modal and cross-modal learning scenarios, such as integrating vision with language or audio signals, which could further enhance contextual understanding. Each modality holds a different type of information, and these can be combined to learn richer feature representations. Distilling the knowledge of each modality into a single network could represent a stepping stone in artificial intelligence.

Another avenue is the development of more efficient masking strategies and reconstruction objectives that better align with downstream tasks beyond classification, such as segmentation, detection, or 3D understanding. While a random strategy may perform well, it has been demonstrated that an informed masking policy that focuses on hiding the salient information of the input is superior \citep{madan-wacv-2024}. Thus, future endeavors may attempt to formulate various guided masking strategies, some of which can be specific to the downstream task. Additionally, there is growing interest in making MIM more resource-efficient.

Another future direction is the integration of physical priors or domain-specific laws into the MIM framework. For domains where data is governed by known physical constraints (e.g.~medical imaging, climate science, robotics), embedding known physical constraints or conservation laws into the masking or reconstruction process could guide the model toward more meaningful and generalizable representations.

Finally, understanding the theoretical underpinnings of why and how MIM works so effectively remains an open challenge. Insights on this path could drive the design of next-generation self-supervised learning algorithms.




\section*{Declarations}

\textbf{Funding.} This research is supported by the project ``Romanian Hub for Artificial Intelligence - HRIA'', Smart Growth, Digitization and Financial Instruments Program, 2021-2027, MySMIS no.~334906. This work was also supported by a grant of the Ministry of Research, Innovation and Digitization, CCCDI - UEFISCDI, project number PN-IV-P7-7.1-PED-2024-1856, within PNCDI IV.

\vspace{0.2cm}
\noindent
\textbf{Conflict of interest.} The authors have no conflicts of interest to declare that are relevant to the content of this article.

\vspace{0.2cm}
\noindent
\textbf{Availability of data and materials.} This paper is a survey paper, hence it does not report new experiments on any dataset.

\backmatter

\begin{appendices}




\end{appendices}


{\small
\bibliography{sn-bibliography}
\bibliographystyle{sn-vancouver}
}

\end{document}